\documentclass[twoside,11pt]{article}

\usepackage[table,xcdraw]{xcolor}
\usepackage{jmlr2e}
\usepackage[utf8]{inputenc}
\usepackage{hyperref }
\usepackage[utf8]{inputenc} 
\usepackage[T1]{fontenc}    
\usepackage{hyperref}       
\usepackage{url}            
\usepackage{booktabs}       
\usepackage{amsmath}
\usepackage{nicefrac}       
\usepackage{microtype}      
\usepackage{optidef}
\usepackage{natbib}
\usepackage{amsmath,amsfonts,amssymb}
\usepackage{bbm}
\usepackage{caption}
\usepackage{subcaption}
\usepackage{mathtools}
\usepackage{booktabs}  
\usepackage{multirow}

\DeclarePairedDelimiter{\ceil}{\lceil}{\rceil}

\newtheorem{prop}{Proposition}

\firstpageno{1}
\usepackage{lastpage}
\jmlrheading{26}{2025}{1-\pageref{LastPage}}{2/24; Revised
4/25}{7/25}{24-0255}{Brian Liu and Rahul Mazumder}
\ShortHeadings{Randomization Can Reduce Both Bias and Variance}{Liu and Mazumder}
\begin{document}

\title{Randomization Can Reduce Both Bias and Variance: A Case Study in Random Forests}

\author{\name Brian Liu \email briliu@mit.edu \\
       \addr Operations Research Center\\
       Massachusetts Institute of Technology\\
       Cambridge, 02139-4307, USA
       \AND
       \name Rahul Mazumder \email rahulmaz@mit.edu \\
       \addr Operations Research Center and Sloan School of Management\\
   Massachusetts Institute of Technology\\
       Cambridge, 02139-4307, USA}

\editor{Florence d'Alche-Buc}

\maketitle

\begin{abstract}
We study the often overlooked phenomenon, first noted in \cite{breiman2001random}, that random forests appear to reduce bias compared to bagging. Motivated by an interesting paper by \cite{mentch2020randomization}, where the authors explain the success of random forests in low signal-to-noise ratio (SNR) settings through regularization, we explore how random forests can capture patterns in the data that bagging ensembles fail to capture. We empirically demonstrate that in the presence of such patterns, random forests reduce bias along with variance and can increasingly outperform bagging ensembles when SNR is high. Our observations offer insights into the real-world success of random forests across a range of SNRs and enhance our understanding of the difference between random forests and bagging ensembles. Our investigations also yield practical insights into the importance of tuning $mtry$ in random forests.


\end{abstract}


\section{Introduction}

For over two decades, random forests \citep{breiman2001random} have enjoyed widespread acclaim in machine learning and remain among the most competitive off-the-shelf supervised learning algorithms today. Random forest  combine excellent predictive performance, comparable to boosting algorithms \citep{caruana2006empirical}, with ease-of-use, since the algorithm is embarrassingly parallelizable and robust to overfitting. As such, random forests are applied in practice across a variety of disciplines that range from finance \citep{liu2015financial} to bioinformatics \citep{diaz2006gene}.

Remarkably, random forests are a simple modification of the original bootstrap aggregating (bagging) algorithm proposed by \cite{breiman1996bagging} to construct tree ensembles. In bagging, a dataset is sampled with replacement repeatedly and a decision tree is fit on each sample. The sampled datasets, which we refer to as bags, are typically of the same size as the original dataset. The predictions of the trees are combined across bags by averaging (regression) or voting (classification) to produce the final prediction of the ensemble. Random forests repeat this procedure with just a single modification. For each split in each tree in the ensemble, only a random subset of  $mtry \in (0,1)$ fraction of the variables in the dataset are considered when constructing the split. This simple modification can  increase the predictive performance of the random forest compared to bagging, as empirical experiments in \cite{breiman2001random} and \cite{caruana2006empirical} show. 

Many empirical explanations for the success of bagging tree ensembles have been proposed in literature. Classically, \cite{friedman1997bias} argues that bagging reduces the variance of an ensemble without substantially increasing ensemble bias. \cite{domingos1997does}, on the other hand, proposes that bagging approximates the optimal procedure for Bayesian model averaging by sampling from the model space. \cite{buhlmann2002analyzing} argues that bagging softens the hard decision surfaces generated by trees, which reduces both ensemble bias and variance. \cite{grandvalet2004bagging} proposes an orthogonal explanation, that bagging equalizes the influence of high leverage points in the data. Sampling with replacement ensures that a single high-leverage point appears in just two-thirds of the generated bags. More recently, \cite{wyner2017explaining} argues that bagging ensembles and by extension random forests, work well because they interpolate the training data while isolating the effects of noisy data points. 

In stark contrast, few empirical explanations exist for why \emph{additional} randomization in a random forest improves ensemble performance over bagging, and the main focus of our paper is to investigate this phenomenon. Canonically, \cite{breiman2001random} suggests that randomizing the features considered per split reduces the correlation between trees, which in turn further reduces the variance of the ensemble. A textbook example from \cite{hastie2009elements} illustrates this effect well; the variance of the average of $B$ i.i.d random variables (trees) each with variance $\sigma^2$ is given by:
\begin{equation}
    \gamma \sigma^2 + \frac{1-\gamma}{B} \sigma^2,
\end{equation}
where $\gamma$ is the pairwise correlation between variables. Given a sufficiently large number of trees $B$, the variance of the average is dominated by the first term. So reducing pairwise correlation reduces variance. While variance reduction in random forests is well understood, Breiman suggests that other factors may contribute to the success of random forests over bagging. In the concluding remarks of \cite{breiman2001random}, Breiman mentions that the accuracy of random forests indicates that they "act to reduce bias" but that the "mechanism for this is not obvious." However, this notion that random forests can reduce bias compared to bagging appears to have been largely overlooked in subsequent works. In fact, the second edition of the canonical textbook \textit{The Elements of Statistical Learning} \citep{hastie2009elements} states in Chapter 15 that "the only hope of improvement [of random forests and bagging] is through variance reduction." As such, the idea that random forests improve performance solely through variance reduction appears to be widely accepted in statistical lore.



Building on the idea that random forests primarily benefit from variance reduction, a recent interesting paper by  \cite{mentch2020randomization} proposes the novel argument that randomization in random forests reduces the effective degrees of freedom (DoF) of the model relative to bagging, and, as such, regularizes the model to outperform bagging ensembles in low signal-to-noise (SNR) ratio settings. The authors demonstrate that this regularization effect occurs through simulation studies and semi-synthetic experiments.

Motivated by the interesting arguments in \cite{mentch2020randomization}, we seek to better explain the success of random forests when SNR is high. In our view, real-world problems span a wide range of SNRs \citep{mazumder2020discussion} and random forests work well in applications that range from financial modeling, where SNR is typically low, \citep{booth2014automated,lohrmann2019classification,baba2020predicting} to signal processing where SNR can be quite high \citep{cerrada2016fault,xu2023research,saki2016smartphone}. A complementary explanation for the success of random forests may be required to account for the popularity of the algorithm in high SNR fields.


In this paper, we revisit the widely-held belief that random forests {\emph{only}} reduce variance and show how random forests can reduce bias compared to bagging, which helps explain their success in high SNR settings. We introduce and formalize the notion of \emph{hidden patterns}, patterns in the data that bagging ensembles have difficulty capturing. By randomly selecting  subsets of features to consider per split, random forests better capture these patterns, which in turn reduces ensemble bias. To provide intuition on hidden patterns, we show an example below.

\subsection*{Hidden Pattern Motivating Example}
\begin{figure}[h]
    \centering
    \includegraphics[width = \textwidth]{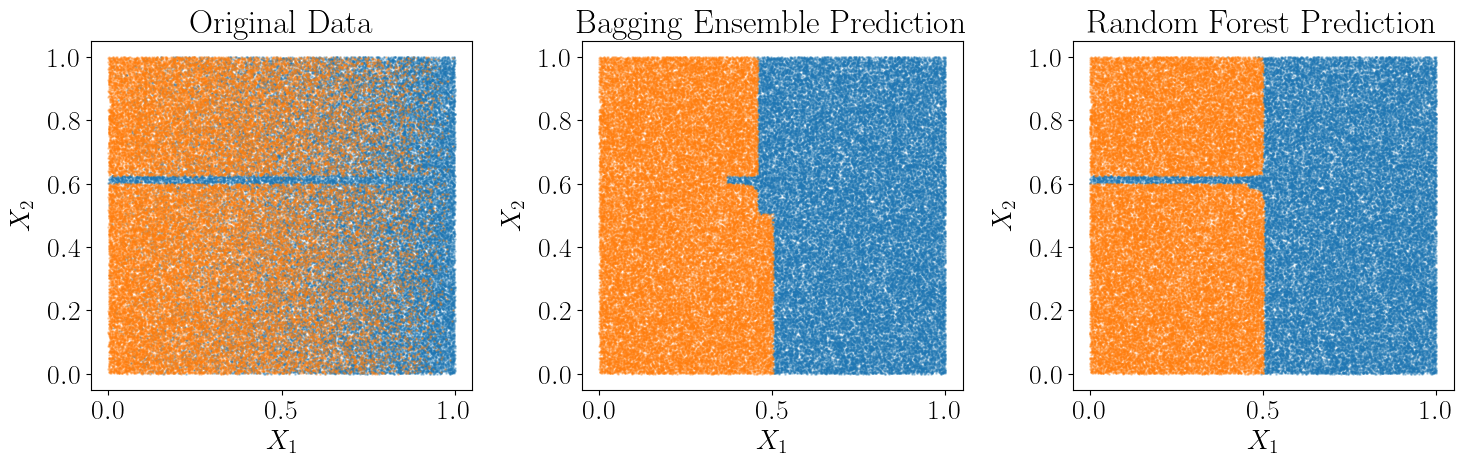}
    \caption{The relationship between $X_2$ and $Y$ is hidden from a bagging ensemble by $X_1$ due to the myopic nature of how splits in a decision tree are constructed.}
    \label{hidden_pattern.fig}
\end{figure}
We consider a 2-dimensional grid of points ($X_1$, $X_2$) where $X_1, X_2 \in [0,1]$ and uniformly generate 10000 points within this grid. We assign these points to classes $Y \in \{0,1\}$ following these rules. All points with $0.6 \leq X_2 \leq 0.65$ are assigned to the positive class ($Y=1$) with probability 0.9. All other points are assigned to the positive class with probability $X_1$.

The leftmost plot in Figure \ref{hidden_pattern.fig} shows a visualization of this data, with positive class points marked in blue and negative class points marked in orange. We see from this plot that we need two splits to capture the relationship between $X_2$ and $Y$: one on $X_2 \leq 0.65$ and one on $X_2 \geq 0.6$. The relationship between $X_1$ and $Y$, on the other hand, can be fairly well captured by a single split down the middle: $X_1 \leq 0.5$.

The relationship between $X_2$ and $Y$ is an example of a hidden pattern. Due to the myopic nature of how splits are constructed, decision trees in a bagging ensemble will typically split on $X_1$ before $X_2$, and, as a result, the relationship between $X_2$ and $Y$ may be hidden from the bagging ensemble by $X_1$. The middle plot in Figure \ref{hidden_pattern.fig} shows the predictions/decision boundaries of a bagging ensemble trained on the data, and we observe that this ensemble fails to capture the relationship between $X_2$ and $Y$. Random forests, however, can uncover hidden patterns by randomly selecting subsets of features to consider per split. As we can see from the rightmost plot in Figure \ref{hidden_pattern.fig}, the decision boundaries of a random forest fit on the data better capture the relationship between $X_2$ and $Y$ compared to bagging ensembles. Later in this paper, we formally define hidden patterns and empirically demonstrate how random forests can reduce the training error and bias of the model through uncovering these patterns.

\subsection*{Main Contributions}
We highlight our main contributions below.
\begin{itemize}

\item We revisit the 
popular belief that random forests {\emph{solely}} reduce variance and we show how random forests can reduce bias.

    \item We introduce and describe the notion of hidden patterns, relationships in the data that bagging ensembles have difficulty capturing, and show that random forests better capture these relationships by randomly selecting the subset of features considered per split.

    \item We show that on datasets with hidden patterns, random forests increasingly outperform bagging ensembles as SNR increases. This finding complements the arguments presented in \cite{mentch2020randomization} by offering additional insight into the success of random forests over bagging when SNR is high.

    \item We empirically demonstrate that random forests reduce bias in addition to variance relative to bagging on datasets with hidden patterns.
    
    \item Our investigation yields the important practical takeaway that $mtry$ should be tuned. We explore how $mtry$ controls the ability of a random forest to reduce bias and analyze how this parameter is sensitive to non-informative features in the data.

\end{itemize}

\section{Preliminaries}

We first introduce notation and provide an overview of bagging ensembles and random forests. We then explore in detail the main arguments in \cite{mentch2020randomization}: that random forests reduce effective DoF compared to bagging, which allows them to increasingly outperform bagging ensembles as SNR decreases. Interestingly, we show here that trimming trees in a bagging ensemble can reduce effective DoF by an equivalent degree to that of random forest, \emph{without} improving performance. This finding suggests that random forests have additional advantages over bagging beyond reducing effective DoF, motivating our investigation into how random forests reduce bias.

\subsection{Notation and Background}
 Let $X \in \mathbb{R}^{n \times p}$ represent our data matrix of $n$ rows and $p$ columns and let $Y \in \mathbb{R}^n$ represent the prediction target. Let $X^{(i)}$ denote the $i^{\text{th}}$ row (observation) in data matrix $X$ and let $X_j$ denote the $j^\text{th}$ column (feature). Assume that we have some, potentially unknown, data-generating procedure that can be represented by function $f: \mathbb{R}^{n\times p} \rightarrow \mathbb{R}^n$ that operates on data matrix $X$. The relationship between $X$ and $Y$ is given by $Y = f(X) + \epsilon$ where $\epsilon \in \mathbb{R}^n$ is a vector of errors. 

Bagging ensembles and random forests are made up of decision trees, and a tree $T(X)$ maps data matrix $X$ to prediction vector $\hat{f}(X) \in \mathbb{R}^n$. Trees consist of internal nodes and leaf nodes. Each internal node represents a split of the feature space on a single feature, and the leaf nodes collectively partition the entire space. Trees are grown greedily, from root to leaf, and each split is selected to minimize the impurity of the directly resulting nodes \citep{breiman2017classification}. Each leaf node assigns a prediction value to the corresponding partition of the feature space, typically the mean of the training data points in the partition for regression or the majority class for classification. Decision trees are highly flexible models; growing a tree to full-depth, where each leaf node contains just a single data point, can produce a model with 0 training loss. However, full-depth trees exhibit high variance and may perform poorly out-of-sample \citep{breiman2017classification, hastie2009elements}.

Bagging ensembles follow this procedure to reduce variance. We generate  $X^1, \ldots, X^m \in \mathbb{R}^{n \times p} $ datasets (bags) by sampling with replacement rows of $X$. We fit a  decision tree $T^k(X)$ on each bag and the trees are averaged to output the final prediction (for regression):
\[\hat{f}(X) = \frac{1}{m}\sum_{k=1}^m T^k(X). \] To form a random forest, we add the restriction that when building each tree, we only consider $mtry < 1.0$ proportion of the features when constructing each split. We refer to this procedure as \textbf{split feature subsetting (SFS)}.

We make the distinction between bagging ensembles and random forests that random forests are fit with $mtry < 1.0$ and bagging ensembles are fit with $mtry = 1.0$, and we emphasize that this leads to estimators with very different operating characteristics. To highlight this distinction, we refer to random forests as SFS ensembles throughout this paper. In other works, random forests are often defined to have $mtry \leq 1.0$, and bagging ensembles are viewed as a special case of random forests with $mtry = 1.0$. Since our paper focuses on the comparison between random forests and bagging we take this additional step to distinguish the two procedures.

\color{black}
As an aside, we briefly discuss tree size in random forests. In the original random forest paper, Breiman recommends growing decision trees to their full depth, where each leaf node contains just a single observation \citep{breiman2001random, zhou2023trees}. This recommendation  is followed by  popular software implementations of the random forest algorithm \citep{randomForestR,scikit-learn}. However, subsequent works suggest that tuning tree depth has the potential to improve the predictive performance of a random forest when SNR is low \citep{zhou2023trees,lin2006random}. Since we are interested in bias reduction in random forests when SNR is high, we focus primarily on random forests with full-depth trees.
\color{black}

\subsection{Random Forest and Effective Degrees of Freedom} \label{rf_effective_dof.section}

\cite{mentch2020randomization} in their interesting paper empirically demonstrate that reducing $mtry$, which controls the proportion of features to consider in SFS, reduces the effective degrees of freedom (DoF) of the ensemble \citep{efron1986bootstrap}. Here, we note that procedures other than SFS can reduce the effective DoF of bagging ensembles. For example, \cite{mentch2020randomization} (Section 3) mentions that limiting the maximum number of leaf nodes per tree ($maxnodes$), a procedure we will refer to as \textbf{TRIM}, naturally reduces  effective DoF. In the next section, we empirically demonstrate that while SFS ensembles outperform bagging when SNR is low, TRIM ensembles, tuned to have the same effective DoF as SFS ensembles, do not outperform bagging under any of the SNR regimes in the experiments in \cite{mentch2020randomization}. Consequently, we believe that SFS randomization in random forests goes beyond reducing effective DoF and model variance, and may in fact reduce model bias compared to bagging. Exploring the mechanisms behind this bias reduction in random forests is the main focus of our paper.

\color{black}
For consistency, we use the same procedure 
discussed in 
\cite{mentch2020randomization} and 
\cite{efron1986bootstrap} to estimate 
effective DoF. \color{black} Let function $\hat{f}$ 
produce estimates $\hat{y}_1, \ldots, \hat{y}
_n$, and assume that these estimates 
differ from the true values $y_1, \ldots, 
y_n$ by errors $\epsilon_1, \ldots, 
\epsilon_n$, which are independent Gaussian $N(0,\sigma^2)$. The effective DoF of 
function $\hat{f}$ is given by:
\begin{align*}
    \text{df}(\hat{f}) = \frac{1}{\sigma^2}\sum_{i=1}^n \text{Cov}(y_i, \hat{y}_i).
\end{align*} \color{black} Following the approach in \cite{mentch2020randomization}, we estimate this using Monte Carlo simulation.\color{black}

\subsubsection{Case Study: TRIM vs. SFS}\label{trim_sfs_simulation.section}
We extend the simulation used in \cite{mentch2020randomization} (Section 4) and generate synthetic data using the following setup. We generate data matrix $X \in \mathbb{R}^{n \times 5}$ where $n \in \{200,1000\}$ and each entry of $X$ is sampled from a uniform $U(0,1)$ distribution. We generate response $Y$ via the following models:

\begin{equation}\label{mars.display}
    Y = 10\sin(\pi X_1 X_2) + 20(X_3 - 0.05)^2 + 10X_4 +5 X_5 + \epsilon,
\end{equation}
and
\begin{equation}\label{mars_add.display}
    Y = 0.1 e^{4 X_1} + \frac{4}{1+ e^{-20(X_2 - 0.5)}} + 3 X_3 + 2 X_4 + X_5 + \epsilon.
\end{equation}

As in \cite{mentch2020randomization}, we refer to (\ref{mars.display}) as the MARS model and (\ref{mars_add.display}) as the MARSadd model in reference to their first appearance in \cite{friedman1991multivariate}. The error term $\epsilon$ follows an independent Gaussian $N(0,\sigma^2)$ distribution and $\sigma^2$ is prespecified to determine the signal-to-noise ratio of the generated data. Recall that SNR is defined by:

\begin{equation}
    \text{SNR} = \frac{\text{Var}\bigl(f(X)\bigr)}{\text{Var}(\epsilon)}.
\end{equation}

 We first generate synthetic datasets of size $n \in \{200,1000\}$ using the MARS and MARSadd functions and vary the SNR of the dataset on 10 evenly log-spaced points from 0.042 to 6. To replicate the results from \cite{mentch2020randomization}, we split the data into a 50\%-50\% train-test split and fit 2 models: a random forest with $mtry = 0.33$ and a bagging ensemble with $mtry = 1.0$. For both models, we grow 100 trees and set $maxnodes = 200$\footnote{We set the maximum number of leaf nodes to 200 so that we are consistent with \cite{mentch2020randomization}.}. We repeat this procedure for 500 trials across each SNR level to obtain estimates of the test error and effective DoF for each model. Finally, we repeat this entire procedure while fitting another bagging ensemble of 100 trees with TRIM and $maxnodes$ set such that the effective DoF of the TRIM ensemble matches the effective DoF of the SFS random forest. We discuss the results of this simulation below.

\begin{figure}[h]
\centering
\begin{minipage}{.5\textwidth}
  \centering
  \includegraphics[width=.975\linewidth]{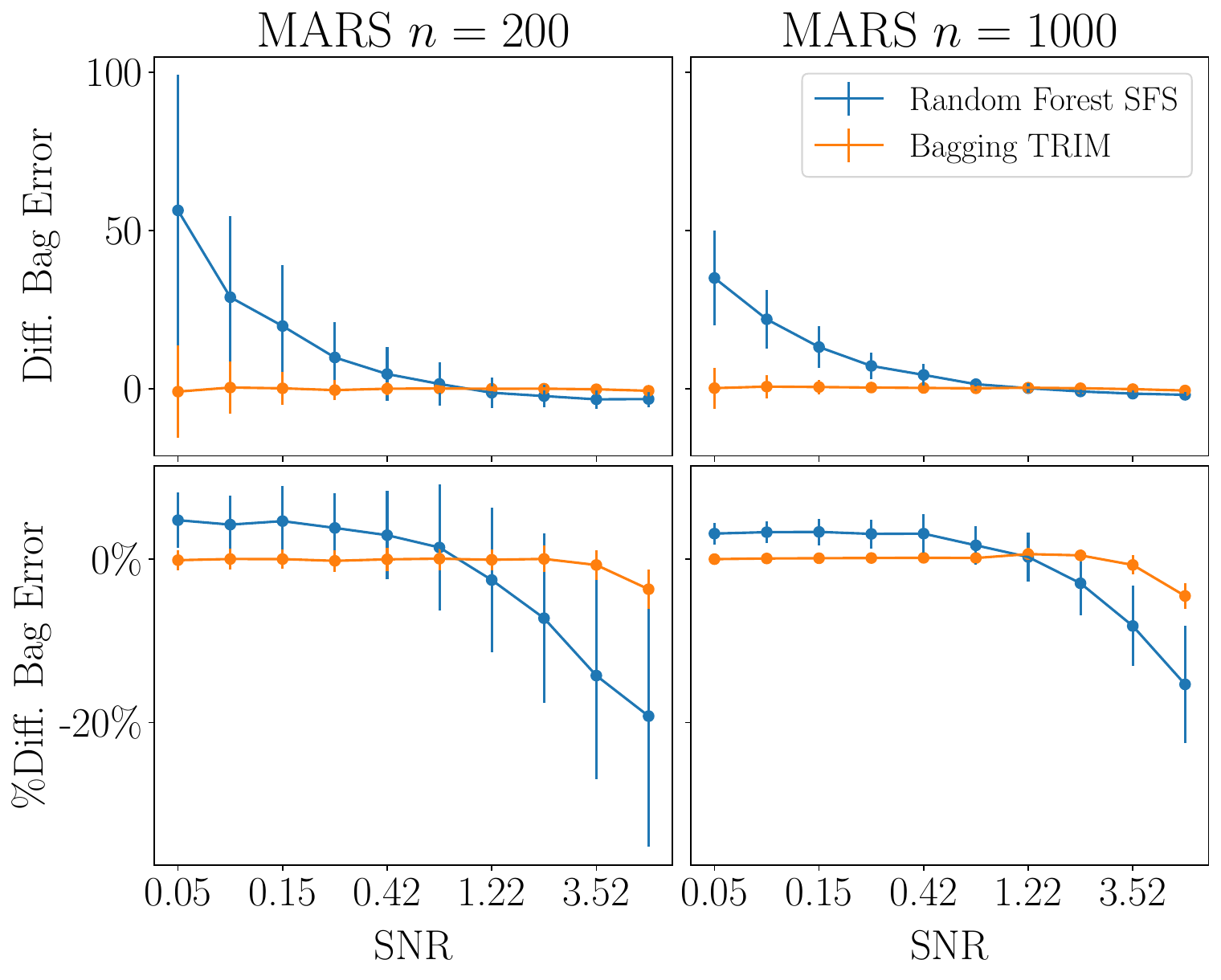}
  \label{fig:test1}
\end{minipage}%
\begin{minipage}{.5\textwidth}
  \centering
  \includegraphics[width=.975\linewidth]{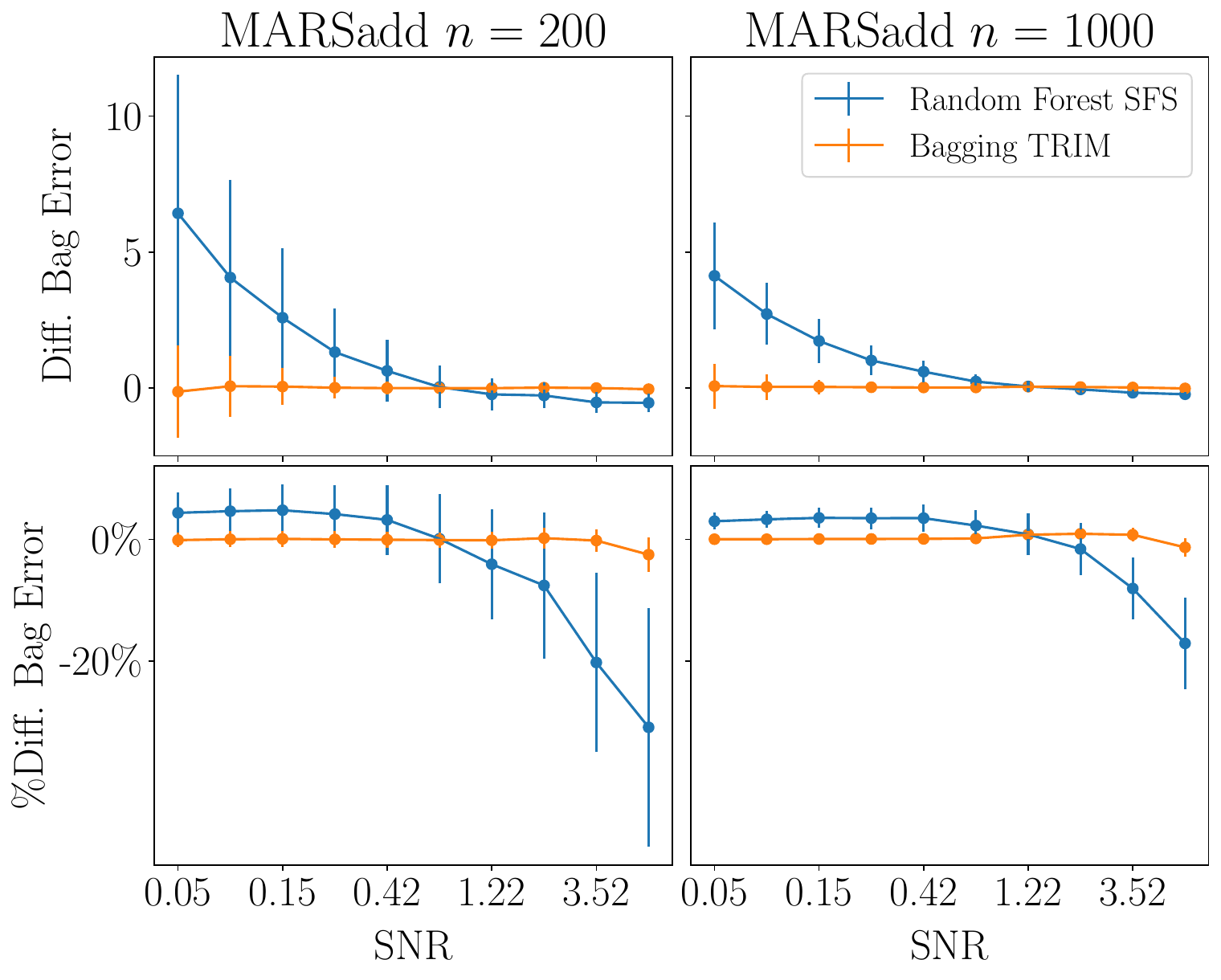}
  \label{fig:test2}
\end{minipage}
\caption{Even after tuning TRIM to match the effective DoF of SFS, TRIM does not improve ensemble performance at any SNR level.}
\label{TRIM_vs_SFS_SNR.fig}
\end{figure}

The plots in Figure \ref{TRIM_vs_SFS_SNR.fig} summarize the results of our simulation study. In the top row of plots, we report the performance of random forest SFS and bagging TRIM relative to vanilla bagging. The vertical axis shows the difference between the errors of the bagging ensemble and the errors of the SFS/TRIM ensemble and the horizontal axes show SNR. The SFS error curves (blue lines) are similar to the ones found in \cite{mentch2020randomization} (Section 4, Figure 4) and show that the relative performance of SFS random forests increases as SNR decreases. The TRIM error curves (orange lines) show that TRIM does not improve out-of-sample performance relative to bagging. We also show the percent difference in error between bagging and SFS/TRIM in the bottom row of plots, since we expect raw errors to be higher when SNR is low. The vertical axis shows:
\[\frac{\text{Error(Bagging)} - \text{Error(SFS/TRIM)}}{\text{Error(bagging)}} \times 100\%, \] and the horizontal axes again show SNR. Our conclusions remain the same, the relative performance of SFS over bagging increases inversely with SNR but the relative performance of TRIM compared to bagging remains the same across all SNR values.

Recall that the TRIM bagging ensembles were tuned to match the effective degrees of freedom (DoF) of the SFS random forests. While both TRIM and SFS reduce effective DoF, only SFS improves ensemble performance in this example. This suggests that factors beyond DoF reduction and variance reduction contribute to the effectiveness of SFS in random forests.

\color{black}
We find this result particularly interesting, as it aligns with an observation first made by \cite{breiman2001random}. Breiman noted that alternative forms of randomization—such as randomizing split selection\footnote{Randomly selecting splits instead of randomizing the features considered per split} or adding random noise to the outputs of individual decision trees—can improve predictive performance to some extent. However, none of these methods match the performance gains achieved through SFS randomization over the features considered at each split (see Section 3 of Breiman, 2001). This indicates that there is something distinctive about SFS randomization, which motivates our investigation into its role in bias reduction.
\color{black}

\subsubsection{TRIM vs. SFS Additional Discussions} \color{black}
As an aside, in \S A.1 of the appendix we report the tuned $maxnodes$ values for our TRIM ensemble. It is interesting to note that for medium to high SNRs, the tuned TRIM ensemble has much fewer leaf nodes compared to the bagging ensemble, even though the predictive performances of the models are similar. This is consistent with the understanding that tuning tree depth seldom affects predictive performance when SNR is high \citep{hastie2009elements}. 

We note that in other cases, trimming tree depth in random forests has been shown to improve predictive performance through regularization when  SNR is low \citep{zhou2023trees}. Various methods have been proposed to reduce the effective degrees of freedom of random forests and bagging ensembles \citep{mentch2022getting,mentch2020randomization, zhou2023trees}, and these regularization techniques yield different impacts on predictive performance. For example, \cite{mentch2020randomization} proposes augmented bagging, a technique to regularize random forests by adding noise features to the training data. We show in \S A.2 of the appendix that similar to TRIM, augmented bagging can reduce the effective DoF of the model without improving performance in the examples discussed above.
\color{black}

Also, as a quick detour, we briefly switch to the 
classification setting to visually 
highlight the difference between TRIM and 
SFS. On a 3-dimensional space 
$(X_1,X_2,X_3)$, consider a sphere of 
radius 1 with center at $(0.5,0.5,0.5)$. 
We uniformly generate $n = 10000$ data 
points and assign points inside of the 
sphere to positive class $Y = 1$ with 
probability $p = 0.9$. We assign points 
outside of the sphere to positive class $y 
= 1$ with probability $p = 0.1$; positive 
class points outside of the sphere and 
negative class $Y = 0$ points inside of 
the sphere can be viewed as noise points. 
Our goal is to test how well bagging 
ensembles, TRIM bagging ensembles, and SFS 
random forests recover the decision 
boundary of the sphere.
\begin{figure}[h]
    \centering
    \includegraphics[width = \textwidth]{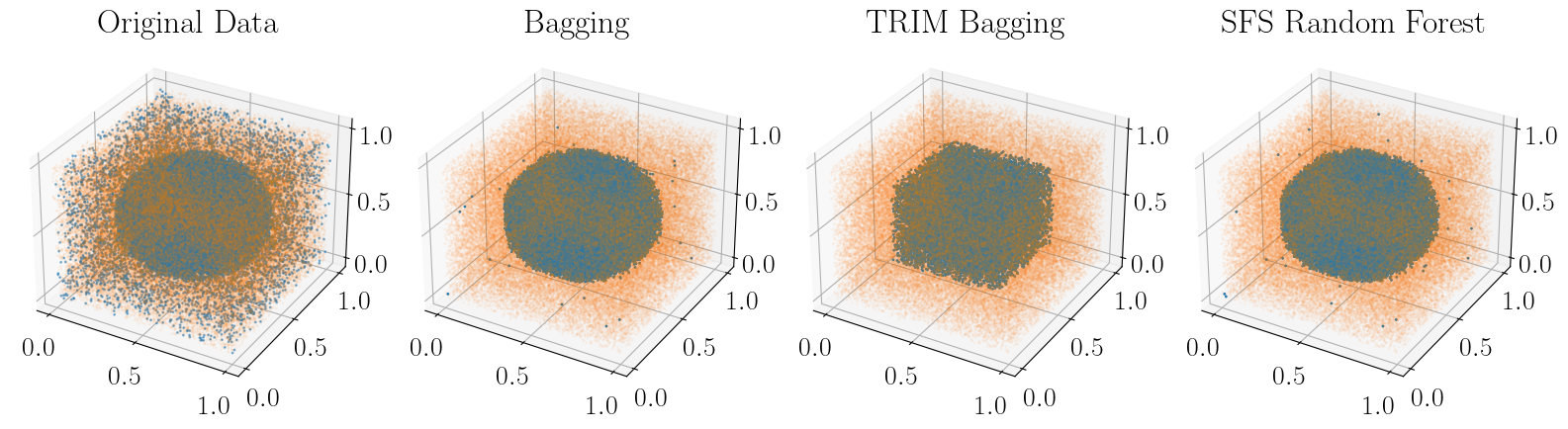}
    \caption{Both TRIM and SFS reduce the effective DoF of the model, however, their effects on the decision boundary of the ensemble differ drastically.}
    \label{3d_classification_plot.fig}
\end{figure}

The leftmost plot in Figure \ref{3d_classification_plot.fig} shows the actual data. The blue points represent $y=1$ and the orange points represent $y = 0$. The 3 right plots in the figure show the predictions of the bagging, TRIM, and SFS random forest ensembles, with $\hat{y} = 1$ in blue and $\hat{y} = 0$ in orange. We observe that the bagging ensemble and SFS random forest recover the true spherical decision boundary of the underlying data. The decision boundary of the TRIM ensemble, however, has degraded into a cube. While TRIM and SFS both reduce effective DoF, they impact the model very differently. Here, TRIM appears to significantly reduce the ability for the ensemble to fit the original training data, while SFS does not. In the following sections, we return to the regression setting to investigate how SFS randomization can in fact \emph{improve} the ability for an ensemble to fit the training data, which leads to a reduction in model bias.

\section{Random Forests and Hidden Patterns}

In the next two sections, we present the main investigations of our paper and show that random forests can reduce both bias and variance on datasets with hidden patterns. We start by discussing an extended case study in two dimensions to visualize how random forests can better fit the training data compared to bagging ensembles, in certain situations. Using this example, we challenge the existing statistical lore that random forests \emph{only} reduce variance compared to bagging. We then characterize our notion of hidden patterns and empirically show that on datasets with such patterns random forests reduce both bias and variance.

\subsection{Case Study: A Hidden Pattern in 2D}

\begin{figure}[h]
    \centering
    \includegraphics[width=\linewidth]{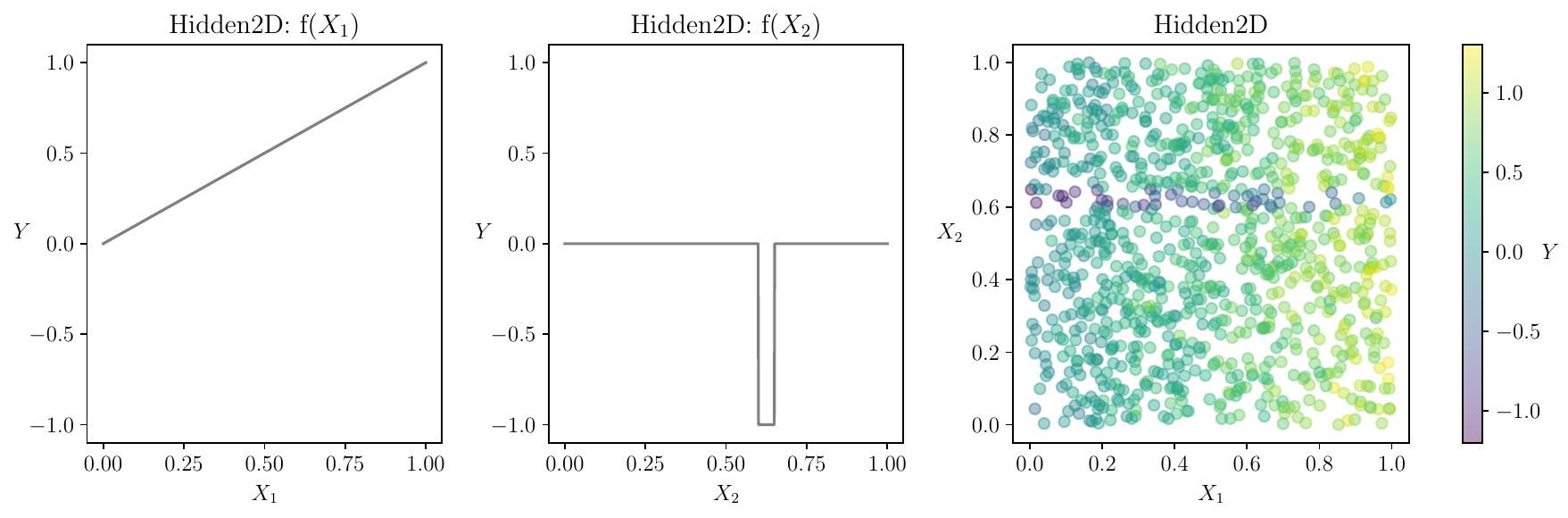}
    \caption{The left and middle display show $f_1(X_1)$ and $f_2(X_2)$, and the rightmost display shows an example of the Hidden2D model with 1000 observations and a SNR of 6.}
    \label{Hidden2D_DGP.fig}
\end{figure}

We present an easy-to-visualize example in two dimensions of how random forests can uncover patterns in the data missed by bagging. Consider the model below, which we refer to as \textbf{Hidden2D}, with the data generating procedure (DGP):
\begin{equation} \label{Hidden2D}
    Y = X_1 -  \mathbbm{1}(0.6 \leq X_2 \leq 0.65) + \epsilon.
\end{equation}
We say that Hidden2D is generated from function $f(X) = f_1(X_1) + f_2(X_2)$,  where $f_1(X_1) = X_1$ and $f_2(X_2) = -  \mathbbm{1}(0.6 \leq X_2 \leq 0.65)$. The DGP function $f(X)$ operates on the data matrix $X$ while the DGP functions $f_1(X_1)$ and $f_2(X_2)$ operate on individual columns of $X$. We show $f_1(X_1)$ and $f_2(X_2)$ in the two leftmost displays in Figure \ref{Hidden2D_DGP.fig}.
Following the procedure from the simulation studies presented in \cite{mentch2020randomization} (Sections 3 and 4), we generate the entries of data matrix $X \in \mathbb{R}^{n \times 2}$ from a uniform $U(0,1)$ distribution, and in the rightmost scatter plot in Figure \ref{Hidden2D_DGP.fig} we show an example of the Hidden2D model with $n = 1000$ observations and $\epsilon$, the independent Gaussian noise vector, set such that the SNR of the model is equal to 6. Each point in the scatter plot represents a observation and the color of the point indicates its corresponding $Y$ value.

Since the entries of $X$ are generated from a $U(0,1)$ distribution, the DGP function on feature $X_2$, $\mathbbm{1}(0.6 \leq X_2 \leq 0.65)$, only affects around $5\%$ of the observations in the data. This is shown in the rightmost plot of Figure \ref{Hidden2D_DGP.fig} by the thin horizontal band in dark blue, located right above $X_2 = 0.6$. As we show in the section below, decision trees often have trouble capturing this relationship between feature $X_2$ and $Y$.

\subsubsection{Visualization: Hidden2D Single Decision Tree}\label{hidden2d_single_viz.section}

\color{black}

\begin{figure}[h]
    \centering
\includegraphics[width=\linewidth]{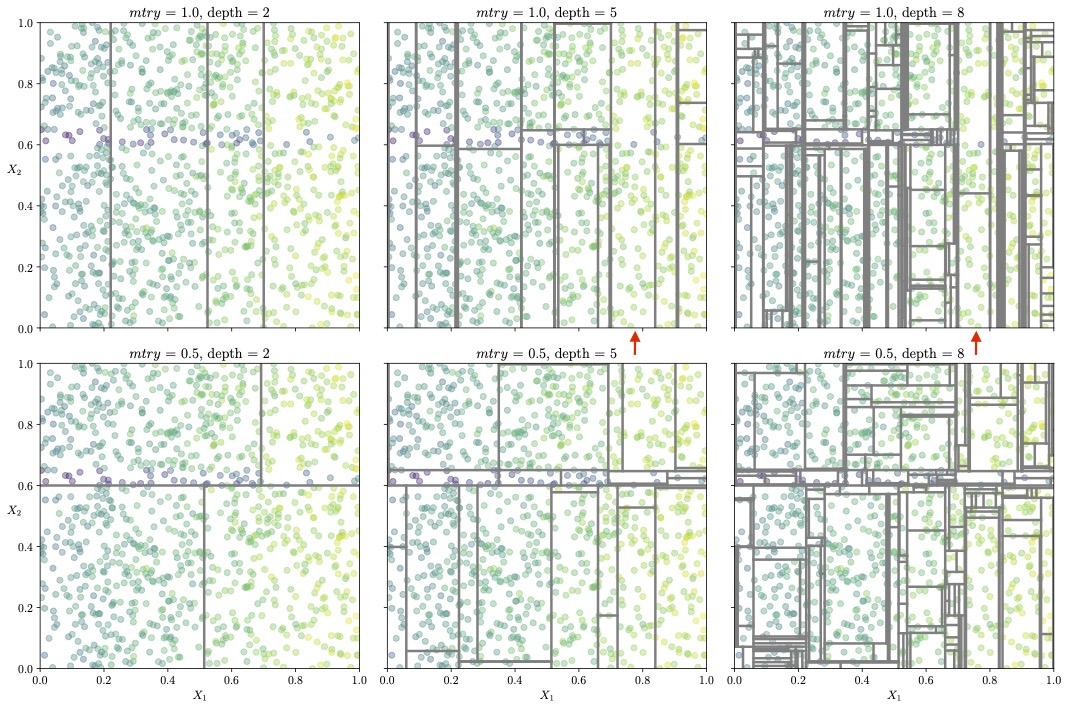}
    \caption{Splits of a decision tree fit on the Hidden2D example at various depths. The rectangular regions delineates the boundaries of the nodes of the tree at each depth level level. The top row of plots shows a tree with $mtry = 1.0$, the bottom row shows a tree with $mtry = 0.5$.}
    \label{Hidden2D_boundaries.fig}
\end{figure}

In the top row of plots in Figure \ref{Hidden2D_boundaries.fig}, we fit a \emph{single} decision tree on the Hidden2D example from above, with 1000 observations and a SNR of 6. The grey lines show the splits constructed by the tree, vertical lines are splits on feature $X_1$ and horizontal lines are splits on feature $X_2$. Each plot shows all of the splits in the tree up to a certain depth level, depth  $ \in \{2,5,8\}$. We observe that the decision tree tends to split on feature $X_1$ before $X_2$; the leftmost plot shows that the first 3 splits of the tree, of depths 1 and 2, are only on feature $X_1$. This makes sense since decision trees are constructed greedily, with the best split selected to minimize the error of the directly resulting partitions. The DGP function on feature $X_1$, $f_1(X_1) = X_1$, affects all of the observations while the DGP function on $X_2$ affects only a small subset ($\sim$5\%) of observations, so initially splitting on feature $X_1$ reduces error by a greater extent than splitting on $X_2$.

Importantly, we note that the subsequent splits on feature $X_2$ are constructed with respect to these initial splits on $X_1$, due to the hierarchal nature of decision trees. Here, we observe that it is difficult for the horizontal splits on $X_2$ to capture the underlying DGP function, $f_2(X_2) = - \mathbbm{1}(0.6 \leq X_2 \leq 0.65)$, when conditioned on the vertical splits on $X_1$. The red arrows in Figure \ref{Hidden2D_boundaries.fig} show the region where this is most apparent. Consider the middle display in the top row of Figure \ref{Hidden2D_boundaries.fig}, this plot shows the splits of our decision tree at depth 5. The red arrow in this display shows the region delineated by the splits $0.70 \leq X_1 \leq  0.81$, and, within
this region, it is hard to construct a horizontal split on $X_2$ that captures  DGP function $f_2(X_2)$. As we can see from the rightmost display in the top row of Figure \ref{Hidden2D_boundaries.fig}, our decision tree still fails to capture this pattern by depth level 8.

In this example, the relationship between feature $X_2$ and $Y$ is hidden by $X_1$. We refer to this relationship as a hidden pattern, a notion we formally define in \S\ref{hidden_pattern_def.section}. For now, consider the bottom row of plots in Figure \ref{Hidden2D_boundaries.fig}. These plots show a single decision tree fit on the same example from above, however, each split is constructed on a \emph{randomly} selected feature, i.e., $mtry$ is set to 0.5 for this decision tree. We can see from the leftmost display that this leads to earlier splits on feature $X_2$. As a result, this tree with $mtry = 0.5$ can better capture the DGP function, $f_2(X_2) = - \mathbbm{1}(0.6 \leq X_2 \leq 0.65)$, compared to the $mtry = 1.0$ tree shown in the row above. Below, we present a simulation study to show how the effect discussed in this section extends to bagging ensembles and random forests.

\color{black}

\subsubsection{Hidden2D Simulation Study: Bagging and SFS Random Forests} \label{hidden2d_first_study.section}

Using the Hidden2D model \eqref{Hidden2D}, we again generate $n = 1000$ data points while setting $\epsilon$  such that the SNR of the data is equal to $6$. Following the setup presented in \S\ref{trim_sfs_simulation.section}, we split the data via a 50-50 train/test split and construct two ensembles of 500 \textbf{full-depth} trees on the train split: a bagging ensemble and a SFS random forest with $mtry = 0.5$. In these full-depth tree ensembles, each tree is grown so that each leaf node contains just a single observation. We compare the test performances of the two ensembles. The bagging ensemble has a test MSE of 0.047 and, interestingly, the SFS random forest has a test MSE of 0.029, a 37\% decrease in test error.

We find it unlikely that this observation can be fully explained by the variance reduction that SFS random forests achieve over bagging. Since random forests reduce effective DoF relative to bagging, we would expect them to \emph{increase} test error when SNR is high, if the advantage of random forests over bagging were solely due to variance reduction. Rather, we believe that this observed decrease test error in an high SNR setting can be explained by the biases of the models. In this example, the \emph{training} MSE of the SFS random forest is reduced to 0.005 from 0.007 for bagging, a 28.6\% decrease, which suggests a reduction in bias.

\begin{figure}[h]
    \centering
    \includegraphics[width = 0.9\textwidth]{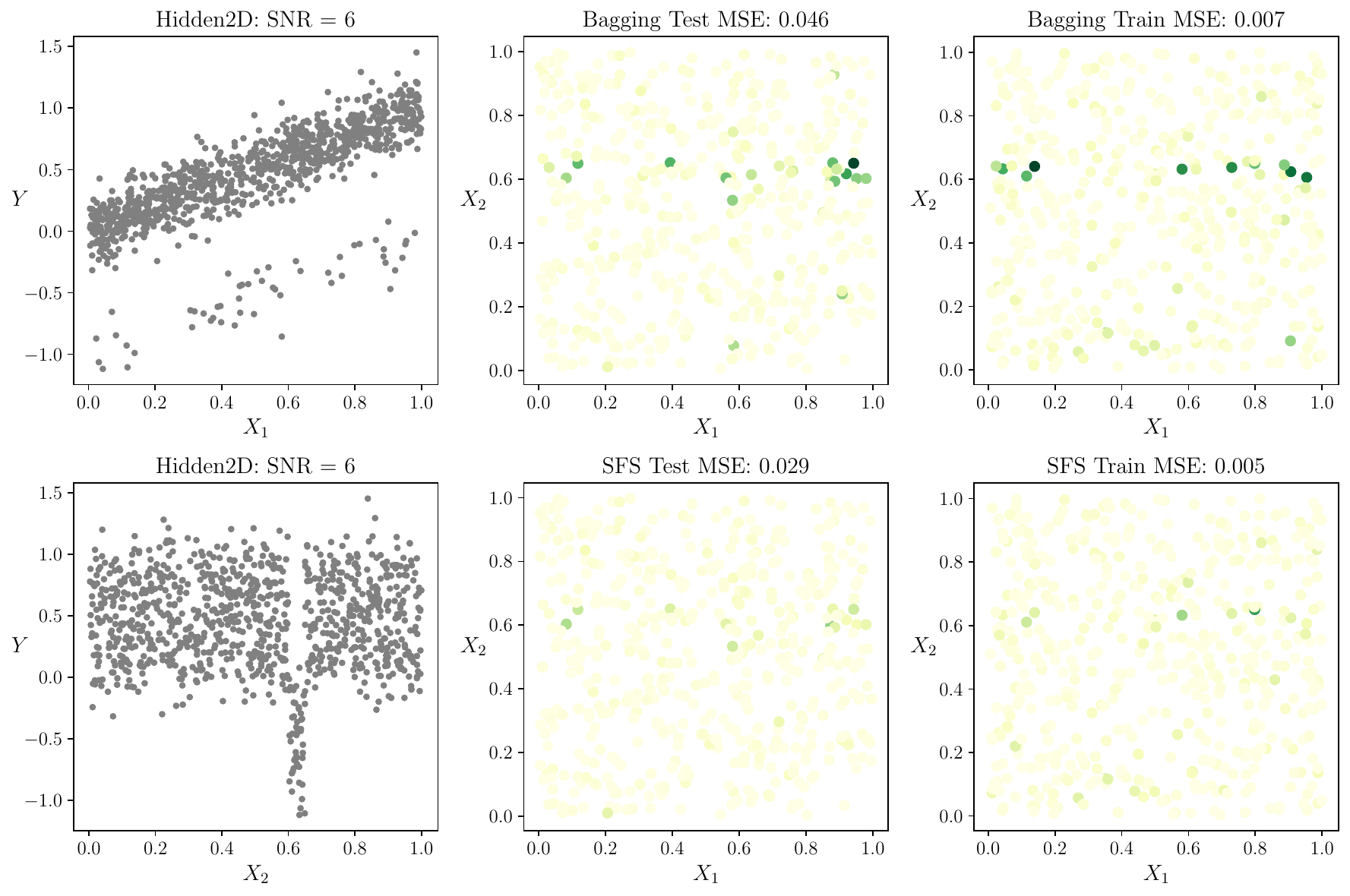}
    \caption{SFS random forests and bagging ensembles fit on Hidden2D generated data with high SNR. SFS allows random forests to better capture the effect of the hidden patterns.}
    \label{hidden2d_high_snr.fig}
\end{figure}

To understand how random forests can reduce bias in this example, we visualize the distribution of errors for each point in the training set and test set. In the center column of Figure \ref{hidden2d_high_snr.fig}, we show scatter plots of the test data points with $X_1$ on the horizontal axes and $X_2$ on the vertical axes. The color gradient shows the squared test error for each data point, darker colors indicate higher errors. The top plot shows the error distribution of the bagging ensemble and the bottom plot shows the error distribution of the SFS random forest and the color gradient is normalized across all plots. The right column in Figure \ref{hidden2d_high_snr.fig} shows these same plots but for the training data and training error.

We observe from these plots that for bagging ensembles (top two plots) data points in the $0.6 \leq X_2 \leq 0.65$ horizontal band have higher training and test errors. This band corresponds to the region of $X_2$ where the DGP function $f_2(X_2) = -\mathbbm{1}(0.6 \leq X_2 \leq 0.65)$ is greater than zero. We also see that in this region, SFS (bottom two plots) reduces both training and test error relative to bagging. We see from this example that the bagging ensemble is unable to capture the hidden pattern in feature $X_2$. The randomization from SFS better allows random forests to capture this effect, which can reduce the bias of the ensemble.

\begin{figure}[h]
\centering
\begin{minipage}{0.5\textwidth}
  \centering
  \includegraphics[width=1\linewidth]{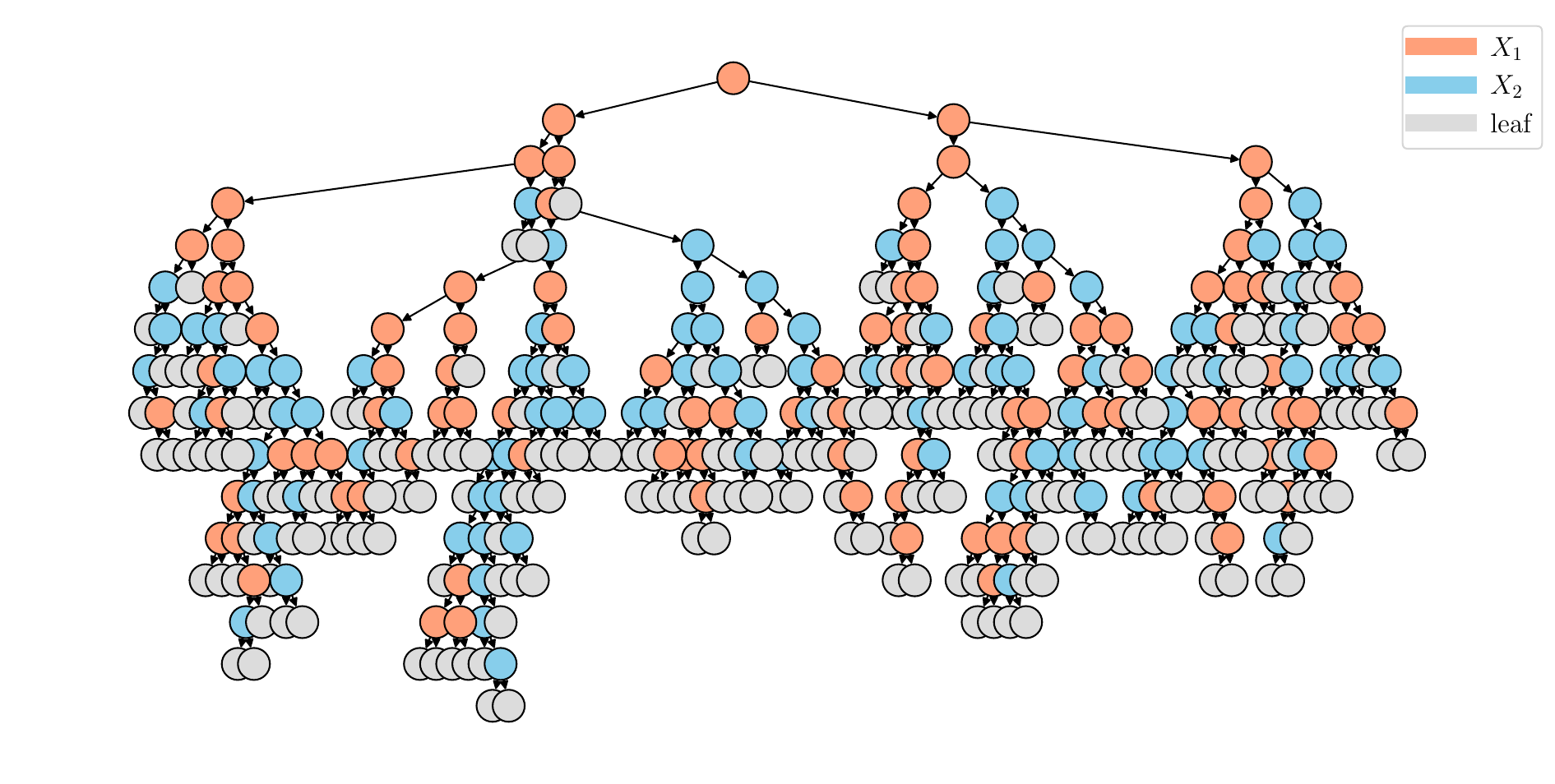}
  \label{2d_bag_tree.fig}
\end{minipage}%
\begin{minipage}{0.5\textwidth}
  \centering
  \includegraphics[width=1\linewidth]{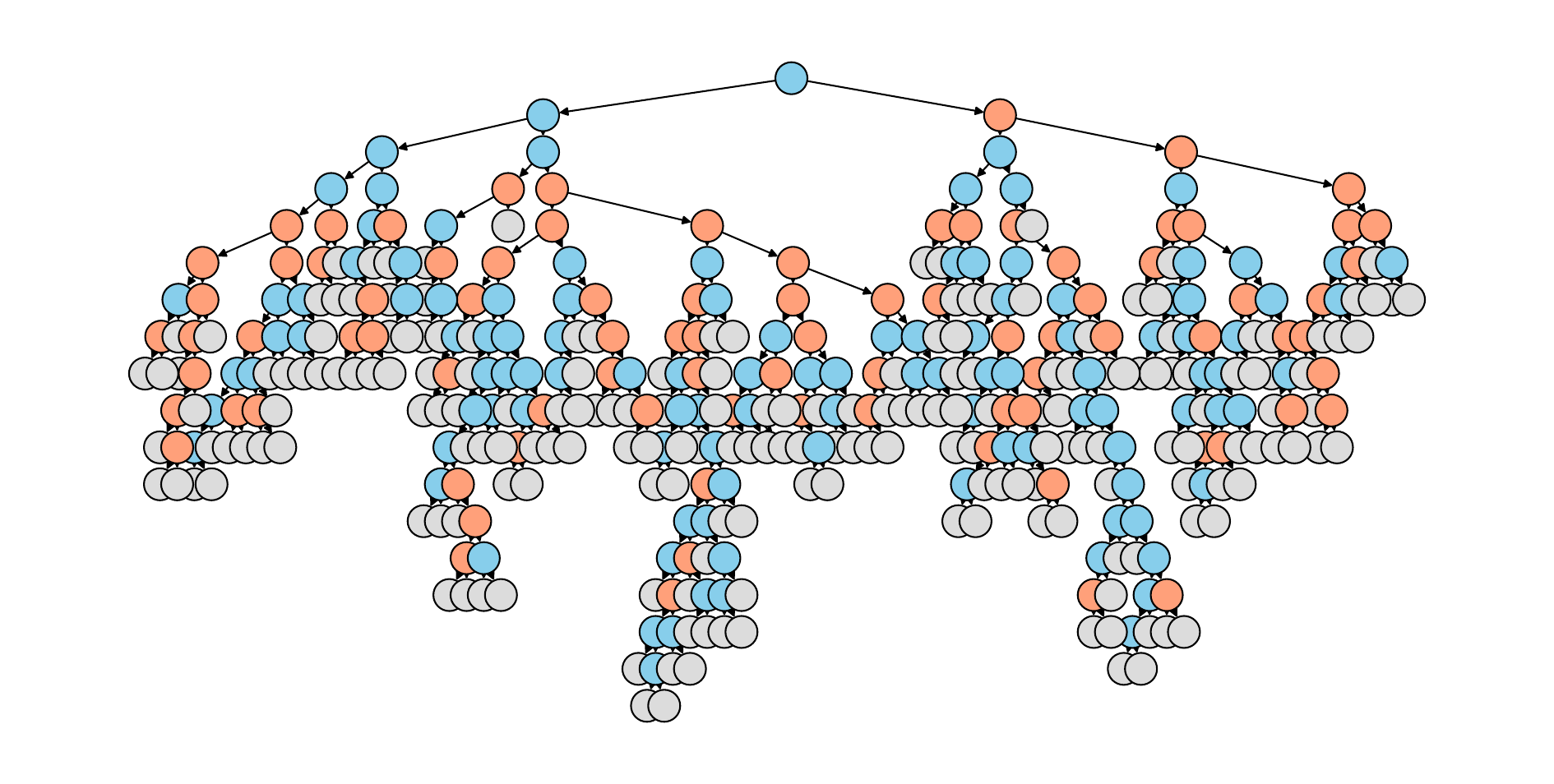}
  \label{2d_sfs_tree.fig}
\end{minipage}
\caption{Visualization of a tree used in bagging ensemble (left) and SFS random forest (right). Feature $X_1$ dominates the top layers of the bagging ensemble tree. The color of each node shows the feature used in the split, orange for feature $X_1$ and blue for feature $X_2$. The grey nodes are leaves. }
\label{hidden2d_tree.fig}
\end{figure}

Bagging ensembles have trouble capturing the hidden pattern in $X_2$ due to the greedy nature of how decision trees are fit. In the full-depth decision trees of the bagging ensemble, feature $X_1$ typically dominates the top layers of the tree, as we see in Figure \ref{hidden2d_tree.fig}. As we discuss in the sections above, in the Hidden2D example it is difficult for the subsequent splits on feature $X_2$ to capture the DGP function $f_2(X_2)$, when conditioned on the preceding splits on feature $X_1$. In SFS random forests, on the other hand, we force half of the splits in each tree to be on $X_2$. In doing so we capture the effect of the hidden pattern.

\subsubsection{Hidden2D Simulation Study: Varying SNR}

\begin{figure}[h]
    \centering
    \includegraphics[width = .95\textwidth]{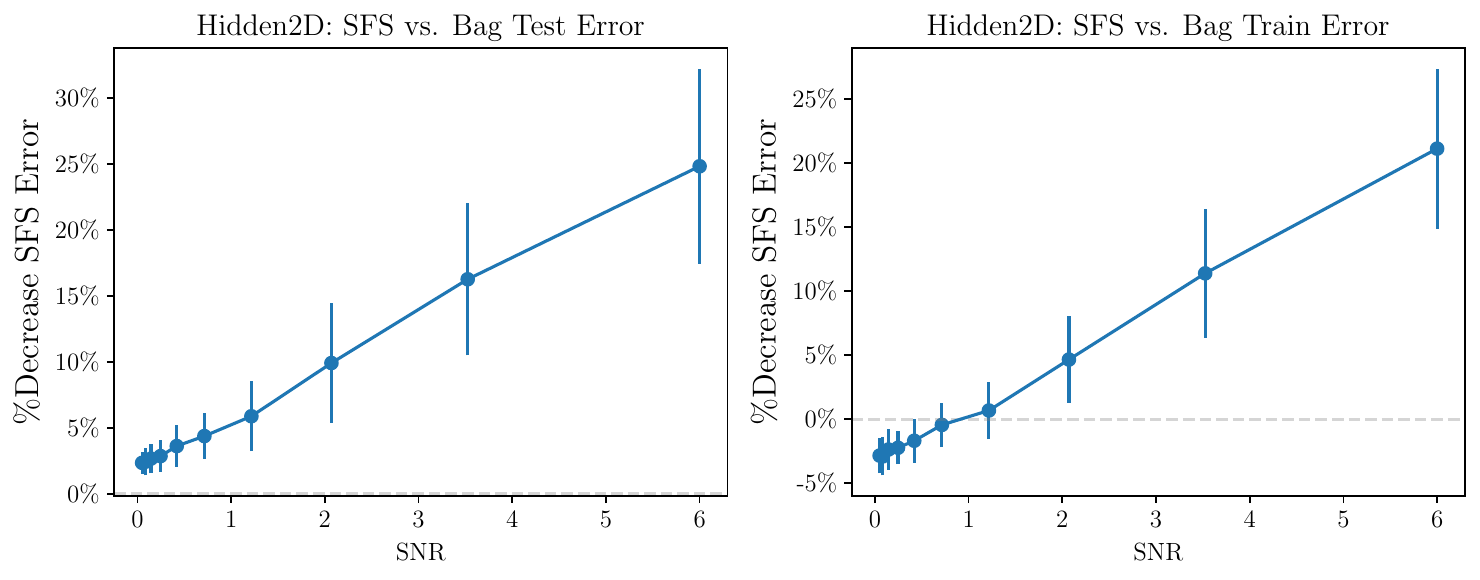}
    \caption{As SNR increases in this example, we observe that SFS increasingly outperforms vanilla bagging. SFS can capture the hidden pattern in feature $X_2$, which is more prominent in higher SNR settings.}
    \label{hidden2d_bag_v_SFS.fig}
\end{figure}

For thoroughness, we evaluate how SFS random forests perform compared to bagging on the Hidden2D model across different SNR levels. We follow the experimental procedure from \S\ref{trim_sfs_simulation.section} and show the results in Figure \ref{hidden2d_bag_v_SFS.fig}. The left plot compares the test performance of SFS against bagging ensembles. The horizontal axis shows SNR and the vertical axis shows the percent decrease in test MSE between SFS and bagging. We observe that SFS outperforms bagging across all SNR ranges in this example; the percent decreases in test MSE are all positive. We see that as SNR increases SFS outperforms bagging by a larger degree.

The right plot in Figure \ref{hidden2d_bag_v_SFS.fig} shows the percent decrease in \emph{training} error between SFS random forests and bagging ensembles. We observe a positive trend between the percent decrease in training error and SNR, and we note that in high SNR settings, random forests have lower training errors than bagging ensembles. This suggests in this example that random forests reduce bias compared to bagging when SNR is high. 

\begin{figure}[h]
    \centering
    \includegraphics[width = 0.9\textwidth]{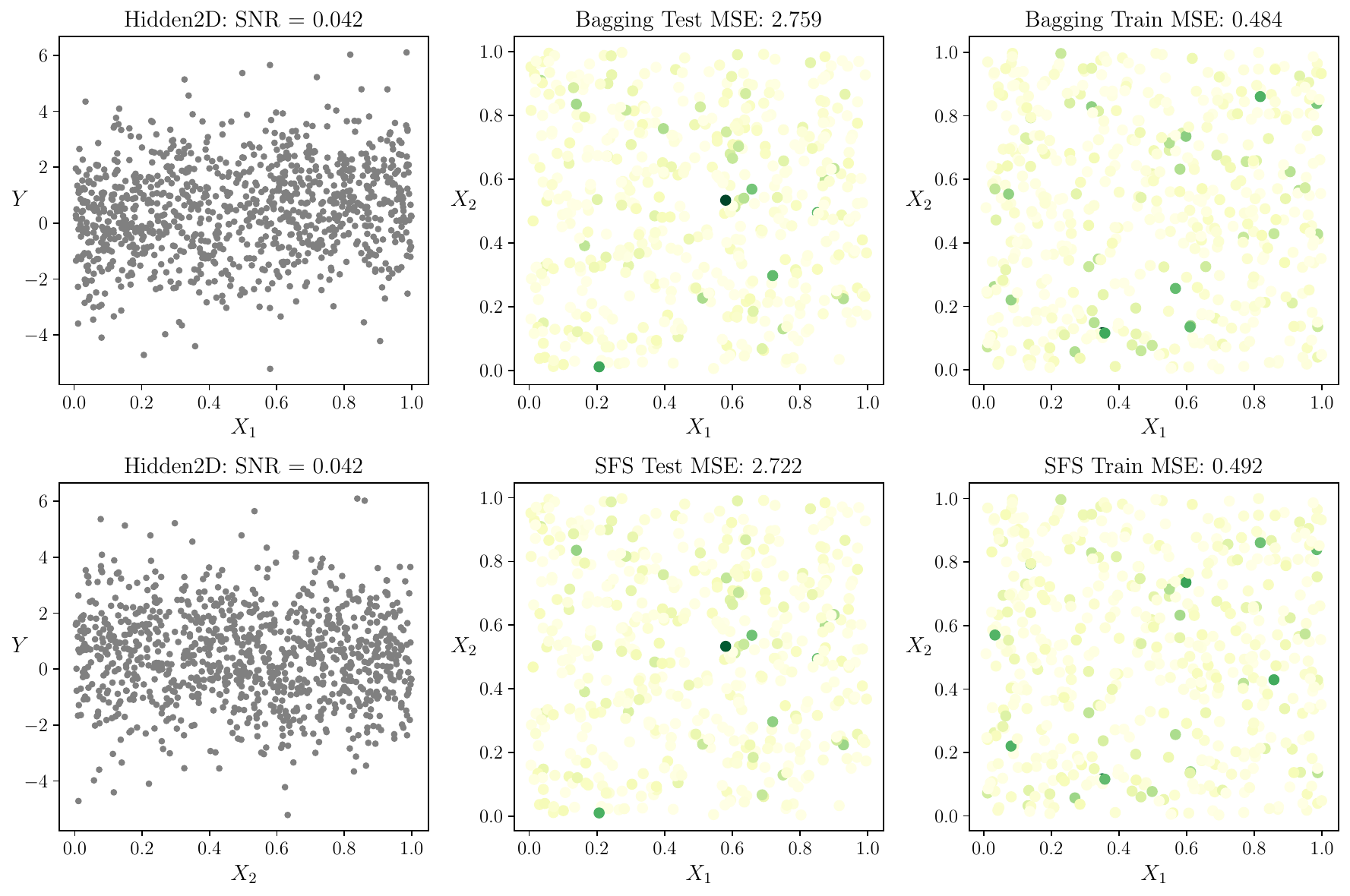}
    \caption{SFS random forests and bagging ensembles fit on Hidden2D generated data with low SNR. In low SNR settings, the effect from hidden pattern $X_2$ is obscured by noise.}
    \label{hidden2d_low_snr.fig}
\end{figure}

We also notice that in the low SNR $<1$ setting, SFS random forests increase training error relative to bagging; the percent decreases in training error here are negative. In this low SNR setting SFS still improves test performance over bagging, as shown in the left plot. This can be explained by the arguments in \cite{mentch2020randomization}, that randomization in random forests acts as regularization in low SNR settings by reducing effective DoF. In this example, however, the improvement of SFS random forests over bagging is more pronounced when SNR is high. We see in this example that the hidden pattern in $X_2$ is obscured by noise when SNR is low. The left column of Figure \ref{hidden2d_low_snr.fig} shows $Y$ plotted against $X_1$ and $X_2$ when SNR = 0.042, and we do not observe the hidden pattern that was apparent in Figure \ref{hidden2d_high_snr.fig}, when SNR = 6. We also see that the distributions of the training and test errors are no longer clustered around the $0.6 \leq X_2 \leq 0.65$ horizontal band. In this case, the hidden pattern in $X_2$ is irrelevant considering the noise in the data, and as such, the improvement of SFS random forests over bagging is  less pronounced.

Our findings build on the arguments in \cite{mentch2020randomization}. While randomization acts as a form of regularization in low SNR settings, SFS randomization can enable random forests to better fit the training data on certain datasets compared to bagging.  As a result, random forests can outperform bagging when SNR is high. In the next section, we empirically demonstrate that this improvement in the high SNR = 6 setting stems from bias reduction.


\color{black}
\subsubsection{Hidden2D Simulation Study: Bias Variance Decomposition} \label{Hidden2D_bvd.section}

In our simulation study above, the underlying data generating procedure is known, and, as such, we can compute the decomposition of the error of our model into a bias term, variance term, and noise term. We follow the general bias-variance decomposition procedure discussed in \cite{domingos2000unified}. Given regression data-generating procedure $Y = f(X) + \epsilon$, where $\epsilon$ is the irreducible error, and prediction model $\hat{f}(X)$ is fit on $X$ to predict $Y$,  the squared error of our model can be decomposed into:
\begin{equation}
    E \bigl[\bigl(Y - \hat{f}(X)\bigr)^2 \bigr] = \bigl(E\bigl[f(X) - \hat{f}(X)\bigr]\bigr)^2 + E\bigl[\bigl(\hat{f}(X) - E\bigl[\hat{f}(X)\bigr] \bigr)^2\bigr] + \text{Var}(\epsilon) 
\end{equation}
where the first term is squared bias, the second term is variance, and the third term is irreducible error. The expectations are taken with respect to randomness in the training data and in the prediction model, e.g., the bootstrap and $mtry$ randomization in the random forest algorithm. We repeat our Hidden2D simulation study (with SNR = 6) 500 times to estimate the bias and variance of both the bagging ensemble and SFS random forest.

\begin{table}[h]
\centering
\begin{tabular}{|c|c|c|}
\hline
               \textbf{Hidden2D}   & $\text{Bias}^2$ & Variance \\ \hline
SFS Random Forest & 0.00235         & 0.00493  \\ \hline
Bagging Ensemble  & 0.00525         & 0.00838  \\ \hline
\end{tabular}
\caption{Bias variance decomposition for Hidden2D simulation study with SNR = 6.}
\label{hidden2d_bvd.table}
\end{table}

We show the results of this bias variance decomposition in Table \ref{hidden2d_bvd.table} and we see that in the high SNR setting of our Hidden2D study, random forests reduce both bias and variance compared to bagging. It is important to note that this finding shows that random forests reduce bias in addition to variance, an observations that appears to have been overlooked in prior literature.


\subsection{Do Random Forests \emph{Only} Reduce Variance?}

In this section, we revisit the classical belief that random forests only reduce variance compared to bagging. Indeed, the second edition of \textit{Elements of Statistical Learning} states in Chapter 15, Section 4.2 that: "improvements in prediction obtained by bagging or random forests are solely a result of variance reduction." The authors argue that this is due to the fact that "the bias of bagged trees is the
same as that of the individual (bootstrap) trees" and that randomization is unlikely to reduce the bias of an individual tree grown to full depth \cite{hastie2009elements}.


\begin{figure}[h]
    \centering
    \includegraphics[width=0.85\linewidth]{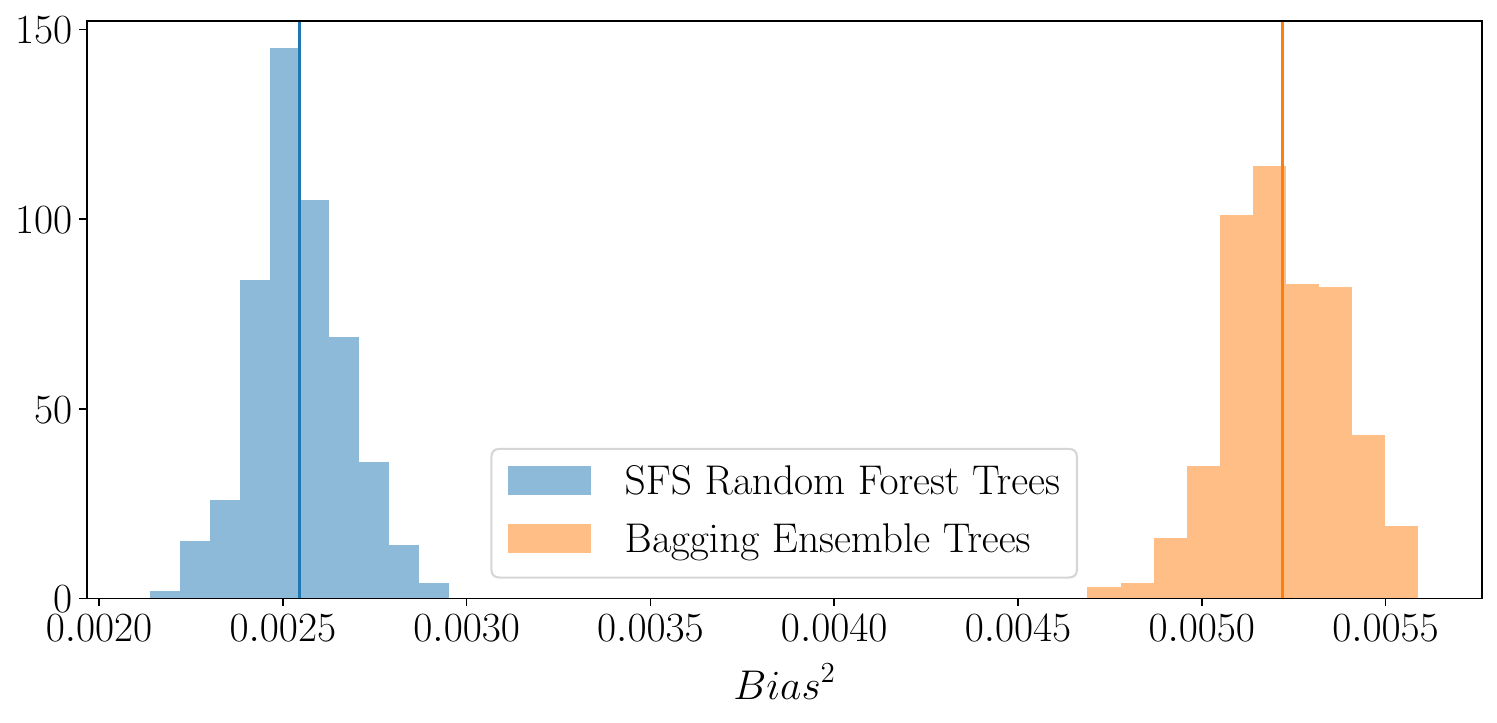}
    \caption{Distribution of the bias of individual trees in an SFS random forest and a bagging ensemble. Each tree is constructed off of a bootstrap sample of the data. SFS random forest trees are constructed with $mtry = 0.5$ for each split, bagging ensemble trees are constructed with $mtry = 1.0$ for each split.}
    \label{individual_tree_bias.fig}
\end{figure}

While randomizing splits can increase tree bias in certain cases, our Hidden2D example suggests that it can also reduce bias in others. In our example, SFS randomization \emph{lowers} the bias of full-depth decision trees. We repeat the bias variance decomposition discussed in \S\ref{Hidden2D_bvd.section} and estimate the bias of each individual tree in both the SFS random forest and the bagging ensemble. The distribution of these biases across trees are shown in Figure \ref{individual_tree_bias.fig}. From this plot, it is apparent that the bias of trees in the random forest is significantly lower than that of trees in the bagging ensemble. Note that the mean of each distribution in the plot, indicated by the vertical line, is equal to the squared bias of the corresponding ensemble as reported in Table \ref{hidden2d_bvd.table}.

Recall that in \S\ref{hidden2d_single_viz.section}, we hint that SFS randomization allows a single decision tree to better capture DGP function $f_2(X_2)$ in the Hidden2D model. In this section, we formally show that SFS randomization reduces the bias of individual trees in a random forests, which reduces model bias compared to bagging.


\subsection{An Analysis of How Random Forests Reduce Training Error}
 Here, we more formally analyze the mechanism behind how random forests reduce training error compared to bagging. Consider a random forest/bagging ensemble that consists of a collection of decision trees, $T$. Each decision tree is grown to full depth on a bootstrap sample of the original data, which means that each leaf node of the tree contains just a single observation\footnote{We reiterate here that trees in a random forest are fit on bootstrap samples of the data and not on the original data. Each decision tree perfectly fits the bootstrap sample but may have non-zero training error on the original data.}. We can express the prediction of the ensemble for a single observation $X^{(i)}$ by:\begin{equation} \label{rf_pred_local.eq}
    \hat{f}(X^{(i)}) = \frac{1}{|T|}\sum_{t \in T} L_t(X^{(i)}),
\end{equation}
where $L_t(X^{(i)})$ denotes the value of the leaf node in tree $t$ that observation $X^{(i)}$ is routed to. Since each decision tree is grown to full depth, the value of the leaf node corresponds to the response of the single observation in the bootstrap sample that the leaf node contains. Functions $\hat{f}(X^{(i)})$ and $L_t(X^{(i)})$  operate element-wise on a single row (observation) in data matrix $X$.

Let $\Psi_i$ represent the set of decision trees that are fit on bootstrap samples of the data that contain $X^{(i)}$:
\begin{align*}
    \Psi_i = \{ \text{Trees fit on bootstrap samples that \textbf{do} contain $X^{(i)}$} \},
\end{align*}
and let $\Omega_i$ represent the set of decision trees that are fit on bootstrap samples of the data that do not contain $X^{(i)}$:
\begin{align*}
    \Omega_i = \{ \text{Trees fit on bootstrap samples that \textbf{do not} contain $X^{(i)}$} \}.
\end{align*} We have that:
\begin{align*}
    |T| = |\Psi_i| + |\Omega_i|,
\end{align*}
and that $|\Psi_i| \approx 0.632|T|$. This follows from properties of the bootstrap when the number of bootstrap samples is large.

Tree $t \in \Psi_i$ is fit on a bootstrap sample that contains $X^{(i)}$ so $L_t(X^{(i)}) = Y_i$ since the observation is routed to its own leaf node. We can now decompose the prediction of the random forest/bagging ensemble for observation $X^{(i)}$ into:
\begin{equation}\label{local_pred_decomp.eq}
\hat{f}(X^{(i)}) = \frac{|\Psi_i|}{|T|} Y_i + \frac{1}{|T|} \sum_{t \in \Omega_i} L_t(X^{(i)}) \approx 0.632 \cdot Y_i + \frac{1}{|T|} \sum_{t \in \Omega_i} L_t(X^{(i)}),
\end{equation}
where the first term is the observed response $Y_i$ times a constant.

The second term of expression \eqref{local_pred_decomp.eq} is interesting. Consider $L_t(X^{(i)})$ for $t \in \Omega_i$. Tree $t$ is fit on a bootstrapped sample $X'$ of $X$ that \textbf{does not} contain observation $X^{(i)}$. As such, observation $X^{(i)}$ is routed to a leaf node in tree $t$ that contains just a single observation $X^{(v)} \in X'$ of the bootstrapped sample, and $L_t(X^{(i)}) = Y_v$. \newline

\noindent{\textbf{Voting Points:}} We refer to $X^{(v)}$ as a voting point for observation $X^{(i)}$, and, as an aside, we briefly discuss why we use this terminology. Voting points in random forests originate from \cite{lin2006random} where the authors analyze random forests as a potential nearest neighbor algorithm. The authors define that point $X_1$ is $k$-potential nearest neighbor ($k$-PNN) to point $X_2$ on dataset $X$ if there exists no other point in dataset $X$ other than $X_1$ that lies in the hyper-rectangle defined by $X_1$ and $X_2$. 
\begin{prop}
    Voting point $X^{(v)}$ is 1-PNN to observation $X^{(i)}$  in bootstrapped sample $X'$.
\end{prop} 
This proposition follows directly from Proposition 1 in \cite{lin2006random}. \cite{lin2006random} define voting points to be data points that are $k$-PNN to an observation of interest and that affect the prediction of that observation. Here, $X^{(v)}$ is 1-PNN to $X^{(i)}$ and it is apparent from display \eqref{local_pred_decomp.eq} that $X^{(v)}$ affects the prediction of $X^{(i)}$. 
\newline

\noindent We return to our training error analysis. Let the function:
\begin{equation}\label{mu_def.eq}
    \mu_{\Omega_i}(X^{(i)}) = \frac{1}{|\Omega_i|} \sum_{t \in \Omega_i} L_t(X^{(i)})
\end{equation} represent the average of all of the voting points for observation $X^{(i)}$. Note that when $|T|$ is large the number of voting points per observation is close to $0.368 |T|$; this again follows from properties of the bootstrap.  We explicitly characterize the training error of bagging ensembles/random forests in terms of this function in the result below.

\begin{prop} Given definitions discussed above, the training mean squared error of a random forest and a bagging ensemble can be given by:
\begin{equation}
   \frac{1}{n} \sum_{i=1}^n (Y_i - \hat{f}(X^{(i)}))^2 = \frac{1}{n} \biggl(\frac{|\Omega_i|}{|T|}\biggr)^2 \sum_{i=1}^n \biggl(Y_i -   \mu_{\Omega_i}(X^{(i)})\biggr)^2 \approx 0.135 \frac{1}{n} \sum_{i=1}^n \biggl(Y_i -   \mu_{\Omega_i}(X^{(i)})\biggr)^2 
\end{equation}
\end{prop}

We derive this expression in Section B.1 of the appendix. This result shows that for random forests/bagging ensembles, the in-sample error of the model depends on $\mu_{\Omega_i}(X^{(i)})$, i.e., the average response of the voting points for each observation of interest $X^{(i)}$. Below, we show how voting points differ in random forests compared to bagging in order to illustrate how random forests reduce training error. \newline

\subsubsection{ Voting Points of Random Forests vs. Bagging:} We return to our Hidden2D example and consider the simulation study discussed in \S\ref{hidden2d_first_study.section}. Consider the rightmost column of plots in Figure \ref{hidden2d_high_snr.fig}; SFS randomization reduces the training error of random forests compared to bagging. To illustrate the effect of voting points, we select the observation in the plot where the difference in training error between bagging and random forests is the largest. We can think of this as the observation where random forests improve the most over bagging; this is our observation of interest.

\begin{figure}[h]
    \centering
    \includegraphics[width=1.0\linewidth]{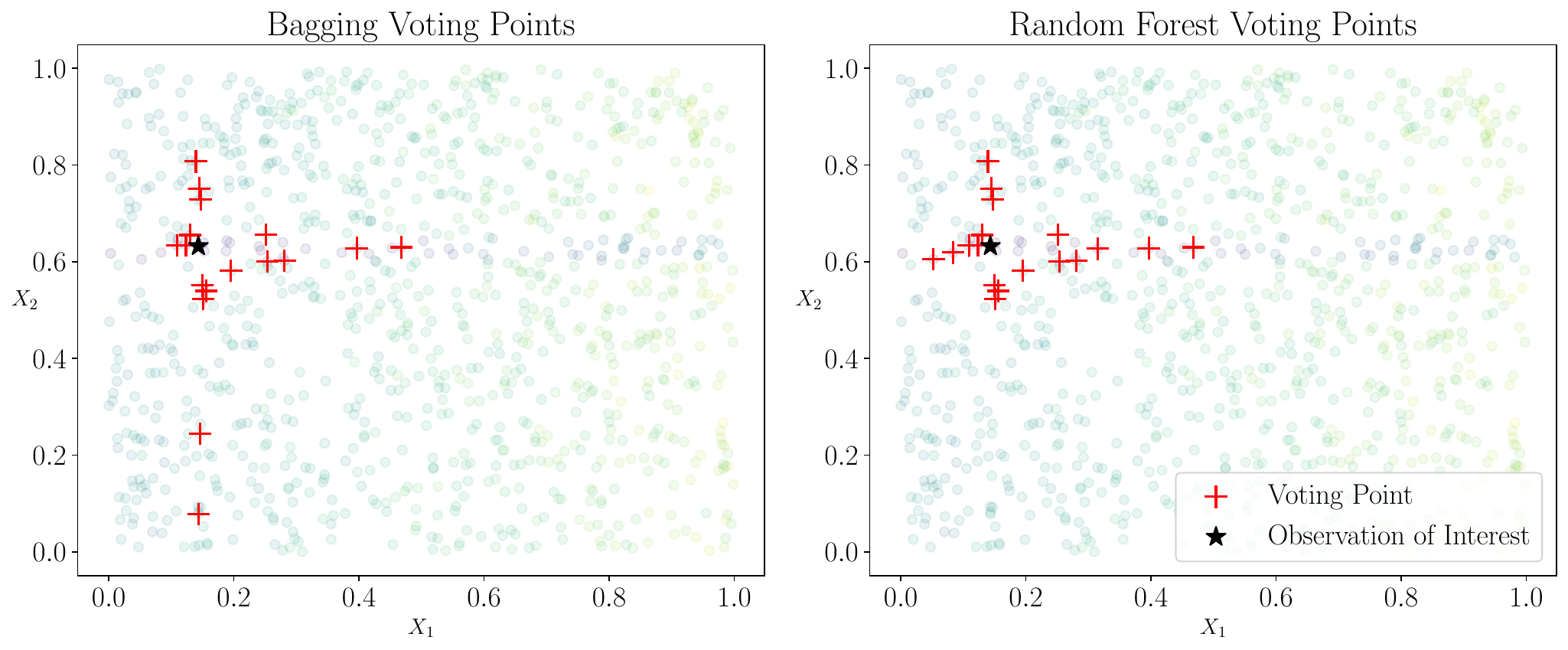}
    \caption{Voting points for our observation of interest in bagging and random forests. The prediction for our observation of interest is a weighted average of these marked points.}
\label{voting_points_viz.fig}
\end{figure}

In Figure \ref{voting_points_viz.fig}, we show the voting points of this observation in both bagging and random forests. Each scatter plot shows the training data, each point corresponds to an observation, and the points are color-coded with respect to response $Y$. The black star indicates our observation of interest and each red cross marks to location of a voting point. 

From this figure we observe that for the bagging ensemble, there are several voting points that are very far from our observation of interest along the vertical axis, which shows $X_2$. Recall that in the Hidden2D model, the underlying DGP function $f_2(X_2) = -  \mathbbm{1}(0.6 \leq X_2 \leq 0.65)$ is only active along the narrow band in $X_2$ between 0.6 and 0.65. Our observation of interest lies within that region, so we should expect voting points to be clustered along $X_2$ to reduce training error. As we can see from the right plot in Figure \ref{voting_points_viz.fig}, the voting points of the random forests are much more clustered along the vertical axis. This is due to the fact that trees in the random forest tend to split on $X_2$ earlier and more frequently compared to bagging. Voting points must belong to the same leaf node as the observation of interest, so there cannot be a split between the two points.

\begin{figure}[h]
    \centering
    \includegraphics[width= \linewidth]{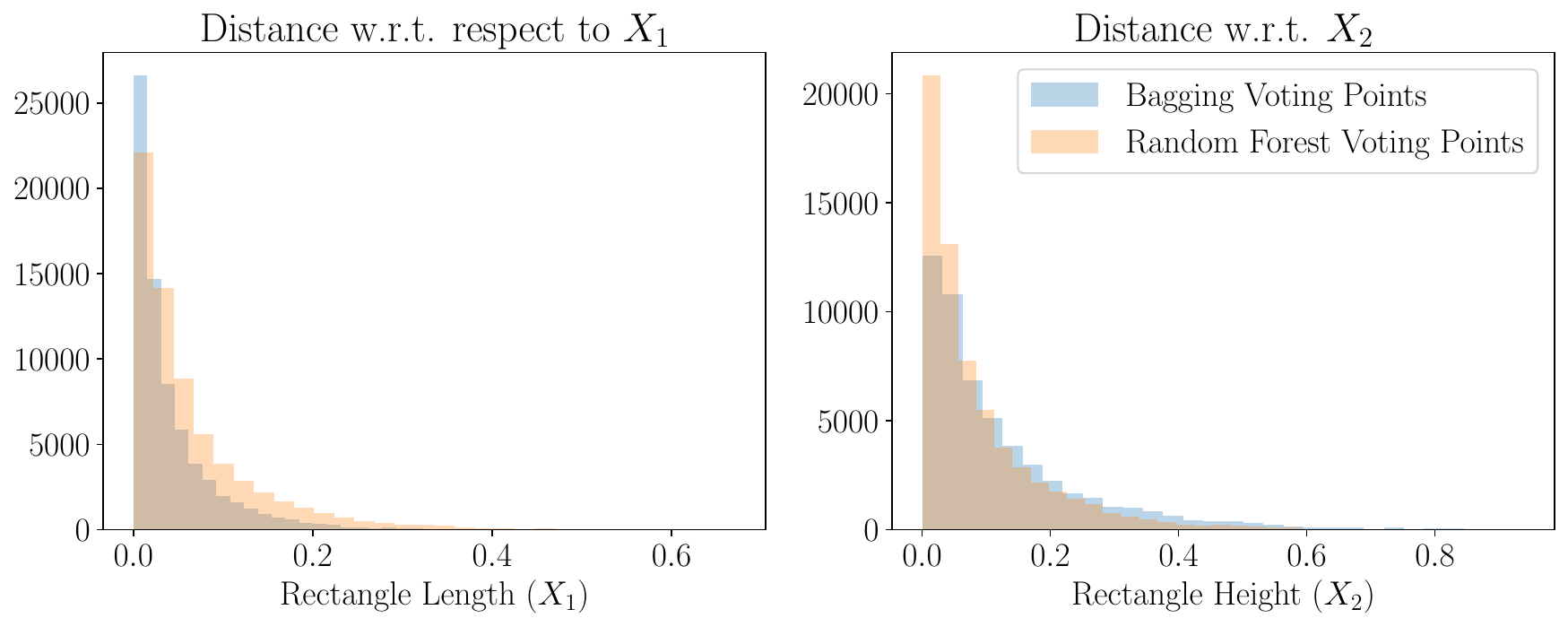}
    \caption{Distribution of distances from observation to voting points for all observations in the data.}
    \label{votingpoints_all.pdf}
\end{figure}

We use this procedure to visualize the difference in voting points between random forests and bagging ensembles across all observations in the data, rather than just a single observation of interest. For each observation $X^{(i)} \in X$, we find all of the voting points and report the height and length of the rectangle defined by $X^{(i)}$ and each voting point. The length of the rectangle is the distance between $X^{(i)}$ and the voting point with respect to feature $X_1$ and the height of the rectangle is the distance between $X^{(i)}$ and the voting point with respect to feature $X_2$. We plot the distribution of all these lengths and heights, for all observations and voting points, in Figure \ref{votingpoints_all.pdf}. From this figure, we can see that compared to bagging, random forest voting points are closer to all observations with respect to  $X_2$ and slightly further to all observations with respect to $X_1$. As such, these voting points better capture the DGP function $f_2(X_2)$ which reduces the training error of a random forest compared to bagging.

\section{Generalizing Hidden Patterns} \label{hidden_pattern_def.section}

Up until now we focus on a single example of a hidden pattern, the DGP function $f_2(X_2) = -  \mathbbm{1}(0.6 \leq X_2 \leq 0.65)$ in our Hidden2D model. Informally, we say that the relationship between feature $X_2$ and $Y$, $f_2(X_2)$, is masked by feature $X_1$ from a bagging ensemble. Trees in a bagging ensemble tend to split on $X_1$ before $X_2$ and it is harder for splits to capture $f_2(X_2)$ when conditioned on the preceding splits on $X_1$. Randomizing the features used in each split allows an ensemble to better capture this hidden pattern and we show in our experiments that this reduces the training error and expected bias of the model. We now formalize this notion of hidden patterns using the definition below.



\begin{definition}{Hidden Pattern}

Consider  data matrix $X \in \mathbb{R}^{n \times p}$ and let  $H$ be a subset of the $p$ features in $X$, i.e., $H \subset \{1,\ldots,p\}$. Let set $H^c$ denote the complement of set $H$, let $M$ be a subset of $H^c$, $M \subset H^c$, and let $M^c$ denote the complement of set $M$. Matrices $X_H$ and $X_{H^c}$ correspond to sub-matrices of $X$ comprising of columns indexed by  $H$ and ${H^c}$ respectively. Assume that response $Y \in \mathbb{R}^n$ is generated from the model:
\begin{align}\label{additive_defn.eq}
    Y = f_{H^c}(X_{H^c}) + f_H(X_H) + \epsilon,
\end{align}
where $\epsilon$ is a noise vector, and  let estimator $g$ be a bagging ensemble with $mtry = 1.0$.

We say that features $H$ are hidden from bagging ensemble by masking features $M$ if:
\begin{equation}\label{defn1.eq}
    \bigl(E[f_H(X_H) - \hat{g}(X)]\bigr)^2 \geq  \bigl( E[f_H(X_H) - \hat{g}(X_{M^c} )] \bigr)^2,
\end{equation}
where $\hat{g}(X)$ and $\hat{g}(X_{M^c})$  are the predictions of bagging ensembles fit to predict $Y$ using the data matrices $X$ and $X_{M^c}$ respectively and the expectations are taken with respect to randomness in the training data and the bootstrap in the bagging algorithm.  Function $f_H(X_H)$ is a hidden pattern that applies to hidden features $H$.

Furthermore, we say that a random forest helps  capture hidden pattern $f_H(X_H)$ if:
\begin{equation} \label{rf_display.eq}
    \bigl(E[f_H(X_H) - \hat{g}(X)]\bigr)^2 \geq  \bigl( E[f_H(X_H) - \hat{r}(X)] \bigr)^2,
\end{equation} where $\hat{r}(X)$ is the prediction of a random forest fit to predict $Y$ using data matrix $X$, with $mtry < 1.0$, and the expectations are taken with respect to randomness in the training data, the bootstrap in the bagging algorithm, and the bootstrap and $mtry$ randomization in the random forest algorithm.
\end{definition}

In our Hidden2D ($n = 1000$, $\text{SNR} = 6$) example, we have that $f_H(X_H) = f_2(X_2) = -\mathbbm{1}(0.6 \leq X_2 \leq 0.65)$ and that $M = \{1\}$. We repeat the Monte Carlo procedure from \cite{domingos1997does}, and which we used for our bias variance decomposition in \S\ref{Hidden2D_bvd.section}, for 500 trials to estimate the following quantities.
\begin{table}[h]\centering
\begin{tabular}{l|l|l|l|}
\cline{2-4}
                                        & $\bigl(E[f_H(X_H) - \hat{g}(X)]\bigr)^2$ & $\bigl(E[f_H(X_H) - \hat{g}(X_{M^c})] \bigr)^2$ & $\bigl( E[f_H(X_H) - \hat{r}(X)] \bigr)^2$ \\ \hline
\multicolumn{1}{|l|}{\textbf{Hidden2D}} & \multicolumn{1}{c|}{0.3326}              & \multicolumn{1}{c|}{0.2492}                              & \multicolumn{1}{c|}{0.3263}                \\ \hline
\end{tabular}
\end{table}

From this table, we see that $f_2(X_2)$ satisfies our definition of a hidden pattern and that SFS randomization helps  random forests capture the hidden pattern. Recall from Table \ref{hidden2d_bvd.table} in \S\ref{Hidden2D_bvd.section} that this corresponds to a decrease in ensemble bias. We note here that our definition of hidden patterns depend on $\bigl(E[f_H(X_H) - \hat{g}(X)]\bigr)^2$. Estimating this quantity requires knowledge of the underlying DGP, $f(X)$, and hidden pattern $f_H(X_H)$, functions that are often unknown in real-world settings. As such, we also focus on the following characteristic of hidden patterns that we observe in practice. \newline

\noindent \textbf{Visualizing Hidden Pattern Characteristics:} Consider again the top plot in the leftmost column of Figure \ref{hidden2d_high_snr.fig}, in \S\ref{hidden2d_first_study.section}. From this plot, we observe that the training errors of the bagging ensemble fit on the Hidden2D example are higher around the horizontal band $0.6 \leq X_2 \leq 0.65$. We say that the hidden pattern is \emph{active} for data points in this band, since the hidden pattern $f_2(X_2)$ has a large impact on response $Y$. In fact, for any data point  $\{X^{(i)}_1, X^{(i)}_2\} \in X$ such that $0.6 \leq X^{(i)}_2\ \leq 0.65$ we have that $|f_2\bigl(X^{(i)}_2\bigr)| \geq |f_1\bigl(X^{(i)}_1\bigr)|$. In the bottom plot in the leftmost column of Figure \ref{hidden2d_high_snr.fig}, we see that SFS random forests decrease the training error of data points around the horizontal band.

We observe that hidden patterns have this observable characteristic: bagging ensembles have higher training errors on data points where the hidden pattern is active and random forests reduce the training errors of these points. In the sections below, we show that this characteristic consistently appears across several different examples of hidden patterns.

\subsection{Examples of Hidden Patterns}\label{hidden_pattern_example.section}


In this section, we extend our concept of hidden patterns beyond the Hidden2D example. We use various functions, such as Gaussian, sine, and box functions to generate $f_H(X_H)$ across various dimensions of $X_H$. Empirically, we show that these hidden patterns satisfy the definition presented above and we visualize their observable characteristics. These examples establish a more general notion of hidden patterns and suggest that hidden patterns may be present in real-world data, an idea we further explore in \S\ref{real_world_case.section}. 
\subsubsection{Hidden Gaussian Spikes}
\begin{figure}[h]
    \centering
    \includegraphics[width=0.95\linewidth]{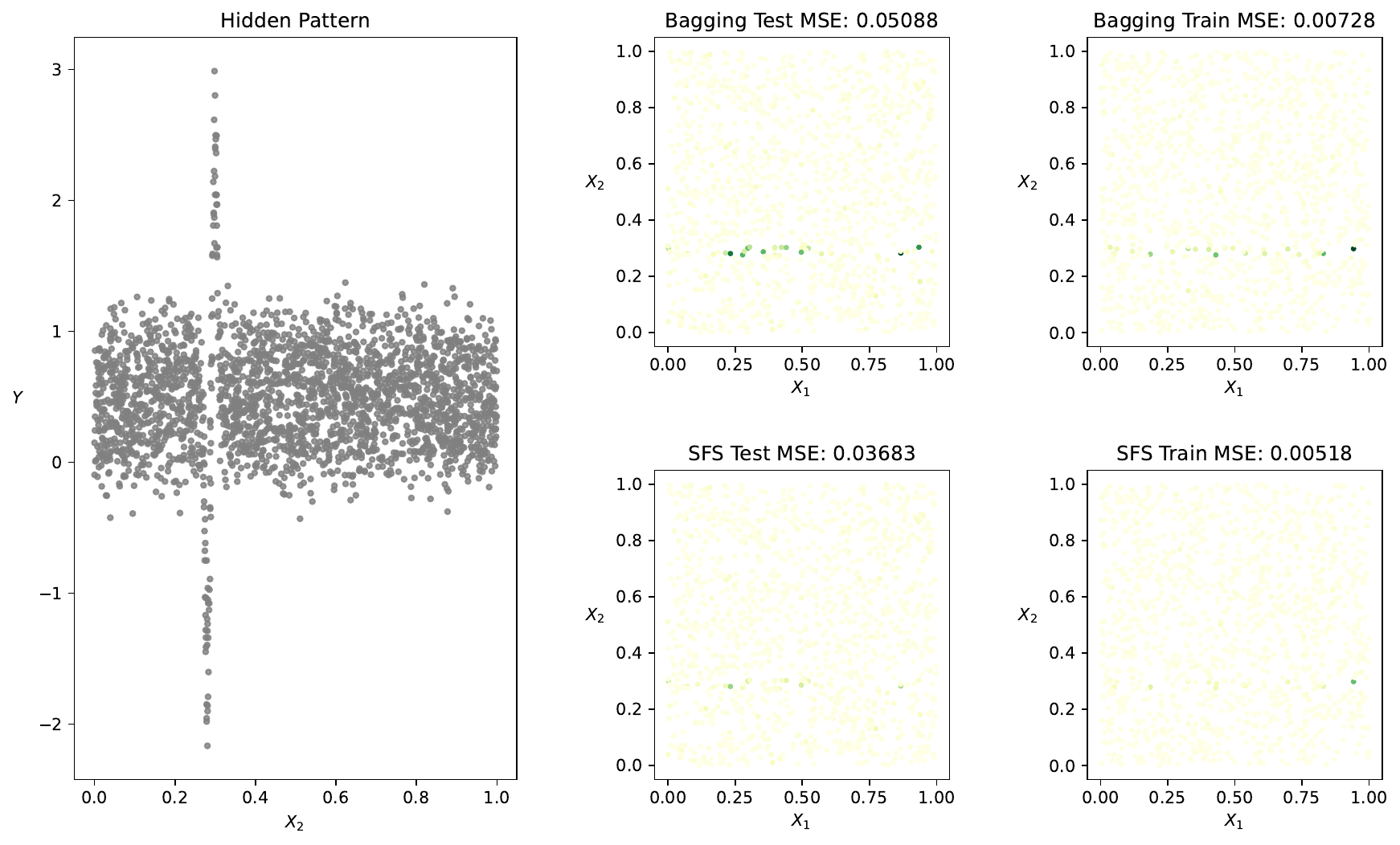}
    \caption{Visualization of a Gaussian spike hidden pattern. The four plots on the right side of this plot show the characteristic of a hidden pattern, that bagging ensembles have higher errors in regions where the hidden pattern is active compared to SFS random forests.}
    \label{hidden_gaussian_spike.fig}
\end{figure}

Consider the model:
\begin{align}
    Y = X_1 - \underbrace{2 \exp\left(-\frac{(X_2 - 0.28)^2}{2(0.005)^2}\right) + 2 \exp\left(-\frac{(X_2 - 0.3)^2}{2(0.005)^2}\right)}_{f_H(X_H)} + \epsilon.
\end{align} The hidden pattern in this model consists of two Gaussian spikes in $X_2$, labeled by the under-brace in the above display, and masking feature $M = \{1\}$. Using this model we generate $n = 2500$ data points, by sampling $X$ from a $U(0,1)$ distribution, and set $\epsilon$ such that the SNR is equal to 6. We show the hidden pattern by plotting $Y$ against $X_2$ in the leftmost plot in Figure \ref{hidden_gaussian_spike.fig}.

On this example, we fit bagging ensembles and SFS random forests ($mtry = 0.5$) with 500 trees and 
 repeat our Monte Carlo procedure (\S\ref{Hidden2D_bvd.section}) across 500 trials to estimate the following quantities.
\begin{table}[h]\centering
\begin{tabular}{l|l|l|l|}
\cline{2-4}
                                                   & $\bigl(E[f_H(X_H) - \hat{g}(X)]\bigr)^2$ & $\bigl(E[f_H(X_H) - \hat{g}(X_{M^c})] \bigr)^2$ & $\bigl( E[f_H(X_H) - \hat{r}(X)] \bigr)^2$ \\ \hline
\multicolumn{1}{|l|}{\textbf{HiddenSpike}} & \multicolumn{1}{c|}{0.3337}              & \multicolumn{1}{c|}{0.2497}                              & \multicolumn{1}{c|}{0.3229}                \\ \hline
\end{tabular}
\end{table}

We note that $f_H(X_H)$ satisfies our definition of a hidden pattern and that randomization helps random forests capture the pattern. We also apply the same Monte Carlo procedure to estimate the bias and variance of a bagging ensemble and a SFS random forest on this example. From the table below, we observe that SFS random forests again reduce ensemble bias and variance compared to bagging. 

\begin{table}[h]
\centering
\begin{tabular}{|c|c|c|}
\hline
               \textbf{HiddenSpike}   & $\text{Bias}^2$ & Variance \\ \hline
SFS Random Forest & 0.00310       & 0.0058  \\ \hline
Bagging Ensemble  & 0.00821        & 0.0107 \\ \hline
\end{tabular}
\end{table} 

Finally, the four plots on the right in Figure \ref{hidden_gaussian_spike.fig} show the training and test errors of a bagging ensemble and a SFS random forest for each data point in the example, using a 50-50 train-test split. From the top right plot we see that training error for bagging ensembles is high for the data-points affected by the hidden pattern, in the region around $X_2 \approx 0.3$. SFS random forests reduces the training error of these points, as expected.

\subsubsection{Hidden Sinusoidal}
\begin{figure}[h]
    \centering
    \includegraphics[width=0.95\linewidth]{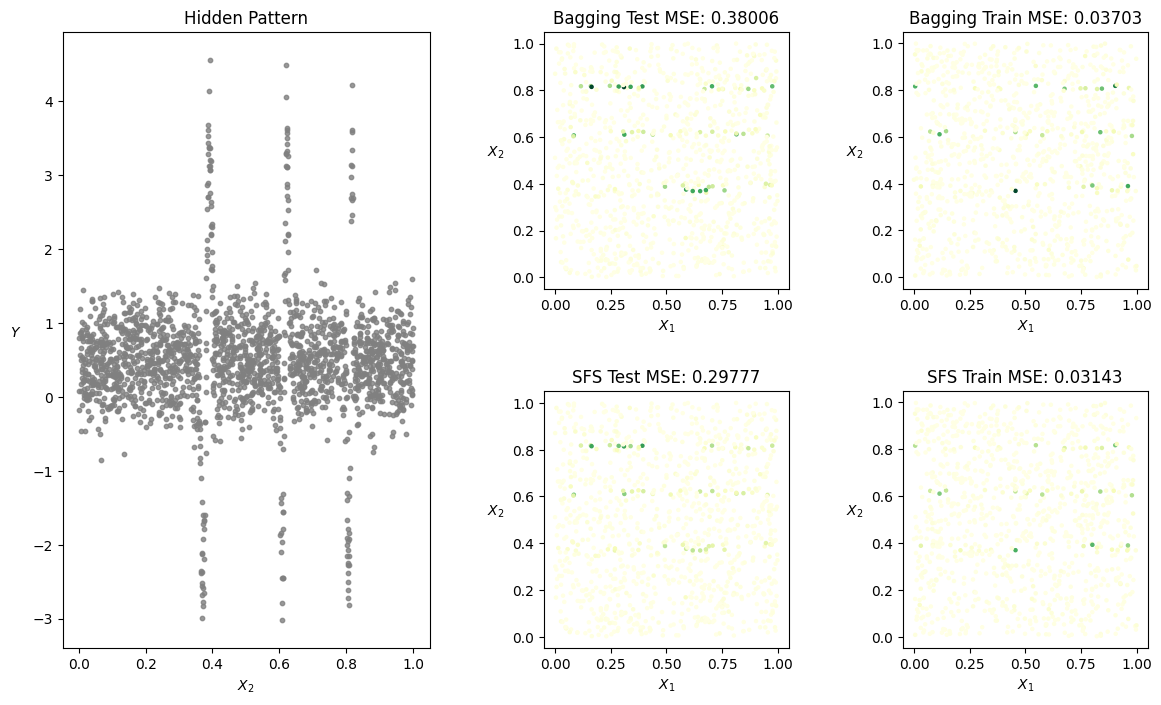}
    \caption{Visualization of the sine hidden pattern.}
    \label{hidden_sine.fig}
\end{figure}

We repeat the analyses discussed in the section above on the following model:
\begin{align}
    Y &= X_1  -\underbrace{\bigl( 3 \sin(50 \pi X_2) \mathbbm{1}_{X_2 \in [0.36, 0.4]} 
    + 3 \sin(70 \pi X_2) \mathbbm{1}_{X_2 \in[0.6, 0.63]} 
    + 3 \sin(90 \pi X_2) \mathbbm{1}_{X_2 \in[0.8, 0.82]} \bigr)}_{f_H(X_H)} + \epsilon,
\end{align}
where the hidden pattern $f_H(X_H)$ consists of 3 sinusoidal patterns defined along intervals of $X_2$ and masking feature $M = \{1\}$. We generate $n = 2000$ data points using this model, with $\epsilon$ set so that $\text{SNR} = 6$; the leftmost plot in Figure \ref{hidden_sine.fig} shows a visualization of the hidden pattern.

We fit bagging ensembles and SFS random forests ($mtry = 0.5$) and repeat the procedure above to estimate the following quantities.
\begin{table}[h]\centering
\begin{tabular}{l|l|l|l|}
\cline{2-4}
                                                   & $\bigl(E[f_H(X_H) - \hat{g}(X)]\bigr)^2$ & $\bigl(E[f_H(X_H) - \hat{g}(X_{M^c})] \bigr)^2$ & $\bigl( E[f_H(X_H) - \hat{r}(X)] \bigr)^2$ \\ \hline
\multicolumn{1}{|l|}{\textbf{HiddenSine}} & \multicolumn{1}{c|}{0.42412}              & \multicolumn{1}{c|}{0.24951}                              & \multicolumn{1}{c|}{0.40307}                \\ \hline
\end{tabular}
\end{table}

Again, we see that $f_H(X_H)$ satisfies our definition of a hidden pattern and furthermore we show in the table below that SFS random forests reduce both bias and variance compared to bagging in this example.
\begin{table}[h]
\centering
\begin{tabular}{|c|c|c|}
\hline
               \textbf{HiddenSine}   & $\text{Bias}^2$ & Variance \\ \hline
SFS Random Forest & 0.06786        & 0.03495  \\ \hline
Bagging Ensemble  & 0.0876        & 0.05338  \\ \hline
\end{tabular}
\end{table}

Finally, the four plots on the right of Figure \ref{hidden_sine.fig} show the training and test errors of each ensemble for each data point. Again, we observe that our HiddenSine model shows the characteristics of a hidden pattern; the training errors of the bagging ensemble are concentrated around the points impacted by the hidden pattern, the three horizontal bands, and these errors are reduced by the random forest.

\subsubsection{Hidden Square and Hidden Circle} \label{hidden_square_circle.section}
\begin{figure}[h]
    \centering
    \includegraphics[width=0.95\linewidth]{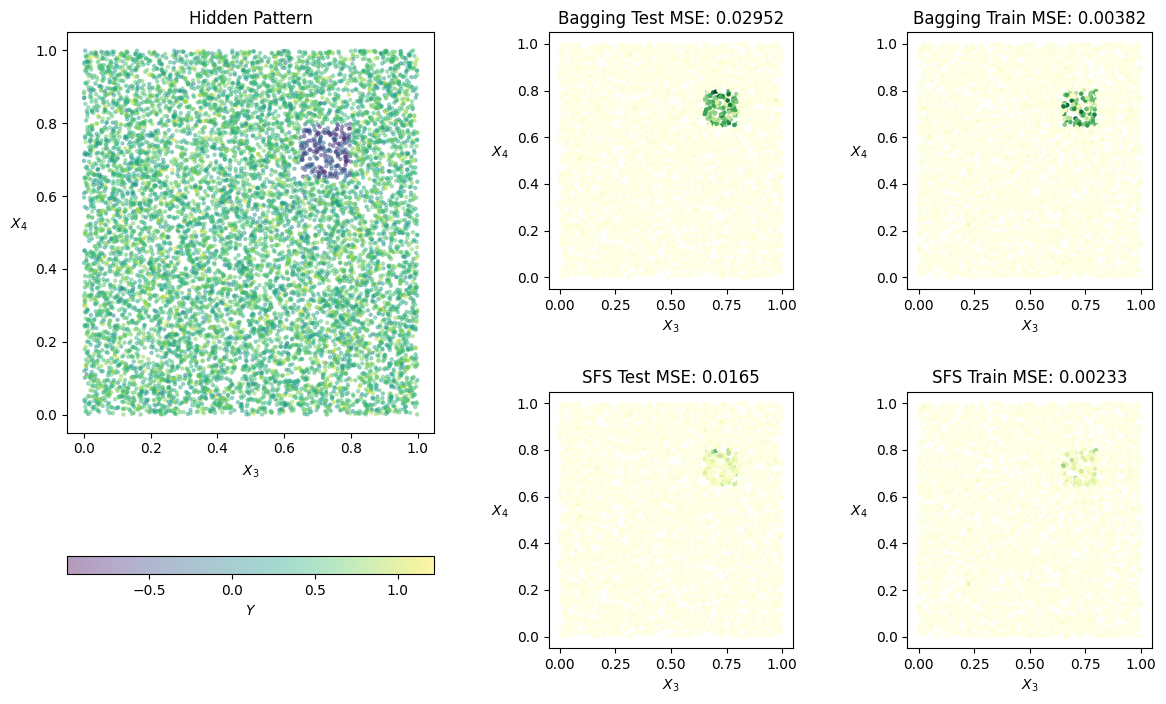}
    \caption{HiddenSquare visualization}
    \label{hidden_square.fig}
\end{figure}

Now consider the model:
\begin{align}
     Y &= 0.5 X_1 + 0.5 X_2 \notag \\
    &\quad - \underbrace{\left( \mathbbm{1}_{X_3 \in [0.65, 0.8], X_4 \in [0.65, 0.8]} \right)}_{f_H(X_H)} + \epsilon,
\end{align}
which we call HiddenSquare. The hidden pattern $f_H(X_H)$ is defined by a box in  $X_3$ and $X_4$ and we have masking features $M = \{1,2\}$; the hidden pattern is jointly a function of these features. Using this HiddenSquare model we generate $n = 10000$ data points with $\epsilon$ set such that the SNR of the data is 6. In the left plot of Figure \ref{hidden_square.fig}, we show a visualization of this hidden pattern by plotting $X_3$ and $X_4$; the color of points in this scatter plot show $Y$.

Now, we fit bagging ensembles and SFS random forests, with $mtry = 0.333$, and repeat the procedures discussed above to estimate the following quantities.
\begin{table}[h]\centering
\begin{tabular}{l|l|l|l|}
\cline{2-4}
                                                   & $\bigl(E[f_H(X_H) - \hat{g}(X)]\bigr)^2$ & $\bigl(E[f_H(X_H) - \hat{g}(X_{M^c})] \bigr)^2$ & $\bigl( E[f_H(X_H) - \hat{r}(X)] \bigr)^2$ \\ \hline
\multicolumn{1}{|l|}{\textbf{HiddenSquare}} & \multicolumn{1}{c|}{0.30046}              & \multicolumn{1}{c|}{0.25046}                              & \multicolumn{1}{c|}{0.28862}                \\ \hline
\end{tabular}
\end{table}
\begin{table}[h]
\centering
\begin{tabular}{|c|c|c|}
\hline
               \textbf{HiddenSquare}   & $\text{Bias}^2$ & Variance \\ \hline
SFS Random Forest & 0.00329        & 0.000871  \\ \hline
Bagging Ensemble  & 0.00991      & 0.002439  \\ \hline
\end{tabular}
\end{table}

 From these results, we see that $f_H(X_H)$ in our HiddenSquare models satisfies our definition of a hidden pattern and that SFS random forests help capture the pattern. Moreover, SFS random forests again reduce both the bias and variance of the ensemble compared to bagging.

In the four right plots in Figure \ref{hidden_square.fig}, we show the errors of each ensemble on each data point in the HiddenSquare model. We plot these errors against $X_3$ and $X_4$ in these scatter plots and show again that the training errors of the bagging ensemble are high for data points impacted by the hidden pattern. These points correspond to the box defined on $X_3$ and $X_4$ and SFS random forests reduce the training error in this region.

\begin{figure}[h]
    \centering
    \includegraphics[width=0.95\linewidth]{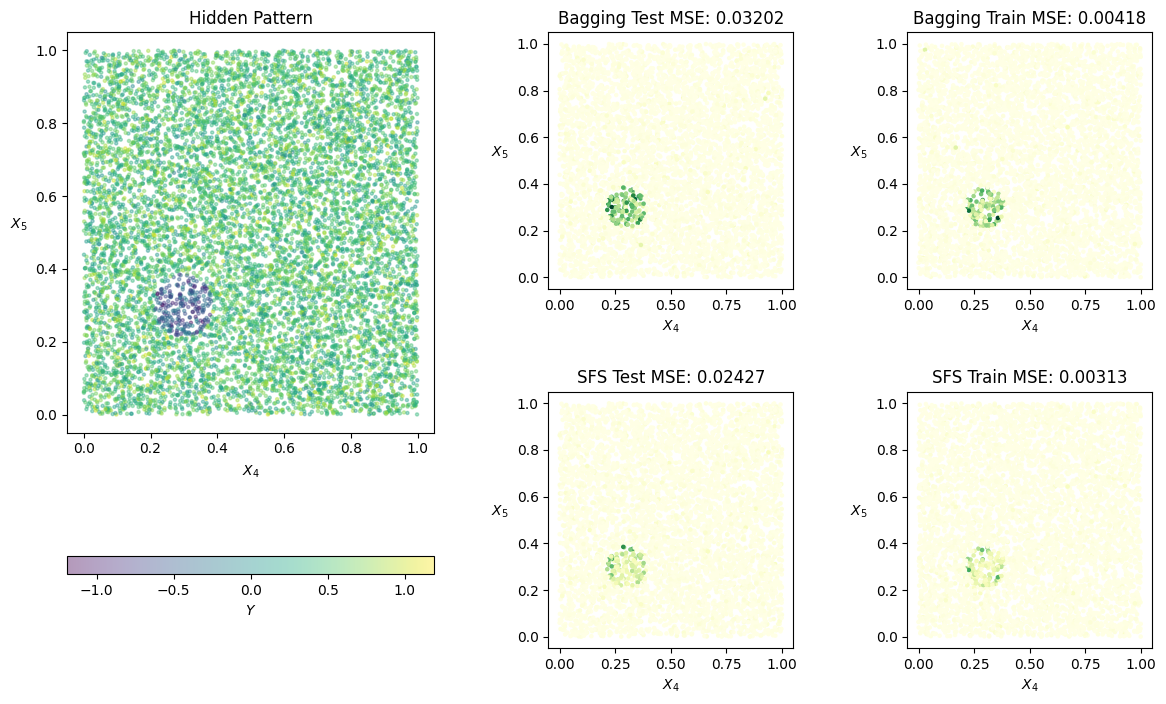}
    \caption{HiddenCircle visualization.}
    \label{hiddencircle.fig}
\end{figure}

We repeat this entire analysis on a similar model, defined by:
\begin{align}
    Y &= 0.74 X_1 + 0.26 X_2 - 0.1 X_3^2 \notag \\
    &\quad - \underbrace{\left(  \mathbbm{1}_{(X_4 - 0.3)^2 + (X_5 - 0.3)^2 \leq 0.0075} \right)}_{f_H(X_H)} + \epsilon,
\end{align}
which we call HiddenCircle. The hidden pattern $f_H(X_H)$ is now defined by a circle in $X_4$ and $X_5$ and masking features $M = \{1,2,3\}$. Using this model, we again generate $n = 10000$ data points with $\text{SNR} = 6$ and show a visualization of the hidden pattern in the left plot in Figure \ref{hiddencircle.fig}.

\newpage
Repeating the analysis from the HiddenSquare example, we fit random forests ($mtry = 0.333$) and bagging ensembles and estimate our quantites of interest.
\begin{table}[h]\centering
\begin{tabular}{l|l|l|l|}
\cline{2-4}
                                                   & $\bigl(E[f_H(X_H) - \hat{g}(X)]\bigr)^2$ & $\bigl(E[f_H(X_H) - \hat{g}(X_{M^c})] \bigr)^2$ & $\bigl( E[f_H(X_H) - \hat{r}(X)] \bigr)^2$ \\ \hline
\multicolumn{1}{|l|}{\textbf{HiddenCircle}} & \multicolumn{1}{c|}{0.28233}              & \multicolumn{1}{c|}{0.21912}                              & \multicolumn{1}{c|}{0.26828}                \\ \hline
\end{tabular}
\end{table}
\begin{table}[h]
\centering
\begin{tabular}{|c|c|c|}
\hline
               \textbf{HiddenCircle}   & $\text{Bias}^2$ & Variance \\ \hline
SFS Random Forest & 0.00798        & 0.000875  \\ \hline
Bagging Ensemble  & 0.01208      & 0.002378  \\ \hline
\end{tabular}
\end{table}
As in the HiddenSquare example, we see that our HiddenCircle hidden pattern satisfies our definition of hidden patterns, and that SFS random forests reduce both bias and variance. From the plots in Figure \ref{hiddencircle.fig}, we again see that the errors of the bagging ensemble are concentrated around points impacted by the hidden pattern, and that SFS random forests reduces the error of these points.

\subsubsection{Hidden Cube}
\begin{figure}[h]
    \centering
    \includegraphics[width=0.95\linewidth]{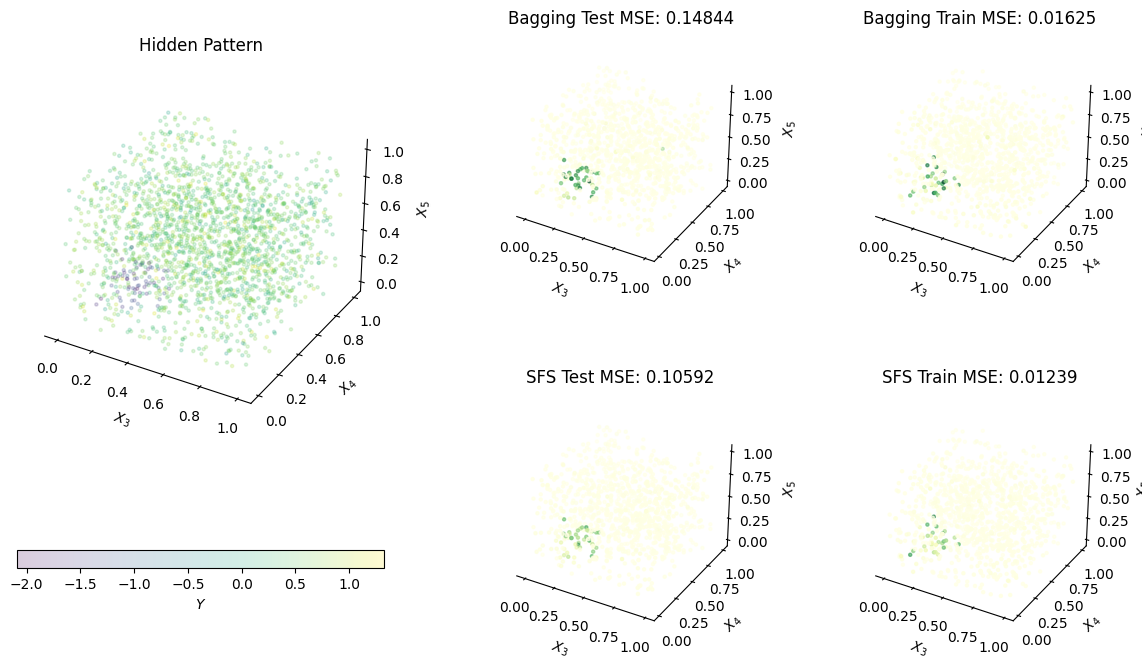}
    \caption{Visualization of cube hidden pattern.}
    \label{hidden_cube.fig}
\end{figure}

Finally, we consider the model:
\begin{align}
    Y &= 0.5 X_1 + 0.5 X_2 - \underbrace{2\left( \mathbbm{1}_{X_3, X_4, X_5 \in [0.1, 0.4]} \right)}_{f_H(X_H)} + \epsilon,
\end{align}
which we call HiddenCube. Here, the hidden pattern $f_H(X_H)$ is defined by the cube in $X_3$, $X_4$, and $X_5$ and masking features $M = \{1,2\}$. We use this model to generate $n = 2000$ data points with an SNR of 6. In the leftmost plot in Figure \ref{hidden_cube.fig}, we show a visualization of this hidden pattern by plotting $X_3$, $X_4$, and $X_5$; the color of each point in the plot shows $Y$.

Repeating the procedures from above, we fit random forests ($mtry = 0.333$) and bagging ensembles and estimate the following quantities.
\begin{table}[h]\centering
\begin{tabular}{l|l|l|l|}
\cline{2-4}
                                                   & $\bigl(E[f_H(X_H) - \hat{g}(X)]\bigr)^2$ & $\bigl(E[f_H(X_H) - \hat{g}(X_{M^c})] \bigr)^2$ & $\bigl( E[f_H(X_H) - \hat{r}(X)] \bigr)^2$ \\ \hline
\multicolumn{1}{|l|}{\textbf{HiddenCube}} & \multicolumn{1}{c|}{0.4201}              & \multicolumn{1}{c|}{0.3698}                              & \multicolumn{1}{c|}{0.4006}                \\ \hline
\end{tabular}
\end{table}
\begin{table}[h]
\centering
\begin{tabular}{|c|c|c|}
\hline
               \textbf{HiddenCube}   & $\text{Bias}^2$ & Variance \\ \hline
SFS Random Forest & 0.1200        & 0.001366  \\ \hline
Bagging Ensemble  & 0.1300     & 0.003410  \\ \hline
\end{tabular}
\end{table}
Again, we see that $f_H(X_H)$ satisfies our definition of a hidden pattern and that SFS random forests help better capture this pattern. As such, SFS random forests have lower bias and variance compared to bagging ensembles when fit on this HiddenCube model.

Finally, the right four plots in Figure \ref{hidden_cube.fig} show the errors of each ensemble plotted against $\{X_3, X_4, X_5\}$. From this plot, we again see that the training errors of the bagging ensemble are high for points impacted by the hidden pattern, the dark cube in the plots, and the SFS random forests reduces the errors of these points.

\newpage
\subsection{Additional Empirical Studies}\label{empirical_study.section}

In the section above, we explore in detail several examples of hidden patterns and discuss their characteristics. Here, we conduct a large scale empirical study to show that random forests reduce both bias and variance across a range of examples that vary in terms of data generating procedures and dataset sizes. We generate models for this study using the formula:
\begin{align}
    Y = \underbrace{f_{H^c}(X_{H^c}) + f_H(X_H)}_{f(X)} + \epsilon,
\end{align}
where DGP function $f(X) = f_{H^c}(X_{H^c}) + f_H(X_H)$. For the functions $f_{H^c}(X_{H^c})$ we consider the MARS and MARSadd functions defined in displays \eqref{mars.display} and \eqref{mars_add.display} along with the following functions:
\begin{align}
     0.3 X_1 - 0.5 X_2 + 0.1 X_3 + 0.1 X_4 + X_5 \tag{Linear}
\end{align}
\begin{align}
   2.0 + 3.5 X_1 - 1.2 X_1^2 + 4.0 X_2 + 0.8 X_2^3 
    \tag{Polynomial}
\end{align}
\begin{align}
    \sqrt{X_1^2 + \left( X_2 X_3 - \frac{1}{X_2 X_4 + 0.1} \right)^2} 
    \tag{Friedman2}
\end{align}
\begin{align}
    \arctan \left( X_2 X_3 - \frac{1}{X_2 X_4+0.1} \right).
    \tag{Friedman3}
\end{align}
For hidden patterns $f_H(X_H)$ we consider functions of the form found in Hidden2D, HiddenSpike, and HiddenSquare.

We consider various problem sizes with $n \in \{1000, 5000, 10000, 20000\}$. Similar to the studies above, we sample data matrix $X$ from a $U(0,1)$ distribution. For all of the models generated, we set $\epsilon$ such that the SNR of the data is equal to 6. The full expression for each model considered in our study can be found in Section C.1 of the appendix.

On each model, we repeat the Monte Carlo procedure used in \cite{domingos2000unified}, and in sections \ref{Hidden2D_bvd.section} and \ref{hidden_pattern_example.section} of our paper, 500 times to estimate the bias and variance of a bagging ensemble ($mtry = 1.0$) and random forest ($mtry < 1.0$) fit on the data. The ensembles are grown to have 500 trees and are identical except for $mtry$.

We show the results of our empirical study in Table \ref{empirical_study.table}. From this table, we observe that random forests can reduce both bias and variance compared to bagging ensembles across a wide range of problems. In each of these models for every problem size, the bias and variance of the random forest is lower than that of the bagging ensemble.

Finally, we note that our definition of hidden patterns, along with the examples and experiments discussed above, assumes that the data-generating procedure is an additive combination of the effects of the hidden pattern and the masking features, see display \eqref{additive_defn.eq}. This assumption may not be necessary, and exploring whether \eqref{defn1.eq} and \eqref{rf_display.eq} hold under other data-generating procedures is an interesting direction for future research.

\newpage

\begin{table}[h]\centering\scalebox{0.7}{
\begin{tabular}{|c|c|c|c|c|c|}
\hline
\textbf{Model Name}              & \textbf{n} & \textbf{Random Forest Bias\(^2\)} & \textbf{Bagging Bias\(^2\)} & \textbf{Random Forest Variance} & \textbf{Bagging Variance} \\ \hline
\textbf{Friedman2 Hidden2D}      & 1000       & \cellcolor[HTML]{9AFF99}0.4543    & 0.4642                      & \cellcolor[HTML]{9AFF99}0.1843  & 0.2429                    \\ \hline
\textbf{Friedman2 Hidden2D}      & 5000       & \cellcolor[HTML]{9AFF99}0.3234    & 0.4113                      & \cellcolor[HTML]{9AFF99}0.1284  & 0.1732                    \\ \hline
\textbf{Friedman2 Hidden2D}      & 10000      & \cellcolor[HTML]{9AFF99}0.2303    & 0.3341                      & \cellcolor[HTML]{9AFF99}0.1078  & 0.1455                    \\ \hline
\textbf{Friedman2 Hidden2D}      & 20000      & \cellcolor[HTML]{9AFF99}0.181     & 0.3135                      & \cellcolor[HTML]{9AFF99}0.0957  & 0.1294                    \\ \hline
\textbf{Friedman3 Hidden2D}      & 1000       & \cellcolor[HTML]{9AFF99}0.071     & 0.0953                      & \cellcolor[HTML]{9AFF99}0.0052  & 0.0158                    \\ \hline
\textbf{Friedman3 Hidden2D}      & 5000       & \cellcolor[HTML]{9AFF99}0.012     & 0.0231                      & \cellcolor[HTML]{9AFF99}0.0043  & 0.0083                    \\ \hline
\textbf{Friedman3 Hidden2D}      & 10000      & \cellcolor[HTML]{9AFF99}0.0138    & 0.0276                      & \cellcolor[HTML]{9AFF99}0.0013  & 0.0048                    \\ \hline
\textbf{Friedman3 Hidden2D}      & 20000      & \cellcolor[HTML]{9AFF99}0.0098    & 0.0223                      & \cellcolor[HTML]{9AFF99}0.001   & 0.004                     \\ \hline
\textbf{Polynomial Hidden2D}     & 1000       & \cellcolor[HTML]{9AFF99}1.3861    & 1.5128                      & \cellcolor[HTML]{9AFF99}0.1029  & 0.2487                    \\ \hline
\textbf{Polynomial Hidden2D}     & 5000       & \cellcolor[HTML]{9AFF99}0.9514    & 1.4184                      & \cellcolor[HTML]{9AFF99}0.0579  & 0.1646                    \\ \hline
\textbf{Polynomial Hidden2D}     & 10000      & \cellcolor[HTML]{9AFF99}0.6536    & 1.1388                      & \cellcolor[HTML]{9AFF99}0.0442  & 0.1375                    \\ \hline
\textbf{Polynomial Hidden2D}     & 20000      & \cellcolor[HTML]{9AFF99}0.4605    & 0.9383                      & \cellcolor[HTML]{9AFF99}0.0364  & 0.1198                    \\ \hline
\textbf{Polynomial HiddenSquare} & 1000       & \cellcolor[HTML]{9AFF99}0.6893    & 0.6923                      & \cellcolor[HTML]{9AFF99}0.0968  & 0.1465                    \\ \hline
\textbf{Polynomial HiddenSquare} & 5000       & \cellcolor[HTML]{9AFF99}0.4917    & 0.555                       & \cellcolor[HTML]{9AFF99}0.0296  & 0.103                     \\ \hline
\textbf{Polynomial HiddenSquare} & 10000      & \cellcolor[HTML]{9AFF99}0.4497    & 0.6177                      & \cellcolor[HTML]{9AFF99}0.0255  & 0.0872                    \\ \hline
\textbf{Polynomial HiddenSquare} & 20000      & \cellcolor[HTML]{9AFF99}0.2923    & 0.5334                      & \cellcolor[HTML]{9AFF99}0.0204  & 0.0752                    \\ \hline
\textbf{Polynomial HiddenSpike}  & 1000       & \cellcolor[HTML]{9AFF99}0.4035    & 0.4457                      & \cellcolor[HTML]{9AFF99}0.1026  & 0.1518                    \\ \hline
\textbf{Polynomial HiddenSpike}  & 5000       & \cellcolor[HTML]{9AFF99}0.2131    & 0.4224                      & \cellcolor[HTML]{9AFF99}0.0359  & 0.102                     \\ \hline
\textbf{Polynomial HiddenSpike}  & 10000      & \cellcolor[HTML]{9AFF99}0.1437    & 0.3254                      & \cellcolor[HTML]{9AFF99}0.0295  & 0.0879                    \\ \hline
\textbf{Polynomial HiddenSpike}  & 20000      & \cellcolor[HTML]{9AFF99}0.0921    & 0.2315                      & \cellcolor[HTML]{9AFF99}0.0469  & 0.0792                    \\ \hline
\textbf{Linear Hidden2D}         & 1000       & \cellcolor[HTML]{9AFF99}0.0813    & 0.0953                      & \cellcolor[HTML]{9AFF99}0.0062  & 0.0162                    \\ \hline
\textbf{Linear Hidden2D}         & 5000       & \cellcolor[HTML]{9AFF99}0.0548    & 0.0867                      & \cellcolor[HTML]{9AFF99}0.0035  & 0.0107                    \\ \hline
\textbf{Linear Hidden2D}         & 10000      & \cellcolor[HTML]{9AFF99}0.044     & 0.0757                      & \cellcolor[HTML]{9AFF99}0.0028  & 0.0094                    \\ \hline
\textbf{Linear Hidden2D}         & 20000      & \cellcolor[HTML]{9AFF99}0.0334    & 0.0624                      & \cellcolor[HTML]{9AFF99}0.0022  & 0.0083                    \\ \hline
\textbf{Linear HiddenSquare}     & 1000       & \cellcolor[HTML]{9AFF99}0.1865    & 0.1928                      & \cellcolor[HTML]{9AFF99}0.0226  & 0.037                     \\ \hline
\textbf{Linear HiddenSquare}     & 5000       & \cellcolor[HTML]{9AFF99}0.0345    & 0.0377                      & \cellcolor[HTML]{9AFF99}0.0023  & 0.0075                    \\ \hline
\textbf{Linear HiddenSquare}     & 10000      & \cellcolor[HTML]{9AFF99}0.0384    & 0.0452                      & \cellcolor[HTML]{9AFF99}0.0019  & 0.006                     \\ \hline
\textbf{Linear HiddenSquare}     & 20000      & \cellcolor[HTML]{9AFF99}0.0343    & 0.0436                      & \cellcolor[HTML]{9AFF99}0.0016  & 0.0053                    \\ \hline
\textbf{Mars Hidden2D}           & 1000       & \cellcolor[HTML]{9AFF99}59.9471   & 70.2535                     & \cellcolor[HTML]{9AFF99}5.1018  & 10.7458                   \\ \hline
\textbf{Mars Hidden2D}           & 5000       & \cellcolor[HTML]{9AFF99}33.4228   & 40.5931                     & \cellcolor[HTML]{9AFF99}2.882   & 6.9979                    \\ \hline
\textbf{Mars Hidden2D}           & 10000      & \cellcolor[HTML]{9AFF99}20.2106   & 24.7232                     & \cellcolor[HTML]{9AFF99}2.1899  & 5.5618                    \\ \hline
\textbf{Mars Hidden2D}           & 20000      & \cellcolor[HTML]{9AFF99}13.431    & 15.426                      & \cellcolor[HTML]{9AFF99}3.1332  & 4.7419                    \\ \hline
\textbf{Mars HiddenSquare}       & 1000       & \cellcolor[HTML]{9AFF99}63.7704   & 64.8151                     & \cellcolor[HTML]{9AFF99}15.1242 & 26.2896                   \\ \hline
\textbf{Mars HiddenSquare}       & 5000       & \cellcolor[HTML]{9AFF99}119.0924  & 124.2567                    & \cellcolor[HTML]{9AFF99}4.849   & 17.9954                   \\ \hline
\textbf{Mars HiddenSquare}       & 10000      & \cellcolor[HTML]{9AFF99}140.8299  & 154.133                     & \cellcolor[HTML]{9AFF99}3.8958  & 14.8447                   \\ \hline
\textbf{Mars HiddenSquare}       & 20000      & \cellcolor[HTML]{9AFF99}121.5901  & 142.0494                    & \cellcolor[HTML]{9AFF99}3.2596  & 12.7916                   \\ \hline
\textbf{Mars HiddenSpike}        & 1000       & \cellcolor[HTML]{9AFF99}103.9615  & 119.2452                    & \cellcolor[HTML]{9AFF99}13.2049 & 22.7455                   \\ \hline
\textbf{Mars HiddenSpike}        & 5000       & \cellcolor[HTML]{9AFF99}32.0277   & 35.1137                     & \cellcolor[HTML]{9AFF99}6.9643  & 13.627                    \\ \hline
\textbf{Mars HiddenSpike}        & 10000      & \cellcolor[HTML]{9AFF99}12.2086   & 14.1591                     & \cellcolor[HTML]{9AFF99}4.4306  & 7.9408                    \\ \hline
\textbf{Mars HiddenSpike}        & 20000      & \cellcolor[HTML]{9AFF99}30.7046   & 39.0582                     & \cellcolor[HTML]{9AFF99}4.1251  & 7.4426                    \\ \hline
\textbf{Marsadd HiddenSquare}    & 1000       & \cellcolor[HTML]{9AFF99}134.3366  & 143.7545                    & \cellcolor[HTML]{9AFF99}4.979   & 15.0782                   \\ \hline
\textbf{Marsadd HiddenSquare}    & 5000       & \cellcolor[HTML]{9AFF99}31.061    & 41.9002                     & \cellcolor[HTML]{9AFF99}4.6501  & 7.9683                    \\ \hline
\textbf{Marsadd HiddenSquare}    & 10000      & \cellcolor[HTML]{9AFF99}23.6029   & 32.295                      & \cellcolor[HTML]{9AFF99}4.3727  & 7.7874                    \\ \hline
\textbf{Marsadd HiddenSquare}    & 20000      & \cellcolor[HTML]{9AFF99}8.8271    & 15.2515                     & \cellcolor[HTML]{9AFF99}2.3843  & 4.7962                    \\ \hline
\textbf{Marsadd Hidden2d}        & 1000       & \cellcolor[HTML]{9AFF99}4.602     & 4.9909                      & \cellcolor[HTML]{9AFF99}0.4208  & 0.9364                    \\ \hline
\textbf{Marsadd Hidden2d}        & 5000       & \cellcolor[HTML]{9AFF99}2.1736    & 2.523                       & \cellcolor[HTML]{9AFF99}0.2286  & 0.5734                    \\ \hline
\textbf{Marsadd Hidden2d}        & 10000      & \cellcolor[HTML]{9AFF99}1.4905    & 1.737                       & \cellcolor[HTML]{9AFF99}0.1782  & 0.4576                    \\ \hline
\textbf{Marsadd Hidden2d}        & 20000      & \cellcolor[HTML]{9AFF99}1.0318    & 1.2295                      & \cellcolor[HTML]{9AFF99}0.1465  & 0.3671                    \\ \hline
\textbf{Marsadd HiddenSpike}     & 1000       & \cellcolor[HTML]{9AFF99}19.7094   & 21.4458                     & \cellcolor[HTML]{9AFF99}2.8458  & 4.7252                    \\ \hline
\textbf{Marsadd HiddenSpike}     & 5000       & \cellcolor[HTML]{9AFF99}1.0263    & 1.6187                      & \cellcolor[HTML]{9AFF99}0.0918  & 0.27                      \\ \hline
\textbf{Marsadd HiddenSpike}     & 10000      & \cellcolor[HTML]{9AFF99}0.7592    & 1.3841                      & \cellcolor[HTML]{9AFF99}0.0734  & 0.2332                    \\ \hline
\textbf{Marsadd HiddenSpike}     & 20000      & \cellcolor[HTML]{9AFF99}0.5599    & 1.1348                      & \cellcolor[HTML]{9AFF99}0.0616  & 0.2009                    \\ \hline
\end{tabular}}
\caption{Results of our empirical study, cells with the lower bias or variance are highlighted in green. Random forests reduce bias and variance compared to bagging across all of the models and settings considered in the study. }
\label{empirical_study.table}
\end{table}
\newpage

\subsection{Case Study: Real World Hidden Pattern} \label{real_world_case.section}

We conclude this section with a case study on what we believe to be a hidden pattern on a real-world dataset: the California Housing Dataset from \cite{pace1997sparse}. This dataset has 20640 observations, each data point corresponds to a census block group in California, and 8 features: latitude, longitude, median income, population, house age, average rooms per household, average bedrooms per household, and average occupancy per household. Our goal is to predict the median home value of each census block group. We note two interesting details about this problem. First, we expect median home value to be highly associated with median income since individuals with higher incomes are more likely to purchase more expensive homes. Second, we expect certain geographical regions in California, such as the San Francisco Bay Area and the Los Angeles Metropolitan Area to have higher home values.

\begin{figure}[h]
    \centering
    \includegraphics[width=0.85\linewidth]{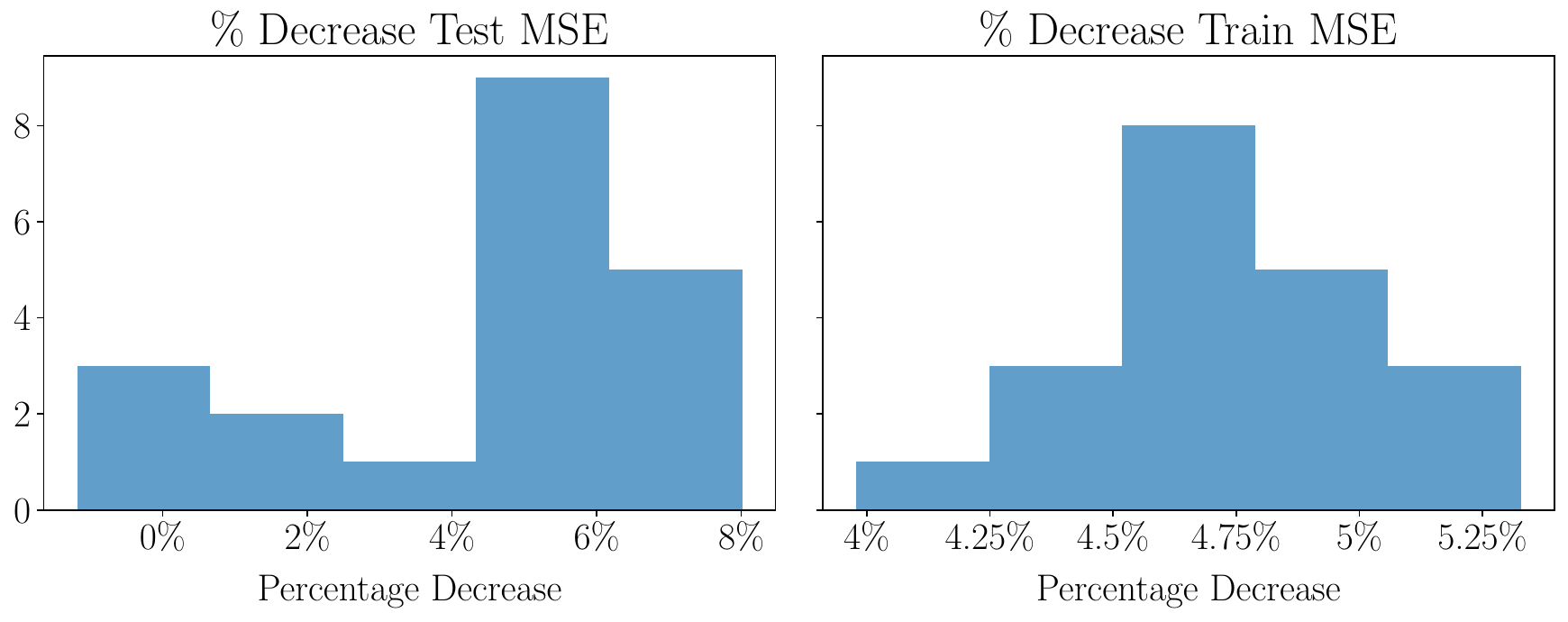}
    \caption{Distributions of percent decreases in training and test MSE for random forests vs. bagging. Random forests consistently decrease training error over bagging in the California Housing example.}
    \label{CA_housing_hist.fig}
\end{figure}

We perform a 20-fold cross validation and on each fold we fit a bagging ensemble with $mtry = 1.0$  and a random forest with $mtry = 0.333$ of 2000 full-depth decision trees. We report the percent decrease in training and test mean-squared error between the two ensembles, given by:
\begin{align*}
   \frac{\text{MSE(bagging)} - \text{MSE(random forest)}}{\text{MSE(bagging)}} \times 100\%,
\end{align*} where percent decreases above 0\% indicate that random forests outperform bagging. The distributions of these percent decreases across all folds are shown in Figure \ref{CA_housing_hist.fig}. We note from the right histogram that random forests consistently reduce the training error of the ensemble over bagging by around $4\%$. This corresponds to an average decrease in test MSE of around $4.4\%$ as well. From these plots, we observe that our random forest consistently outperforms bagging on this example.

\begin{figure}[h]
    \centering
    \includegraphics[width=0.75\linewidth]{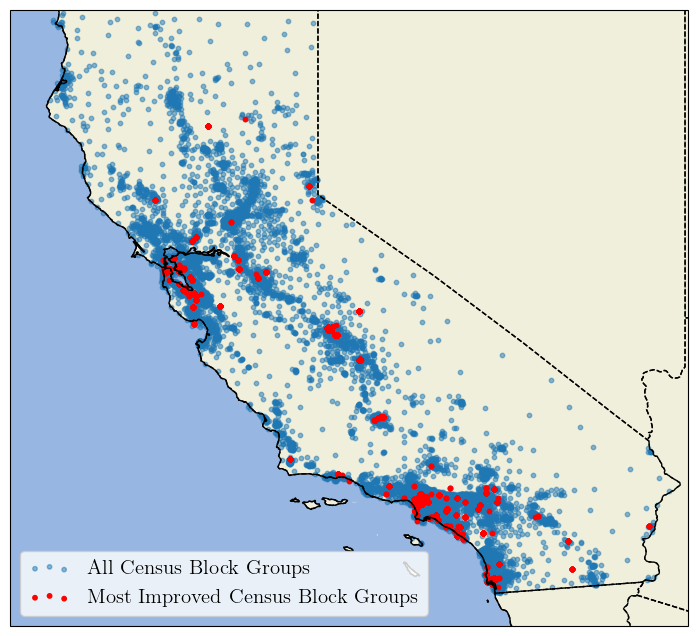}
    \caption{Visualization of census block groups in California. The census block groups where random forests improve the most over bagging are highlighted in red.}
    \label{CA_map.fig}
\end{figure}

On each fold, we also record the 50 training data points where the difference in squared error between bagging and random forests is the largest. These data points can be viewed as the census block groups where random forests improve the most over bagging. In Figure \ref{CA_map.fig} we show a map of all the census block groups in California, each census block group is shown by a blue point. The red points show the census block groups where random forests reduce training error the most compared to bagging, across all folds. From this plot, we observe that random forests improve over bagging the most in the SF Bay Area and the LA Metropolitan Area.

We hypothesize that the latitude and longitude features in this dataset contain hidden patterns similar to the one in the HiddenSquare model discussed in \S\ref{hidden_square_circle.section}. Census block groups within the latitude-longitude boundaries of the SF Bay Area or LA Metropolitan Area are likely to have significantly increased median home values compared to surrounding regions. These hidden patterns, however, may be obscured by other influential features such as median income, which has a nearly linear relationship with median home value (appendix D.1). Random forests may help uncover this hidden pattern; this would explain the consistent decrease in training error compared to bagging shown in Figure \ref{CA_housing_hist.fig} as well as the geographical distribution in improvements shown in Figure \ref{CA_map.fig}.

We test our hypothesis using this sample splitting procedure. Note that in this real-world example, the underlying hidden pattern function $f_H(\text{latitude},\text{longitude})$ is unknown. To estimate this function, we first randomly take a 50-50 split of the dataset. On the first half of the data, we fit a bagging ensemble of 1000 decision trees to predict housing price using \emph{only} the hidden features in $H = \{\text{latitude}, \ \text{longitude}\}$. On the second half of the data, we use this bagging ensemble to produce a plug-in estimate for the underlying hidden pattern function: $\hat{f_H}(X_H)$. We then fit the following ensembles to predict housing price on the second half of the dataset:
\begin{itemize}
    \item $g(X)$: a bagging ensemble that uses all of the features. 
    \item $g(X_{M^c})$: a bagging ensemble that uses all of the features except median income, our hypothesized masking feature $X_M$.
    \item $r(X)$: a random forest with $mtry = 0.333$.
\end{itemize}
We repeat this data splitting procedure 500 times to estimate the following quantities.
\begin{table}[h]\centering
\begin{tabular}{l|l|l|l|}
\cline{2-4}
                                        & $\bigl(E[\hat{f_H}(X_H)- \hat{g}(X)]\bigr)^2$ & $\bigl(E[\hat{f_H}(X_H) - \hat{g}(X_{M^c})] \bigr)^2$ & $\bigl( E[\hat{f_H}(X_H) - \hat{r}(X)] \bigr)^2$ \\ \hline
\multicolumn{1}{|l|}{\textbf{CAHousing}} & \multicolumn{1}{c|}{0.240}              & \multicolumn{1}{c|}{0.223}                              & \multicolumn{1}{c|}{0.231}                \\ \hline
\end{tabular}
\end{table}

The distributions for these estimates can be found in \S D.2 of the appendix. These results support the hypothesis that features latitude and longitude contain a hidden pattern masked by median income and that random forests help capture this feature.

\section{Bias Reduction, $mtry$, and Random Forest Tuning}\label{mtry.section} \color{black}
We conclude this paper by analyzing the effect of the $mtry$ parameter on bias reduction in random forests and its implications for random forest tuning. First, we present an empirical study to explore how $mtry$ impacts bias reduction. We consider two models, hMARS:
\begin{equation}\label{hmars}
\begin{split}
    Y = 10sin(\pi X_1 X_2) + 20(X_3 - 0.05)^2 + 10X_4 +5 X_5  - 30 \times \mathbbm{1}( 0.6 \leq X_6 \leq 0.65) \\ - 35 \times \mathbbm{1}( 0.55 \leq X_7 \leq 0.6) + \epsilon,
    \end{split}
\end{equation}
and hMARSadd:
\begin{equation}\label{hmars_add}
\begin{split}
    Y = 0.1 e^{4 X_1} + \frac{4}{1+ e^{-20(X_2 - 0.5)}} + 3 X_3 + 2 X_4 + X_5  - 10 \times \mathbbm{1}( 0.6 \leq X_6 \leq 0.65) \\ - 7.5 \times \mathbbm{1}( 0.55 \leq X_7 \leq 0.6) + \epsilon,
    \end{split}
\end{equation}
where each model contains hidden patterns in features $X_6$ and $X_7$ that are similar to the ones found in our Hidden2d model. Using each model, we generate $n = 1000$ observations by sampling $X$ from a $U(0,1)$ distribution, with $\epsilon$ set such that the SNR $= 6$. We repeat 500 trials of the Monte Carlo estimation procedure used in \S\ref{empirical_study.section}  while varying $mtry \in \{0.1, 0.2, \ldots, 0.9, 1.0 \}$ to estimate the bias and variance of bagging ensembles ($mtry = 1.0$) and random forests ($mtry < 1.0$). 
\begin{figure}[h]
    \centering
    \includegraphics[width = \textwidth]{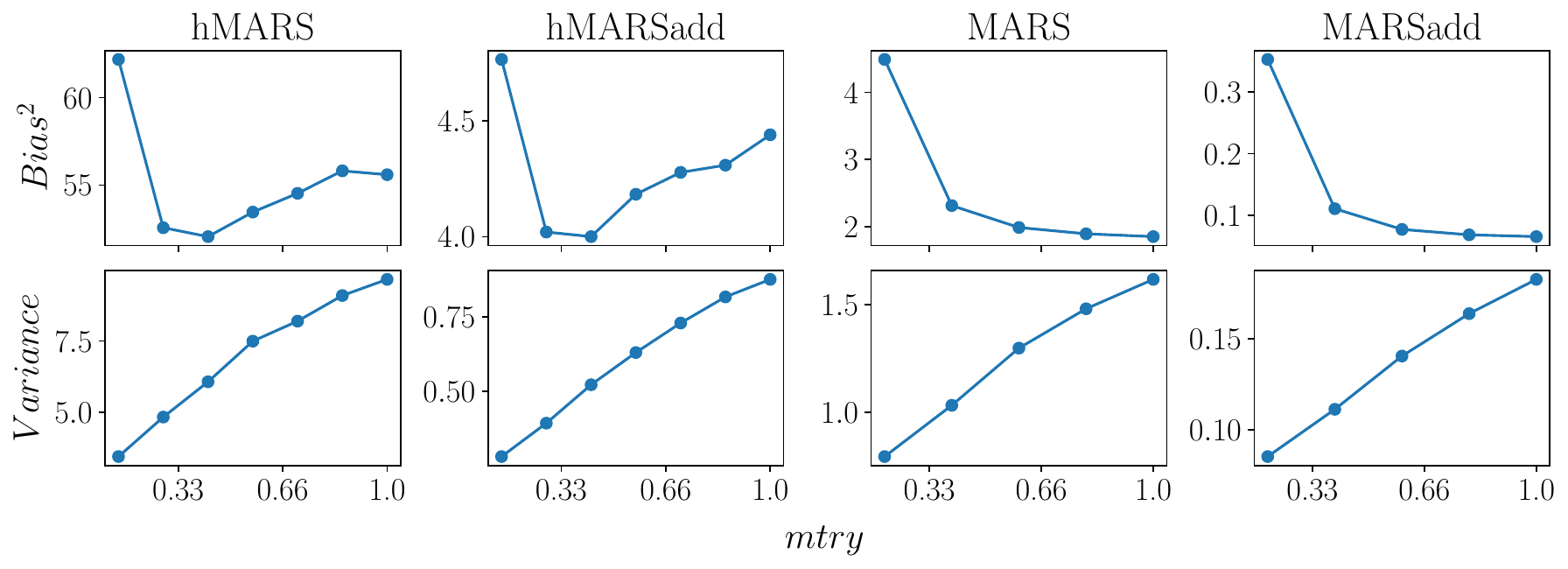}
    \caption{Bias-variance decomposition for random forests and bagging ensembles while varying $mtry$; the right-most points show bagging ensembles with $mtry = 1.0$.}
    \label{SNR_BVD_6.fig}
\end{figure}

Figure \ref{SNR_BVD_6.fig} shows the results of this study. We see that on the hMARS and hMARSadd models, increasing randomization by reducing $mtry$ from 1.0 to 0.333 reduces both bias and variance. However, we also observe that setting $mtry$ to the smallest value drastically increases bias. In this case, a single randomly selected feature is considered for each split and the resulting ensemble has high bias. We see here that selecting an appropriate value for $mtry$ is crucial for reducing bias in a random forest.

\subsection{Non-informative Features and $mtry$} \label{noninformative.section}
In this section, we empirically demonstrate that selecting the best value of $mtry$ for bias reduction also depends on the number of non-informative features in the data.
\begin{figure}[h]
    \centering
    \includegraphics[width = 0.8\textwidth]{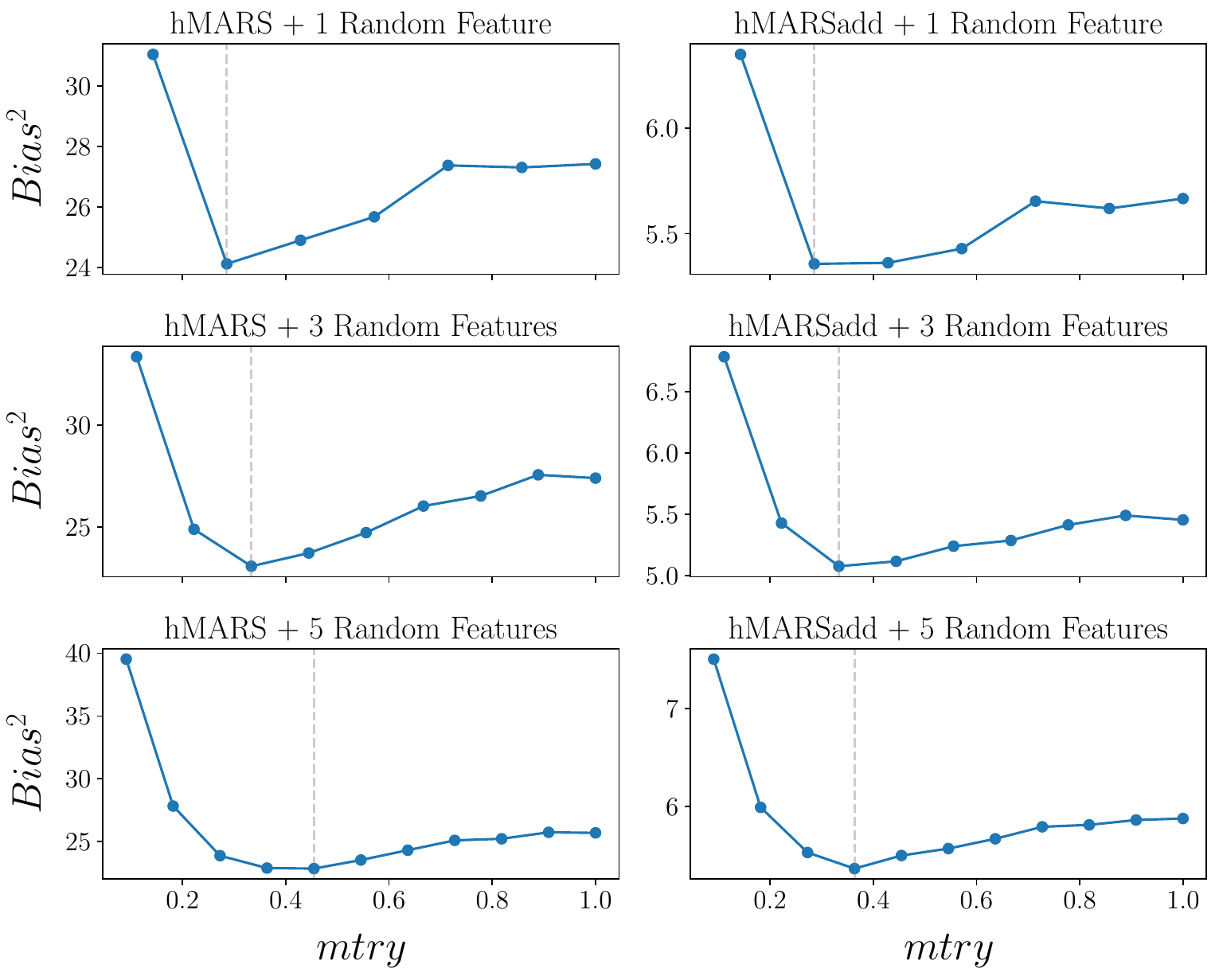}
    \caption{Adding non-informative features to the hMARS and hMARSadd datasets increases the best value of $mtry$ for bias reduction.}
    \label{bias_random_feats.fig}
\end{figure}

Consider the case where we add 1, 3, and 5 non-informative (randomly generated) features to the hMARS and hMARSadd dataset (with 1000 datapoints and SNR = 6) and repeat the bias-variance analysis conducted above.  We show in Figure \ref{bias_random_feats.fig} the bias term (vertical axes) plotted against $mtry$ (horizontal axes) for each dataset\footnote{We show plots of the variance terms in the Appendix E.}. The vertical dashed gray line in each panel shows the value of $mtry$ that minimizes the bias term. From this figure, we observe that as the number of non-informative random features increases, the value of $mtry$ that corresponds to the smallest bias increases as well. We explore why this occurs below.

In this simulation study, we empirically demonstrate that $mtry$ controls the trade-off between the ability of a random forest to capture hidden patterns and its sensitivity to non-informative features. As such, it may be beneficial in certain scenarios to increase $mtry$ on datasets with many non-informative features. Each potential split in a random forest considers a random subset of  $\ceil{mtry \times p}$ features. Consider the case where in this subset, every single feature is non-informative. The resulting split would be essentially random, i.e., a \emph{non-informative} split, which may increase bias.

\begin{figure}[h]
    \centering
    \includegraphics[width = 0.8\textwidth]{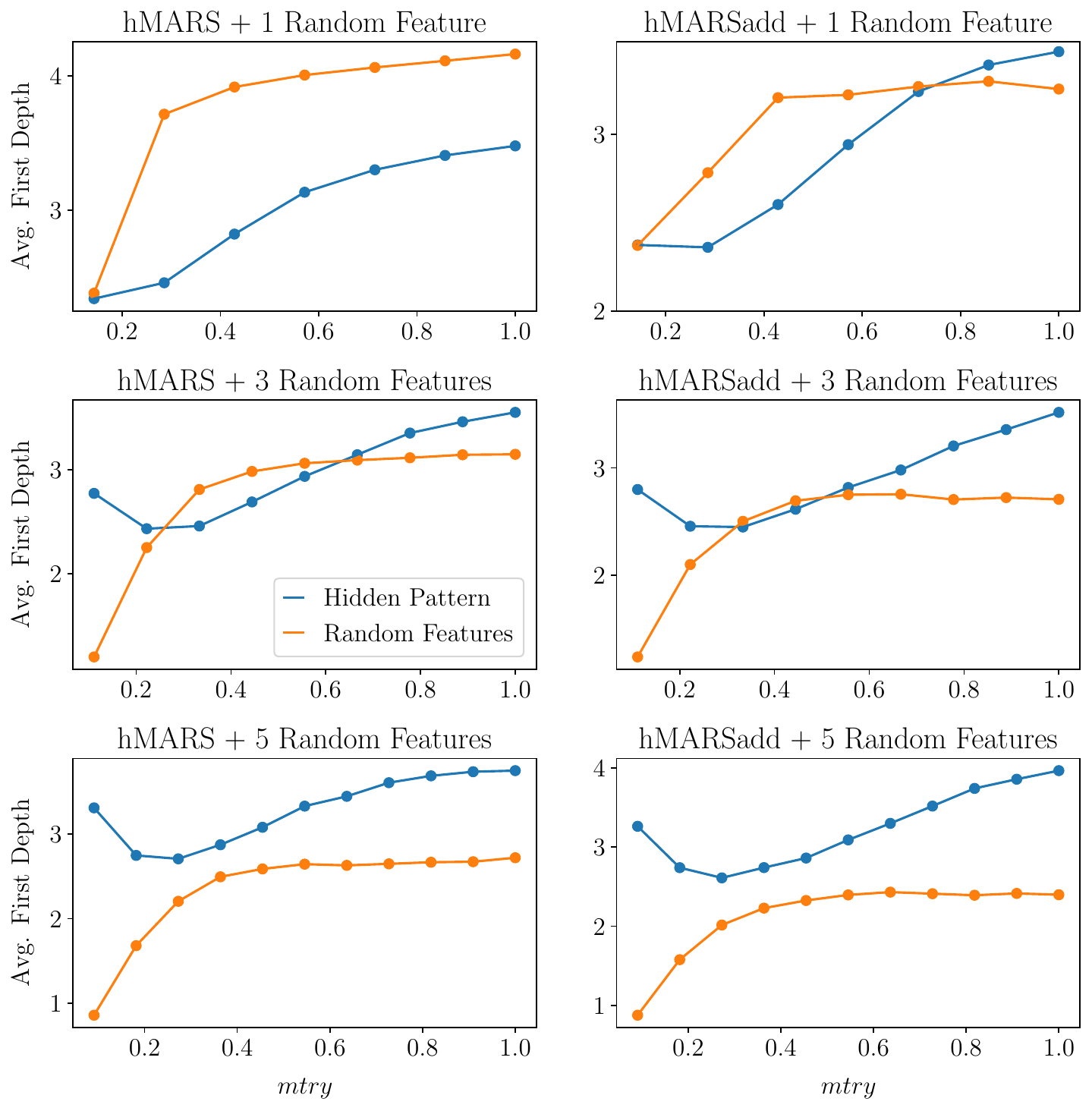}
    \caption{Feature importance measured via average first depth for hMARS and hMARSadd datasets (1000 datapoints and SNR = 6) with added random features. Lower values of average first depth means that the feature is more influential to the model.}
    \label{avg_feature_depth.fig}
\end{figure}

We use this procedure to assess how the influence of \emph{non-informative} splits, splits that only consider non-informative features, varies with $mtry$. For each feature, we examine in each tree the depth of the first split that uses the feature; this can be thought of as a proxy for how influential the feature is \citep{breiman2017classification}. We average across all trees in the ensemble to obtain the average first depth for each feature;  features with a lower average first depth should be more influential.  Following the procedure discussed above, we assess how the average first depth of non-informative features and features with hidden patterns  vary with $mtry$ when we fit random forests on the hMARS and hMARSadd models with $n = 1000$ and SNR = 6.  The results are shown in Figure \ref{avg_feature_depth.fig}. 

In this figure, the horizontal axes show $mtry$ and the vertical axes show the average first depth for the non-informative features (orange) and the features with hidden patterns (blue). We observe that in all panels, decreasing $mtry$ beyond a certain point drastically decreases the average first depth of non-informative features, which corresponds to a large increase in the influence of non-informative splits. We also observe that decreasing $mtry$ up to a certain point decreases the average first depth of hidden pattern features, which corresponds to an increase in the influence of features with hidden pattern. Here, $mtry$ balances the influence of features with hidden patterns with the influence of non-informative features. Increasing the influence of features with hidden patterns can reduce bias, but increasing the influence of non-informative features can increase bias. 
\color{black}
 It is important to mention that adding random features to a dataset and inducing non-informative splits in a random forest reduces the effective DoF of the ensemble, a key insight highlighted by \cite{mentch2022getting}. The authors propose augmented bagging, where non-informative features are injected into the dataset, and show that the procedure can improve the predictive performance of bagging ensembles in low SNR settings. In low SNR settings, increasing the influence of non-informative features may in fact improve the predictive performance of the model by reducing variance.

In many real-world datasets the underlying data generating procedure, as well as the SNR of the data, the presence of hidden patterns, and the number of non-informative features, are unknown. Consequently, it may be important to tune $mtry$, as well as other parameters in the random forest, to improve performance. In the section below, we explore the effects of tuning $mtry$, and other parameters, on the predictive performance of random forests on real world data.

\subsection{Exploring Random Forest Tuning} \label{tuning.experiment}

Throughout this paper, we mention several parameters in a random forest that can improve predictive performance through different mechanisms. We focus on the $mtry$ parameter, and show that $mtry$ can reduce both bias and variance. In \S\ref{rf_effective_dof.section}, we also discuss the TRIM procedure which reduces the effective DoF and variance of an ensemble by controlling tree size. \cite{zhou2023trees} further explores this idea, and the authors show that tuning tree depth can improve predictive performance in low SNR settings. Finally, in the section above, we mention the augmented bagging procedure introduced in \cite{mentch2022getting}. Adding non-informative noise features to a dataset can also regularize bagging ensembles when SNR is low.

\begin{table}[h]
\centering
\begin{tabular}{|c|c|c|c|c|}
\hline
\textbf{Dataset Name} & \textbf{Observations} & \textbf{Features} & \textbf{$R^2$ Bagging} & \textbf{Best Tuning Method} \\ \hline
autoMpg               & 398                   & 25                & 0.818                  & $mtry$                      \\ \hline
no2                   & 500                   & 7                 & 0.602                  & $mtry$                      \\ \hline
boston                & 506                   & 22                & 0.877                  & tree depth                  \\ \hline
stock                 & 950                   & 9                 & 0.985                  & $mtry$                      \\ \hline
socmob                & 1156                  & 39                & 0.804                  & $mtry$                      \\ \hline
spacega              & 3107                  & 6                 & 0.662                  & tree depth                  \\ \hline
abalone               & 4177                  & 10                & 0.546                  & tree depth                  \\ \hline
winequality          & 6497                  & 11                & 0.528                  & $mtry$                      \\ \hline
wind                  & 6574                  & 14                & 0.783                  & tree depth                  \\ \hline
puma32H               & 8192                  & 32                & 0.933                  & $mtry$                      \\ \hline
bank32nh              & 8192                  & 32                & 0.513                  & $mtry$                      \\ \hline
cpusmall             & 8192                  & 12                & 0.977                  & $mtry$                      \\ \hline
kin8nm                & 8192                  & 8                 & 0.718                  & $mtry$                      \\ \hline
pol                   & 15000                 & 48                & 0.988                  & $mtry$                      \\ \hline
elevators             & 16599                 & 18                & 0.839                  & $mtry$                      \\ \hline
houses                & 20640                 & 8                 & 0.821                  & $mtry$                      \\ \hline
house16H             & 22784                 & 16                & 0.635                  & $mtry$                      \\ \hline
mv                    & 40768                 & 14                & 0.982                  & $mtry$                      \\ \hline
\end{tabular}
\caption{List of datasets used in our parameter tuning experiment. The datasets considered span a range of dimensions and SNRs. The rightmost column shows the  tuning method that works the best on each dataset.}
\label{data_details.table}
\end{table}

In this section, we compare the impact of tuning these parameters on real-world data, following the experimental procedure discussed below. We use 17 regression datasets from OpenML and use 5-fold CV on each dataset; dataset names and dimensions are shown in the 3 leftmost columns of Table \ref{data_details.table}. We split each training fold 70-30 into a training set and validation set and fit bagging ensembles with 1000 decision trees, while tuning the following methods.
\begin{itemize}
    \item We tune $\mathbf{mtry}$ across the following 12 values: \{0.1, 0.18, 0.26, 0.35, 0.43, 0.51, 0.59, 0.67, 0.75, 0.8, 0.86, 0.92\}.
    \item We tune the minimum number of samples per leaf node as a proxy for tree depth; this procedure was used in \cite{zhou2023trees}. We vary \textbf{node size} across the following 12 values: \{1, 3, 5, 0.06$n$, 0.11$n$, 0.17$n$, 0.22$n$, 0.28$n$, 0.33$n$, 0.39$n$, 0.44$n$, 0.5$n$\}, where $n$ is the number of observations in the dataset.
    \item We tune the number of additional \textbf{noise features} $q$ and the \textbf{correlation} $r$ of the features added in augmented bagging, following the procedure discussed in \cite{mentch2022getting}. We  vary $q \in \{p/2, \ p, \ 3p/2, \ 2p\}$, where $p$ is the number of features in the dataset, and $r \in \{0, \ 0.4, \ 0.9 \}$, where $r$ is the correlation between each noise feature and a feature in the dataset. This corresponds to a grid of 12 settings for $q$ and $p$.
\end{itemize}

For each method, we select the parameters that yield the minimum validation error, refit the model on the entire training set, and evaluate the $R^2$ of the model on the test set.

\begin{figure}[h]
    \centering
    \includegraphics[width=0.9\linewidth]{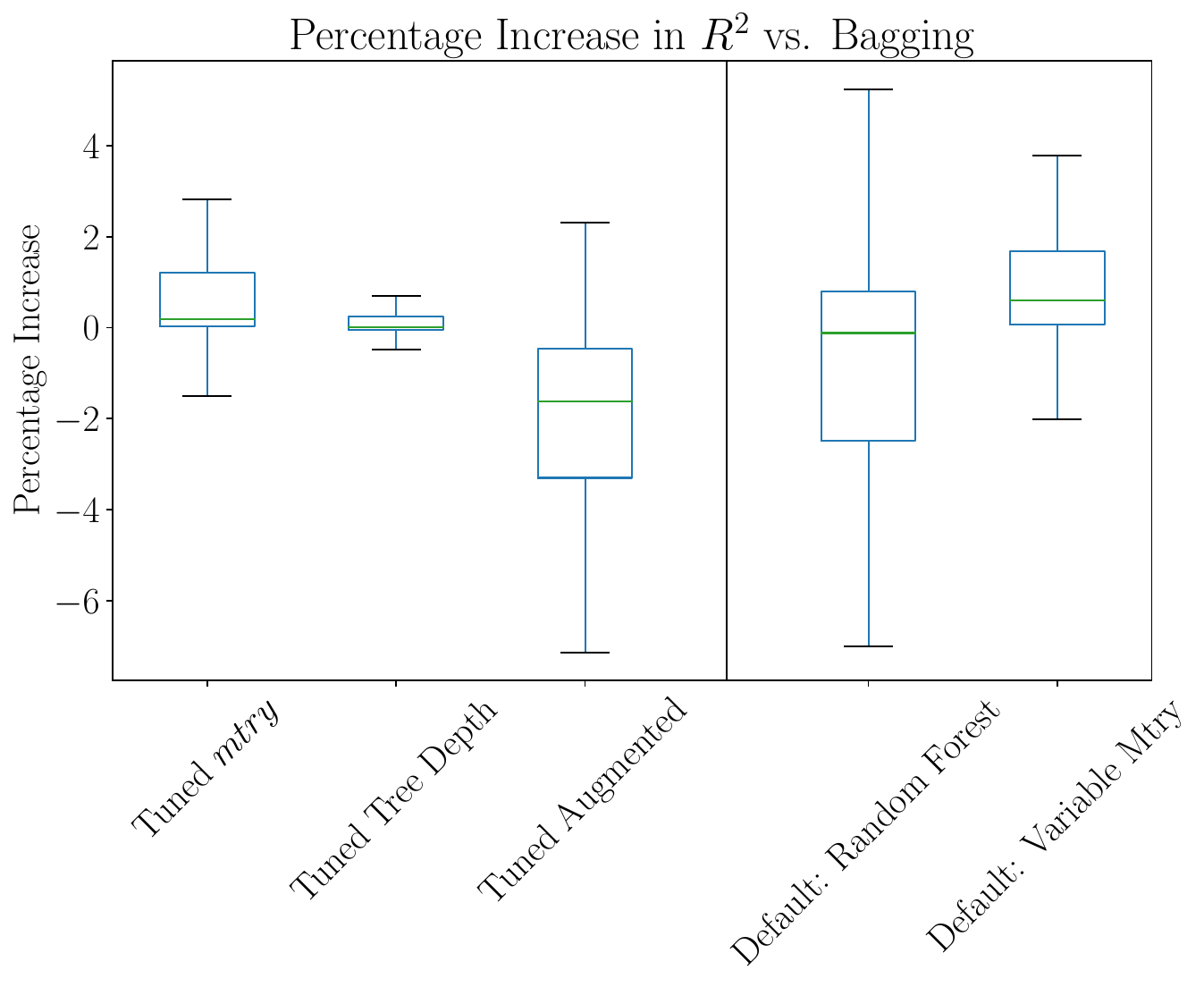}
    \caption{Comparison of various tuning methods against bagging. The vertical axis shows the percentage increase in $R^2$, values above 0 indicate the method outperforms bagging.}
    \label{real_world_study.fig}
\end{figure}

Across all datasets and folds in our experiment, we report the percent increase in test $R^2$ between the best model found by each tuning method and a standard bagging ensemble, of 1000 decision trees grown to full depth with $mtry = 1.0$. This is given by the expression: 
\begin{align*}
    \frac{\text{Tuned Method }R^2 - \text{Bagging } R^2} {\text{Tuned Method } R^2}.
\end{align*}
We show the distribution of our results in the left three boxplots of Figure \ref{real_world_study.fig}. The vertical axis shows the percentage increase in $R^2$ for each method compared to bagging; higher values indicate greater improvements of the tuned methods over bagging.

From this plot, we see that tuning $mtry$ outperforms both tuning tree depth and tuning augmented bagging. We believe that this is due in part to the fact that the datasets we consider in our experiment span a wide range of SNRs. In the rightmost column of Table \ref{data_details.table}, we show the $R^2$ of the standard bagging ensembles (1000 full-depth trees) fit on each dataset, averaged across folds.  In several cases, bagging ensembles achieve $R^2$ scores above 0.80, which suggests that the SNR of the dataset is high.  Since $mtry$ can reduce both bias and variance, tuning this parameter appears to be effective across a variety of real-world datasets. In the rightmost column of Table \ref{data_details.table}, we report the best tuning method for each dataset in our experiment; across the majority of datasets, tuning $mtry$ performs the best.

\subsubsection{Default values for $mtry$}

We conclude with some brief remarks on the default value for $mtry$. One main advantage of random forests---particularly when compared to other methods such as boosting---is that they are considered easier to use “off the shelf,” requiring relatively little parameter tuning (see \cite{efron2021computer}, Chapter 17). This aligns with the classical view that random forests primarily reduce variance, whereas boosting ensembles mainly reduce bias \citep{efron2021computer}, and, as such, random forests are often fit with $mtry$ set to the default value of $0.333$ \citep{breiman2001random,efron2021computer,mentch2020randomization}


In our paper, we show that $mtry$ controls both bias and variance reduction. As such, we would expect tuning $mtry$ to improve the predictive performance of random forests on real world datasets, which span a range of SNRs, and we show here that this is the case. In the second from the right plot of Figure \ref{real_world_study.fig}, we show the percent increase in $R^2$ between a random forest with the default $mtry = 0.333$ over a bagging ensemble. Compare this plot with the leftmost plot in Figure \ref{real_world_study.fig}, which shows the percent increase in $R^2$ between a tuned random forest over bagging. We see that tuning $mtry$ consistently improves the performance of the ensemble.

Recall from \S\ref{noninformative.section} that while reducing $mtry$ can help random forests capture hidden patterns, doing so may also increase the influence of non-informative splits, which may increase bias. Now, consider non-informative splits in the context of individual decision trees in the ensemble, which have hierarchal structure. Influential non-informative splits in the top layers (for example depth $<$ 3) of a decision tree may not significantly increase bias \emph{if} subsequent splits in the deeper layer of the tree can correct for their effect. However, $mtry$ typically applies to all splits in a tree regardless of depth, so this correction is unlikely to occur.

With this in mind, we propose a new default setting for $mtry$ in random forests, called \textbf{variable} $\mathbf{mtry}$, where we only randomize splits in the top layers of each tree in the ensemble. For each tree, we set $mtry = 0.333$ for splits of depth 1 and 2, and then we set $mtry = 1.0$ for the remainder of the deeper splits, until the tree is grown to full depth. We aim to detect hidden patterns in the top layers of a decision tree while letting lower layers correct for the non-informative splits introduced by setting $mtry < 1.0$.

We evaluate this default setting using the procedure discussed above. The rightmost plot in Figure \ref{real_world_study.fig} shows the percent increase in $R^2$ of our variable $mtry$ method compared to standard bagging. We see that using the variable $mtry$ method outperforms the default random forest method of setting $mtry = 0.333$ across all splits in the trees. In fact, our variable $mtry$ default even outperforms tuning $mtry$. These results suggest that randomizing only the top layer splits in a random forest may allow the ensemble to capture hidden patterns while decreasing the impact of non-informative splits. Varying $mtry$ by depth-layer presents an interesting direction for future research and may potentially yield new default parameter settings for random forests that improve predictive performance while preserving their ease of use.

\color{black}
\section{Conclusion}

In this paper, we show that random forests can reduce both bias and variance compared to bagging. Our findings help explain the observed success of random forests across a wide range of SNRs and allow us to re-examine the longstanding notion that random forests only reduce variance. We introduce the concept of hidden patterns, patterns in the data missed by bagging, and show that random forests can better capture these patterns by randomizing the features considered per split. Empirically, we show that this leads random forests reducing both bias and variance on datasets with hidden patterns. Using these findings, we analyze the impact of $mtry$ on bias reduction in random forests and offer several practical insights towards parameter tuning in random forests.

 \section{Acknowledgments} We would like to thank the anonymous referees for their thoughtful comments that resulted in significant improvements in the manuscript.  This research is funded in part by a grant from the Office of Naval Research (ONR-N00014-21-1-2841).

\newpage




\newpage

\newpage

\newpage

\newpage
\appendix
\section*{Appendix A.}

\subsection*{Appendix A.1}
\begin{table}[h]
\centering
\begin{minipage}{0.48\linewidth}
\centering
\begin{tabular}{|l|c|c|}
\hline
\multicolumn{1}{|c|}{SNR}  & \begin{tabular}[c]{@{}c@{}}Trimmed \\ Max Nodes\end{tabular} & \begin{tabular}[c]{@{}c@{}}Bagging/RF \\ Max Nodes\end{tabular} \\ \hline
\multicolumn{1}{|c|}{0.05} & 175                                                          & 200                                                             \\ \hline
0.085                      & 160                                                          & 200                                                             \\ \hline
\multicolumn{1}{|c|}{0.145} & 155                                                          & 200                                                             \\ \hline
0.247                      & 150                                                          & 200                                                             \\ \hline
0.42                       & 145                                                          & 200                                                             \\ \hline
0.715                      & 140                                                          & 200                                                             \\ \hline
1.216                      & 90                                                           & 200                                                             \\ \hline
2.071                      & 80                                                           & 200                                                             \\ \hline
3.525                      & 55                                                           & 200                                                             \\ \hline
\multicolumn{1}{|c|}{6.0}  & 45                                                           & 200                                                             \\ \hline
\end{tabular}
\caption{Tuned $maxnodes$ values, to match the effective DoF of a random forest, on the MARS dataset.}
\end{minipage}
\hfill
\begin{minipage}{0.48\linewidth}
\centering
\begin{tabular}{|l|c|c|}
\hline
\multicolumn{1}{|c|}{SNR}  & \begin{tabular}[c]{@{}c@{}}Trimmed \\ Max Nodes\end{tabular} & \begin{tabular}[c]{@{}c@{}}Bagging/RF \\ Max Nodes\end{tabular} \\ \hline
\multicolumn{1}{|c|}{0.05} & 180                                                          & 200                                                             \\ \hline
0.085                      & 165                                                          & 200                                                             \\ \hline
\multicolumn{1}{|c|}{0.145} & 155                                                          & 200                                                             \\ \hline
0.247                      & 150                                                          & 200                                                             \\ \hline
0.42                       & 145                                                          & 200                                                             \\ \hline
0.715                      & 135                                                          & 200                                                             \\ \hline
1.216                      & 90                                                           & 200                                                             \\ \hline
2.071                      & 80                                                           & 200                                                             \\ \hline
3.525                      & 60                                                           & 200                                                             \\ \hline
\multicolumn{1}{|c|}{6.0}  & 45                                                           & 200                                                             \\ \hline
\end{tabular}
\caption{Tuned $maxnodes$ values, to match the effective DoF of a random forest, on the MARSadd dataset.}
\end{minipage}
\end{table}

\clearpage
\newpage

\subsection*{Appendix A.2}
In this section, we repeat the experimental procedure discussed in \S\ref{trim_sfs_simulation.section}, however, instead of our TRIM procedure, we assess the augmented bagging procedure discussed in \cite{mentch2022getting}. We tune $q$, the number of additional noise features to add, and $r$ the correlation between the noise features and informative features, until the effective DoF of the augmented bagging ensemble matches that of a SFS random forest. We report the relative difference of each method compared to bagging in the plots below, for both the MARS and MARSadd example. From these plots, we observe that like for TRIM, augmented bagging reduces effective DoF without improving predictive performance on these examples.

\begin{figure}[h]
    \centering
    \begin{subfigure}[b]{0.45\linewidth}
        \centering
        \includegraphics[width=\linewidth]{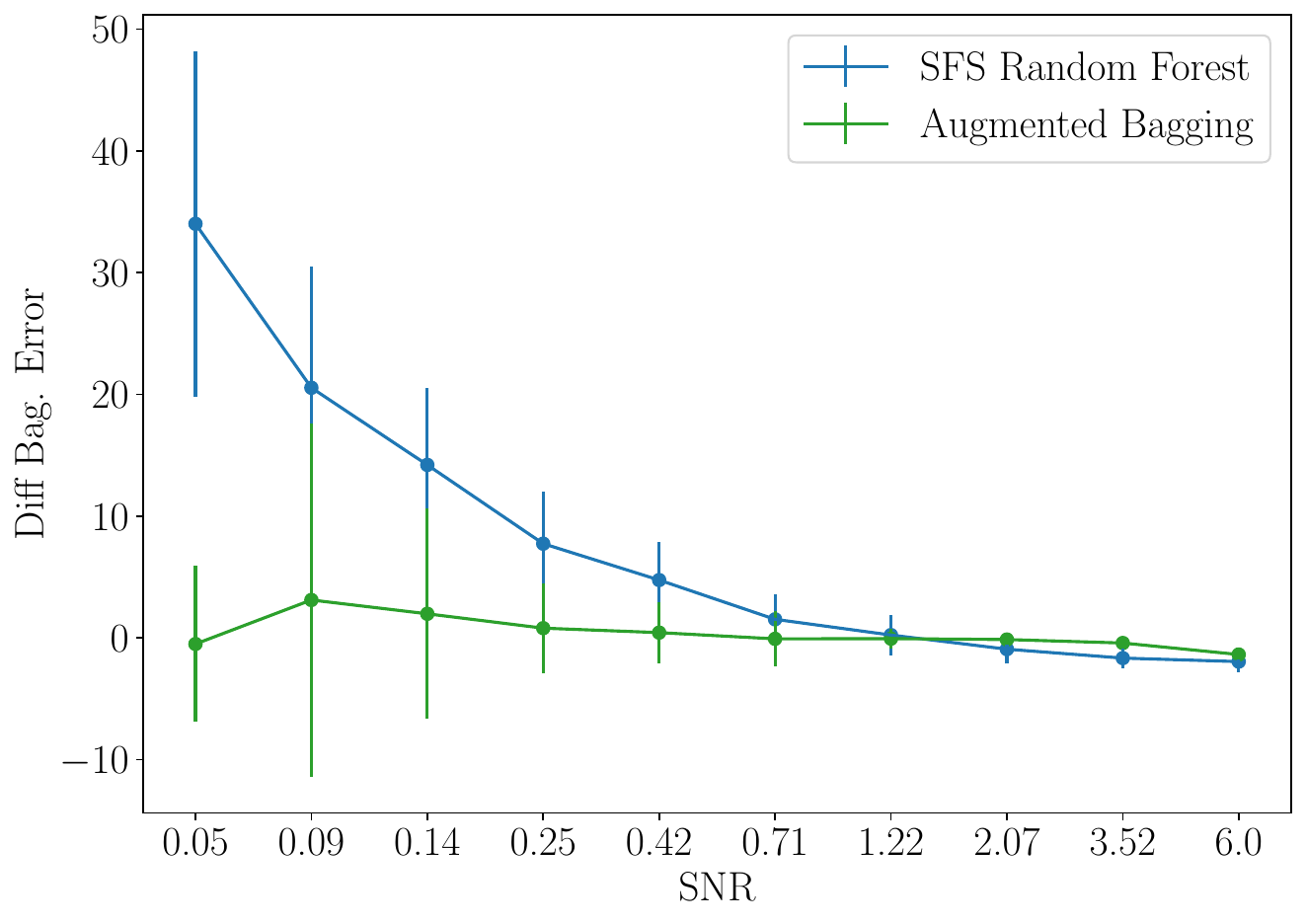}
        \caption{Relative Error of SFS random forests and augmented bagging ensembles on the MARS example.}
        \label{fig:enter-label1}
    \end{subfigure}
    \hfill
    \begin{subfigure}[b]{0.45\linewidth}
        \centering
        \includegraphics[width=\linewidth]{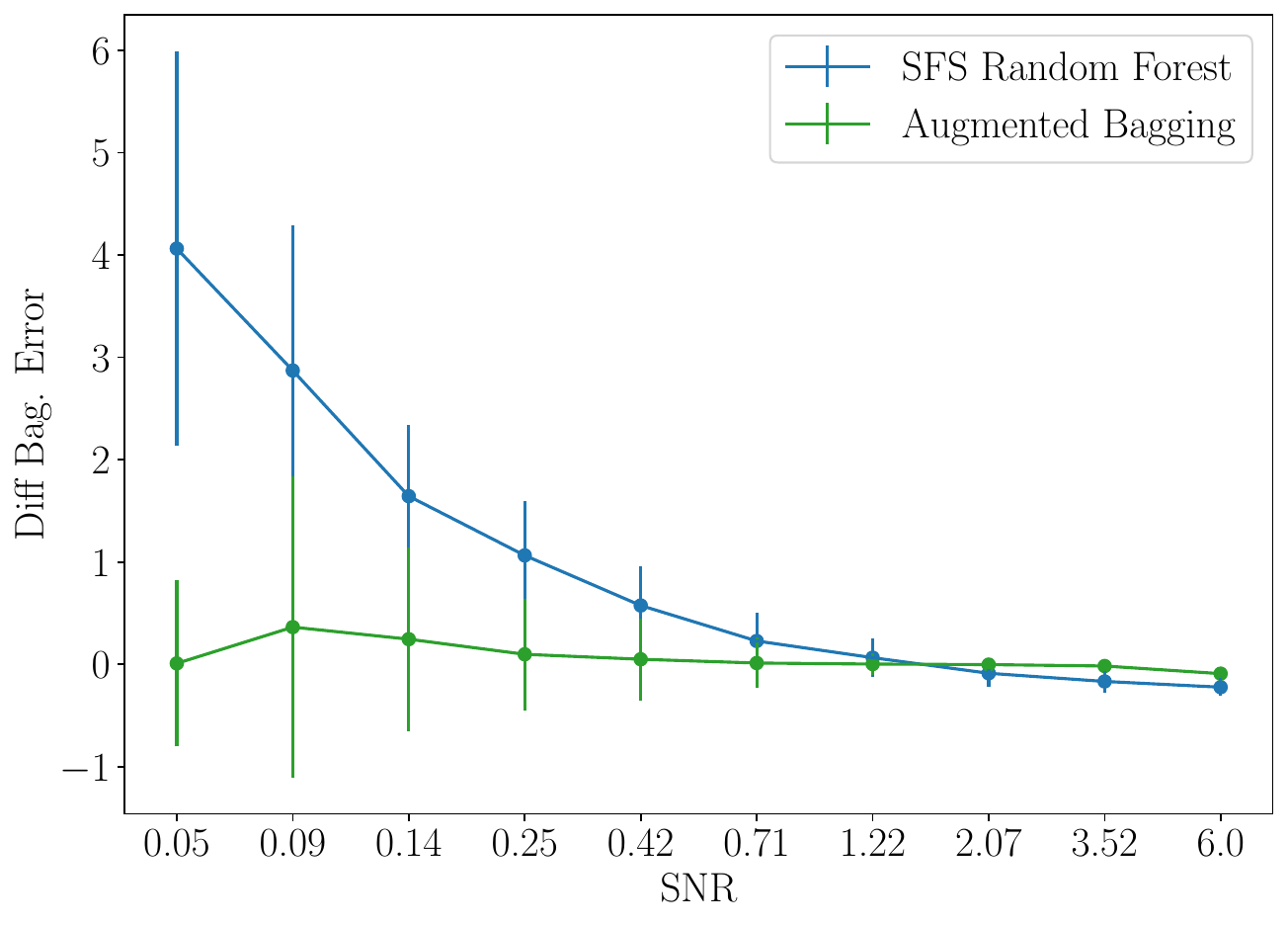}
        \caption{Relative Error of SFS random forests and augmented bagging ensembles on the MARSadd example.}
        \label{fig:enter-label2}
    \end{subfigure}
\end{figure}

\clearpage
\newpage
\section*{Appendix B.}

\subsection*{B.1 Proof of Proposition 2}

We start from the LHS of the expression and plug in \eqref{local_pred_decomp.eq}.
\begin{align*}
    \frac{1}{n} \sum_{i=1}^n \bigl(Y_i - \hat{f}(X^{(i)}))^2 =  \frac{1}{n} \sum_{i=1}^n \biggl (Y_i -  \frac{|\Psi_i|}{|T|} Y_i - \frac{1}{|T|} \sum_{t \in \Omega_i} L_t(X^{(i)})\biggr)^2 
\end{align*}
\begin{align*}
=  \frac{1}{n} \sum_{i=1}^n \biggl (Y_i \bigl(\frac{|T| - |\Psi_i|}{|T|}\bigr) - \frac{1}{|T|} \sum_{t \in \Omega_i} L_t(X^{(i)})\biggr)^2 
\end{align*}
We have that $|T| = |\Psi_i| + |\Omega_i|$.
\begin{align*}
=  \frac{1}{n} \sum_{i=1}^n \biggl ( \frac{|\Omega_i|}{|T|}Y_i - \frac{1}{|T|} \sum_{t \in \Omega_i} L_t(X^{(i)})\biggr)^2 
\end{align*}
\begin{align*}
=  \frac{1}{n} \biggl(\frac{|\Omega_i|}{|T|}\biggr)^2 \sum_{i=1}^n \biggl ( Y_i -\frac{1}{|\Omega_i|} \sum_{t \in \Omega_i} L_t(X^{(i)})\biggr)^2 
\end{align*}
We plug in expression \eqref{mu_def.eq} to get the desired result. When the number of decision trees in a bagging ensemble/random forest is large we have:
\begin{align*}
     \biggl(\frac{|\Omega_i|}{|T|}\biggr)^2 \approx 0.135,
\end{align*}
which follows from the properties of the bootstrap.

\section*{Appendix C.}

\subsection*{C.1 Empirical Study Models}
\begin{itemize}

\item \textbf{friedman2\_2d}
\[
\begin{aligned}
y \;&=\; \sqrt{x_{1}^{2} \;+\;
\Bigl(x_{2}x_{3} \;-\;\frac{1}{x_{2}x_{4} \;+\;0.1}\Bigr)^{2}} \\
&\quad-\;5\,\mathbbm{1}(\,0.625 < x_{5} \le 0.65)\;+\;\epsilon
\end{aligned}
\]

\item \textbf{friedman3\_2d}
\[
\begin{aligned}
y \;&=\; \arctan\!\Bigl(x_{2}x_{3} \;-\;\frac{1}{x_{2}x_{4}}\Bigr) \\
&\quad+\;2\,\mathbbm{1}(\,0.625 < x_{5} \le 0.65)\;+\;\epsilon
\end{aligned}
\]

\item \textbf{polynomial\_2d}
\[
\begin{aligned}
y \;&=\; 2 \;+\; 3.5\,x_{1} \;-\;1.2\,x_{1}^{2}
\;+\;4\,x_{2} \;+\;0.8\,x_{2}^{3} \\
&\quad-\;5\,\mathbbm{1}(\,0.625 < x_{6} \le 0.65)
\;-\;4\,\mathbbm{1}(\,0.575 < x_{7} \le 0.60)
\;-\;6\,\mathbbm{1}(\,0.58 < x_{8} \le 0.60)
\;+\;\epsilon
\end{aligned}
\]

\item \textbf{polynomial\_box}
\[
\begin{aligned}
y \;&=\; 2 \;+\; 3.5\,x_{1} \;-\;1.2\,x_{1}^{2}
\;+\;4\,x_{2} \;+\;0.8\,x_{2}^{3} \\
&\quad-\;10\,\mathbbm{1}(\,0.625 < x_{6} \le 0.70,\;0.625 < x_{7} \le 0.70)
\;+\;\epsilon
\end{aligned}
\]

\item \textbf{polynomial\_spike}
\[
\begin{aligned}
\text{G1}(x_{5}) &= 6\,\exp\!\Bigl(-\tfrac{(x_{5}-0.28)^{2}}{2\,(0.005)^{2}}\Bigr),\\
\text{G2}(x_{5}) &= 6\,\exp\!\Bigl(-\tfrac{(x_{5}-0.30)^{2}}{2\,(0.005)^{2}}\Bigr),
\end{aligned}
\]
\[
\begin{aligned}
y \;=\; 2 \;+\; 3.5\,x_{1} \;-\;1.2\,x_{1}^{2}
\;+\;4\,x_{2} \;+\;0.8\,x_{2}^{3}
\;-\;\text{G1}(x_{5})
\;+\;\text{G2}(x_{5})
\;+\;\epsilon
\end{aligned}
\]

\item \textbf{linear\_2d}
\[
\begin{aligned}
y \;&=\; 0.3\,x_{1} \;-\;0.5\,x_{2}
\;+\;0.1\,x_{3} \;+\;0.1\,x_{4}
\;+\;x_{5} \\
&\quad-\;\mathbbm{1}(\,0.625 < x_{6} \le 0.65)
\;-\;2\,\mathbbm{1}(\,0.575 < x_{7} \le 0.60)
\;-\;0.5\,\mathbbm{1}(\,0.58 < x_{8} \le 0.60)
\;+\;\epsilon
\end{aligned}
\]

\item \textbf{linear\_box}
\[
\begin{aligned}
y \;&=\; 0.3\,x_{1} \;-\;0.5\,x_{2}
\;+\;0.1\,x_{3} \;+\;0.1\,x_{4}
\;+\;x_{5} \\
&\quad-\;3\,\mathbbm{1}(\,0.63 < x_{6} \le 0.65,\;0.63 < x_{7} \le 0.65)
\;+\;\epsilon
\end{aligned}
\]

\item \textbf{linear\_spike}
\[
\begin{aligned}
\text{G1}(x_{5}) &= 4\,\exp\!\Bigl(-\tfrac{(x_{5}-0.28)^{2}}{2\,(0.005)^{2}}\Bigr),\\
\text{G2}(x_{5}) &= 4\,\exp\!\Bigl(-\tfrac{(x_{5}-0.30)^{2}}{2\,(0.005)^{2}}\Bigr),
\end{aligned}
\]
\[
\begin{aligned}
y \;=\; 0.3\,x_{1} \;-\;0.5\,x_{2}
\;+\;0.1\,x_{3} \;+\;0.1\,x_{4}
\;+\;x_{5}
\;-\;\text{G1}(x_{5})
\;+\;\text{G2}(x_{5})
\;+\;\epsilon
\end{aligned}
\]

\item \textbf{hmars\_2d}
\[
\begin{aligned}
y \;&=\; 10\,\sin(\,\pi\,x_{1}x_{2}) 
\;+\;20\,(x_{3}-0.05)^{2}
\;+\;10\,x_{4}
\;+\;5\,x_{5} \\
&\quad-\;30\,\mathbbm{1}(\,0.60 < x_{6} \le 0.65)
\;-\;35\,\mathbbm{1}(\,0.55 < x_{7} \le 0.60)
\;+\;\epsilon
\end{aligned}
\]

\item \textbf{hmars\_box}
\[
\begin{aligned}
y \;&=\; 10\,\sin(\,\pi\,x_{1}x_{2})
\;+\;20\,(x_{3}-0.05)^{2}
\;+\;10\,x_{4}
\;+\;5\,x_{5} \\
&\quad-\;250\,\mathbbm{1}(\,0.65 < x_{6}\le0.70,\;0.65 < x_{7}\le0.70)
\;+\;\epsilon
\end{aligned}
\]

\item \textbf{hmars\_spike}
\[
\begin{aligned}
\text{G1}(x_{6}) &= 100\,\exp\!\Bigl(-\tfrac{(x_{6}-0.28)^{2}}{2\,(0.005)^{2}}\Bigr),\\
\text{G2}(x_{6}) &= 100\,\exp\!\Bigl(-\tfrac{(x_{6}-0.30)^{2}}{2\,(0.005)^{2}}\Bigr),
\end{aligned}
\]
\[
\begin{aligned}
y \;&=\; 10\,\sin(\,\pi\,x_{1}x_{2})
\;+\;20\,(x_{3}-0.05)^{2}
\;+\;10\,x_{4}
\;+\;5\,x_{5} \\
&\quad-\;\text{G1}(x_{6})
\;+\;\text{G2}(x_{6})
\;+\;\epsilon
\end{aligned}
\]

\item \textbf{hmars\_add\_box}
\[
\begin{aligned}
y \;&=\; 0.1\,\exp(4\,x_{1})
\;+\;\frac{4}{\,1 + \exp[-20\,(x_{2}-0.5)]\,}
\;+\;3\,x_{3}
\;+\;2\,x_{4}
\;+\;x_{5} \\
&\quad-\;105\,\mathbbm{1}(\,0.45 < x_{6}\le0.55,\;0.45 < x_{7}\le0.55)
\;+\;\epsilon
\end{aligned}
\]

\item \textbf{hmars\_add\_2d}
\[
\begin{aligned}
y \;&=\; 0.1\,\exp(4\,x_{1})
\;+\;\frac{4}{\,1 + \exp[-20\,(x_{2}-0.5)]\,}
\;+\;3\,x_{3}
\;+\;2\,x_{4}
\;+\;x_{5} \\
&\quad-\;10\,\mathbbm{1}(\,0.60 < x_{6} \le 0.65)
\;-\;7.5\,\mathbbm{1}(\,0.55 < x_{7} \le 0.60)
\;+\;\epsilon
\end{aligned}
\]

\item \textbf{hmars\_add\_spike}
\[
\begin{aligned}
\text{G1}(x_{6}) &= 10\,\exp\!\Bigl(-\tfrac{(x_{6}-0.28)^{2}}{2\,(0.005)^{2}}\Bigr),\\
\text{G2}(x_{6}) &= 10\,\exp\!\Bigl(-\tfrac{(x_{6}-0.30)^{2}}{2\,(0.005)^{2}}\Bigr),
\end{aligned}
\]
\[
\begin{aligned}
y \;&=\; 0.1\,\exp(4\,x_{1})
\;+\;\frac{4}{\,1 + \exp[-20\,(x_{2}-0.5)]\,}
\;+\;3\,x_{3}
\;+\;2\,x_{4}
\;+\;x_{5} \\
&\quad-\;\text{G1}(x_{6})
\;+\;\text{G2}(x_{6})
\;+\;\epsilon
\end{aligned}
\]

\end{itemize}

\newpage 
\section*{Appendix D.}
\subsection*{D.1}
\begin{figure}[h]
    \centering
    \includegraphics[width=0.5\linewidth]{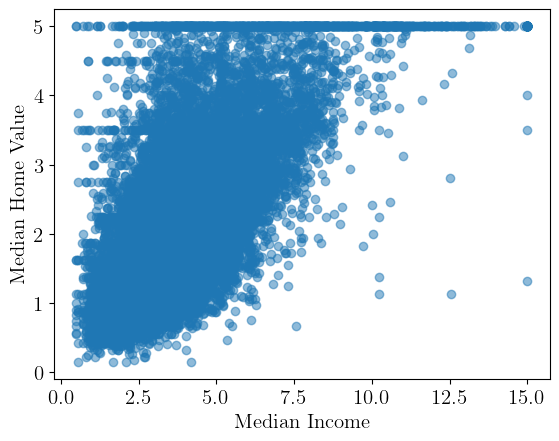}
    \caption{Median home value vs. median income for the California housing dataset.}
    \label{CA_home_income.fig}
\end{figure}
\subsection*{D.2}

\begin{figure}[h]
    \centering
    \includegraphics[width=0.5\linewidth]{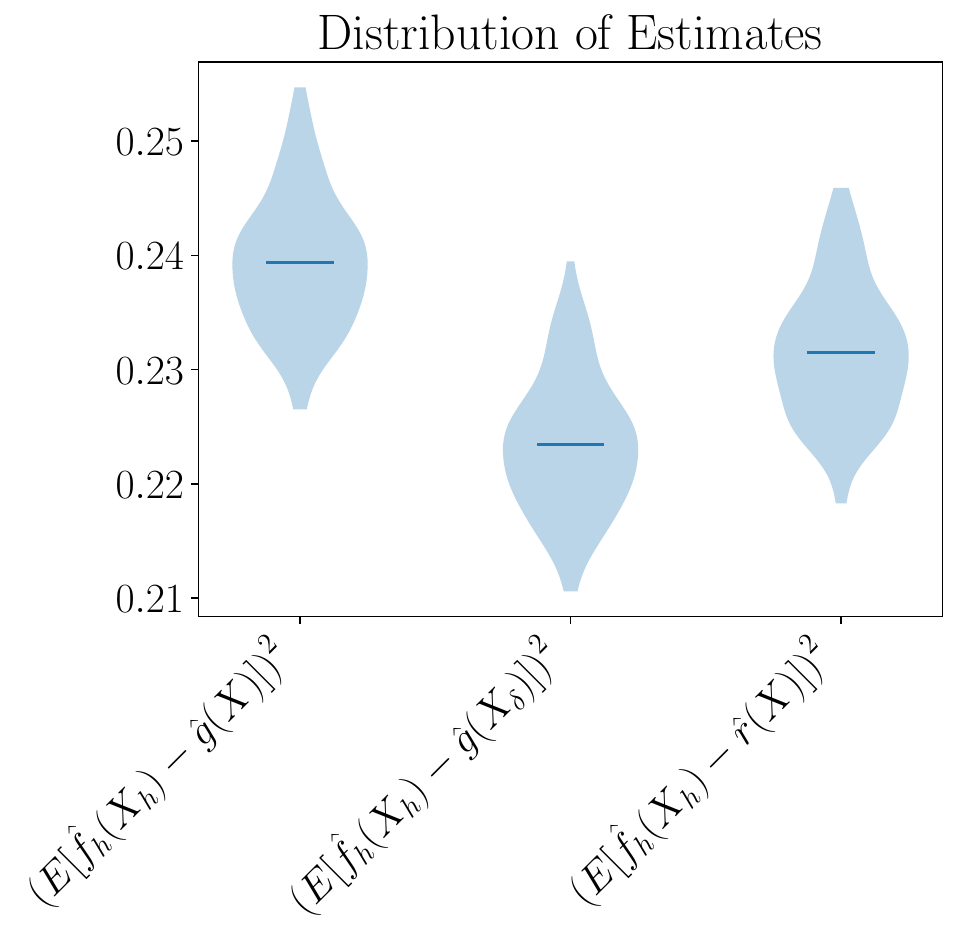}
    \caption{Distribution of estimated quantities in the California housing example.}
    \label{estimate_distribution.fig}
\end{figure}

\clearpage
\newpage

\section*{Appendix E.}

\begin{figure}[h]
    \centering
    \includegraphics[width = 0.8\textwidth]{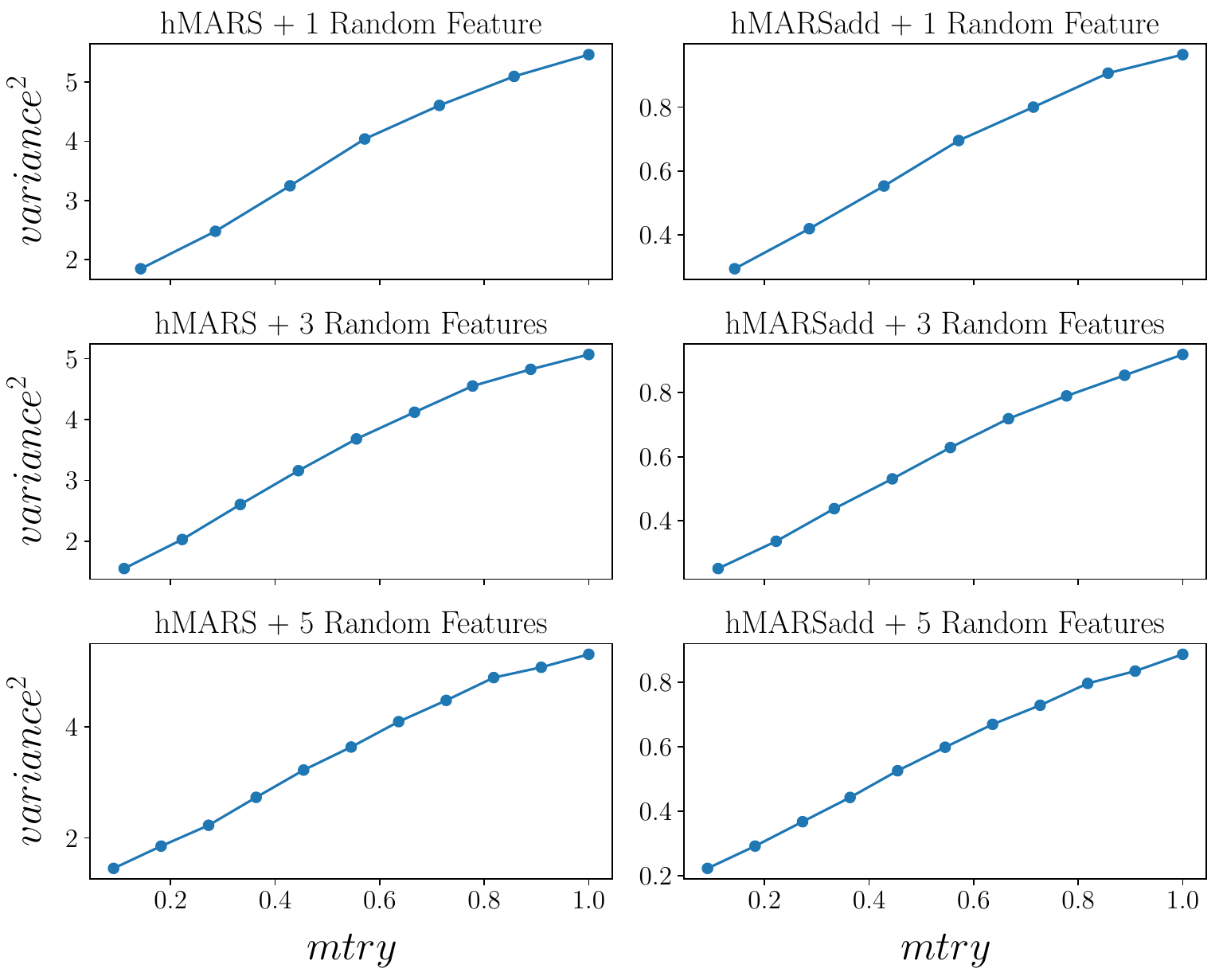}
    \caption{Variance component of bias-variance decomposition for experiment in \S\ref{mtry.section}.}
    \label{variance.fig}
\end{figure}

\newpage
\section*{Appendix F.}
\begin{table}[h] \centering
\begin{tabular}{|c|c|c|}
\hline
\textbf{Dataset Name} & \textbf{$R^2$ Bagging} & \textbf{Best Tuned $mtry$} \\ \hline
autoMpg               & 0.818                  & 0.100                      \\ \hline
no2                   & 0.602                  & 0.787                      \\ \hline
boston                & 0.877                  & 0.542                      \\ \hline
stock                 & 0.985                  & 0.493                      \\ \hline
socmob                & 0.804                  & 0.362                      \\ \hline
spacega               & 0.662                  & 0.902                      \\ \hline
abalone               & 0.546                  & 0.738                      \\ \hline
winequality           & 0.528                  & 0.493                      \\ \hline
wind                  & 0.783                  & 0.869                      \\ \hline
puma32H               & 0.933                  & 0.885                      \\ \hline
bank32nh              & 0.513                  & 0.575                      \\ \hline
cpusmall              & 0.977                  & 0.493                      \\ \hline
kin8nm                & 0.718                  & 0.656                      \\ \hline
pol                   & 0.988                  & 0.673                      \\ \hline
elevators             & 0.839                  & 0.836                      \\ \hline
houses                & 0.821                  & 0.656                      \\ \hline
house16H              & 0.635                  & 0.624                      \\ \hline
mv                    & 0.982                  & 0.885                      \\ \hline
\end{tabular}
\caption{Tuned values of $mtry$, averaged across folds, for each dataset used in the experiments in \S\ref{tuning.experiment}.}
\end{table}
\vskip 0.2in
\bibliography{ref}

\end{document}